\documentclass{article}

\usepackage{arxiv}
\usepackage{framed}

\usepackage{amssymb}
\usepackage{latexsym}
\usepackage{url}

\usepackage{graphicx, subcaption}
\usepackage{amsmath}
\usepackage{amssymb}
\usepackage{booktabs}
\usepackage{multirow}
\usepackage{mathtools}
\usepackage{algorithm}
\usepackage{algpseudocode} 
\usepackage[table]{xcolor}
\usepackage{booktabs,tabularx}  
\usepackage{array}    
\usepackage[framemethod=tikz]{mdframed}
\usepackage{color, soul}
\usepackage{pifont} 
\usepackage{natbib}

\newcommand{\myparagraph}[1]{\smallskip\noindent\textbf{#1}}

\usepackage{hyperref}

\usepackage[capitalize]{cleveref}
\crefname{section}{Sec.}{Secs.}
\crefname{section}{Section}{Sections}
\crefname{table}{Table}{Tables}
\crefname{table}{Tab.}{Tabs.}

\newcommand{\RN}[1]{%
  \textup{\uppercase\expandafter{\romannumeral#1}}%
}
\title{Text-Driven Weakly Supervised OCT Lesion Segmentation with Structural Guidance}%

\author{Jiaqi Yang\thanks{Email: jyang2@gradcenter.cuny.edu} \\
	CUNY Graduate Center \\
	\And
	Nitish Mehta  \\
	New York University Department of Ophthalmology \\ NYU Langone Health \\
        \And
	Xiaoling Hu \\
	Massachusetts General Hospital \\and Harvard Medical School \\
        \And
	Chao Chen \\
	Stony Brook University \\
        \And
	Chia-Ling Tsai \\
	CUNY Queens College \\
}

\renewcommand{\shorttitle}{J. Yang \textit{et~al.}}

\begin{document}
\maketitle

\begin{abstract}
Accurate segmentation of Optical Coherence Tomography (OCT) images is crucial for diagnosing and monitoring retinal diseases. However, the labor-intensive nature of pixel-level annotation limits the scalability of supervised learning for large datasets. Weakly Supervised Semantic Segmentation (WSSS) offers a promising alternative by using weaker forms of supervision, such as image-level labels, to reduce the annotation burden. Despite its advantages, weak supervision inherently carries limited information.
We propose a novel WSSS framework with only image-level labels for OCT lesion segmentation that integrates structural and text-driven guidance to produce high-quality, pixel-level pseudo labels. The framework employs two visual processing modules: one that processes the original OCT images and another that operates on layer segmentations augmented with anomalous signals, enabling the model to associate lesions with their corresponding anatomical layers.
Complementing these visual cues, we leverage large-scale pretrained models to provide two forms of textual guidance: label-derived descriptions that encode local semantics, and domain-agnostic synthetic descriptions that, although expressed in natural image terms, capture spatial and relational semantics useful for generating globally consistent representations. By fusing these visual and textual features in a multi-modal framework, our method aligns semantic meaning with structural relevance, thereby improving lesion localization and segmentation performance.
Experiments on three OCT datasets demonstrate state-of-the-art results, highlighting its potential to advance diagnostic accuracy and efficiency in medical imaging. The code and pretrained models are publicly available at \url{https://github.com/YangjiaqiDig/WSSS-AGM/tree/master/structure_guided}.
\end{abstract}



\section{Introduction}\label{sec:intro}
Semantic segmentation, which categorizes every pixel in an image, is essential in autonomous driving, geospatial analysis, and medical imaging~\citep{schlegl2019f, feng2020deep}. In retinal lesion segmentation, automating lesion detection and retinal conditions holds immense potential to enhance clinical diagnosis and management. Optical Coherence Tomography (OCT) provides high-resolution cross-sectional views of the retina, enabling accurate lesion segmentation as a valuable biomarker for guiding clinical decisions~\citep{schmidt2021ai}. 

While fully supervised methods can produce precise segmentation, pixel-level annotation is costly, and OCT lesions often have ambiguous boundaries with high inter-expert variability. These factors motivate weakly supervised semantic segmentation (WSSS), which uses weaker annotations such as image-level labels~\citep{jiwoon2018, KolesnikovL16, Kwak2017WeaklySS, niu2023fine}, scribbles~\citep{Vernaza_2017, luo2022scribbleseg, valvano2021self}, or bounding boxes~\citep{9684236, oh2021background, DaiH015}. For OCT, image-level WSSS is especially practical because experts can indicate lesion presence more easily than delineate precise boundaries. However, such image-level labels are coarse and lack the granularity required for accurate delineation, often resulting in imprecise outputs. To address these limitations, we draw on clinical practice. Retinal specialists do not rely solely on local pixel intensities. They assess structural context, such as a lesion’s alignment with retinal layers, and summarize findings with high-level descriptions in medical reports. These structural priors and textual reasoning are crucial for distinguishing lesions from artifacts, especially in ambiguous or low-contrast cases. Motivated by this, we introduce structural and semantic information into the WSSS pipeline to emulate clinicians’ diagnostic reasoning. 

\begin{figure}[ht!]
\centering
\subfloat[]{
\includegraphics[trim={0 11cm 0 3cm},clip,width=0.17\linewidth]{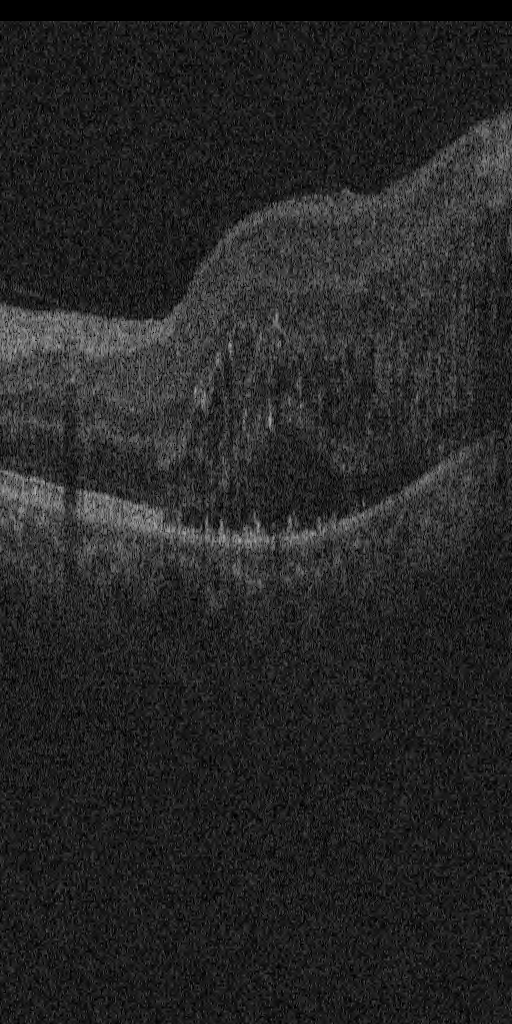}
}
\subfloat[]{
\includegraphics[trim={0 11cm 0 3cm},clip,width=0.17\linewidth]{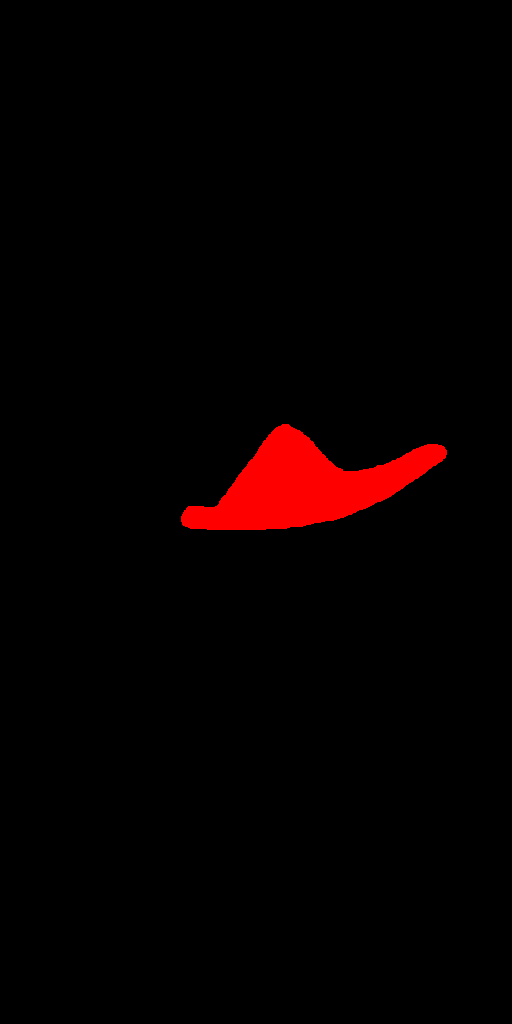}
}
\subfloat[]{
\includegraphics[trim={0 11cm 0 3cm},clip,width=0.17\linewidth]{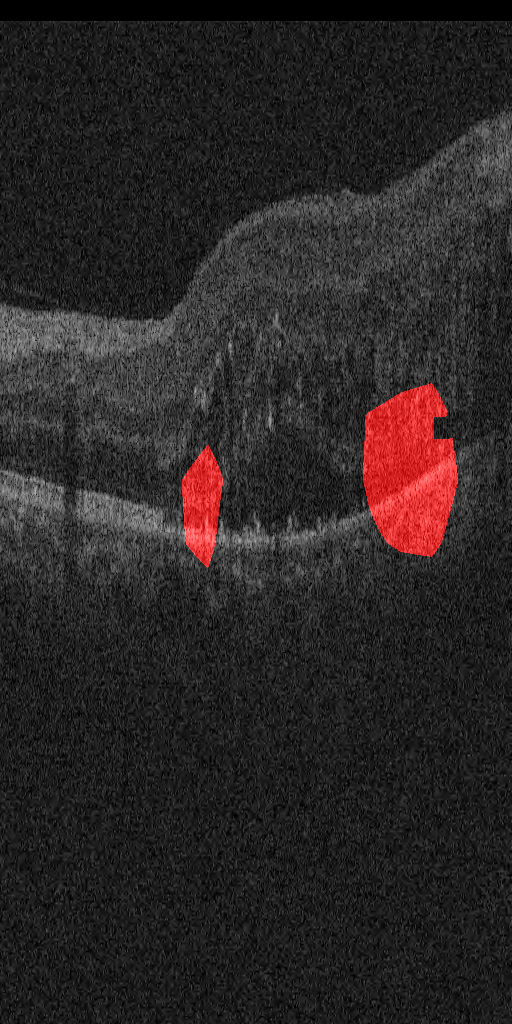}
}
\subfloat[]{
\includegraphics[trim={0 11cm 0 3cm},clip,width=0.17\linewidth]{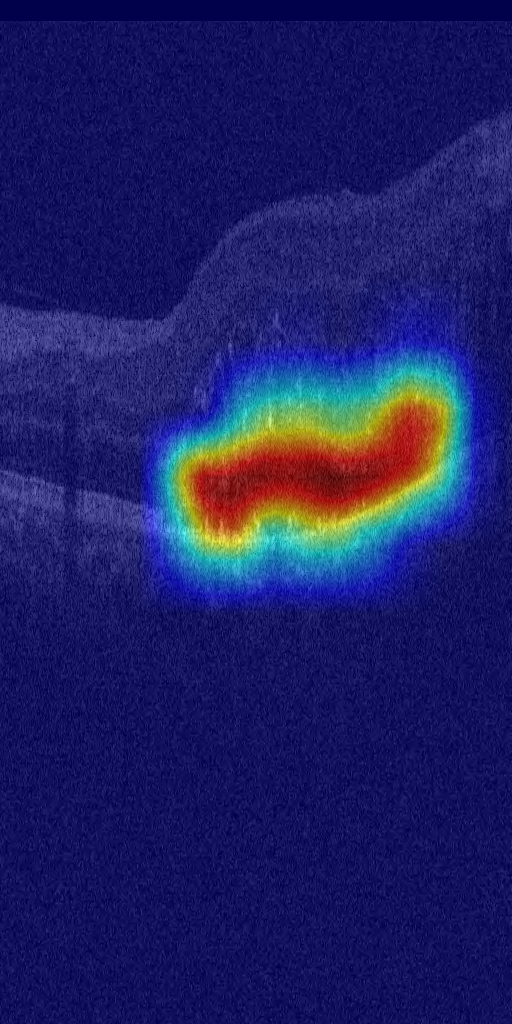}
}
\subfloat[]{
\includegraphics[trim={0 11cm 0 3cm},clip,width=0.17\linewidth]{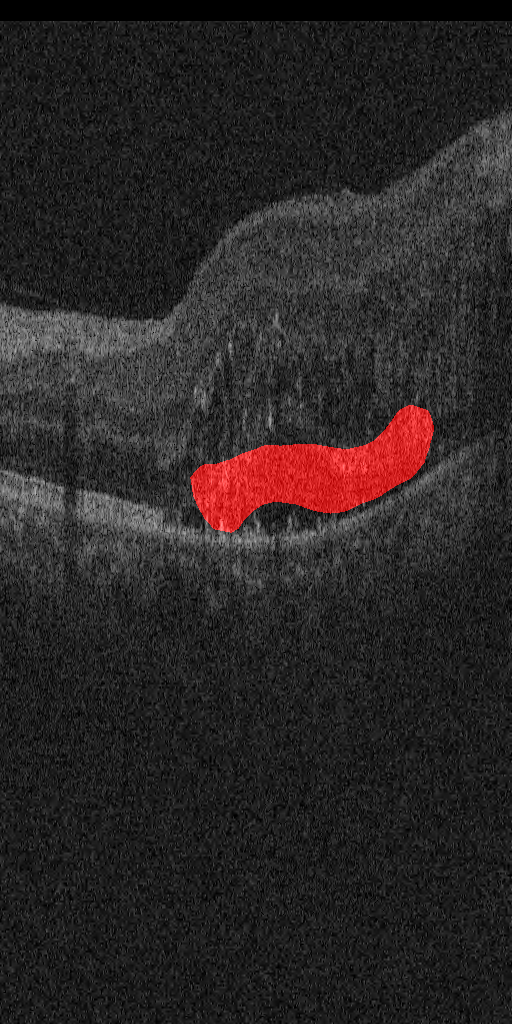}
}
\caption{(a) OCT image; (b) Ground truth (red: SRF lesion); (c) Baseline pseudo label generated by thresholding the CAM of a ResNet-50; (d) SRF lesion CAM from our method; (e) Pseudo label generated by thresholding the CAM of our model.}
\label{fig:teaser_1}
\end{figure}

A common WSSS pipeline trains an \textit{image classifier} with image-level labels to produce \textit{class activation maps (CAMs)}~\citep{ZhouKLOT15}, which are then thresholded into \textit{pixel-level pseudo labels} as the final segmentation, or treated as pseudo ground truth annotation to supervise a \textit{separate segmentation} model. CAM quality is critical because it directly determines pseudo labels and thus segmentation performance. As shown in~\cref{fig:teaser_1}(c), image-level WSSS often yields coarse CAMs. Motivated by clinical practice, we incorporate structural and textual information beyond weak labels. Vision-language models (VLMs) such as CLIP~\citep{clip} align textual descriptions with visual components, offering signals beyond image-only cues. Although WSSS has begun using VLMs~\citep{clip-es, deng2024question}, OCT-specific image–text pairs are scarce, and medical variants like MedCLIP~\citep{wang2022medclip} do not target OCT. We therefore adapt VLMs to OCT using two complementary textual sources: \textbf{label-derived descriptions} that encode category-level semantics and \textbf{domain-agnostic synthetic descriptions} that provide global relational cues. In parallel, we integrate \textbf{structural guidance} based on retinal layers and anomalous patterns to improve spatial specificity. Together, these components strengthen WSSS performance for OCT images. As illustrated in~\cref{fig:teaser_1}, the CAM (d) and corresponding pseudo label (e) generated by our method demonstrate higher quality than the baseline pseudo label (c), which is derived solely from the original image.

\begin{figure}[ht!]
\centering
    \includegraphics[width=0.9\linewidth]{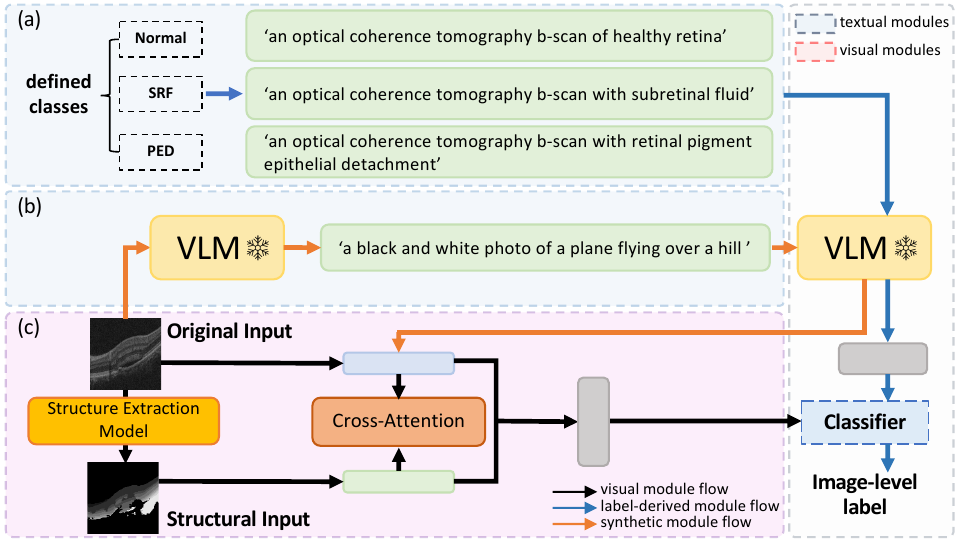}
\caption{High-level visualization of our text and structure-guided classifier, which improves CAMs quality and pseudo labels for downstream segmentation. (a) Label-derived description on RESC dataset. (b) Domain-agnostic synthetic description generated by a VLM, capturing shape and object relationships (e.g., `a plane' over `a hill'). (c) Integration of original and structural inputs via cross-attention mechanism.}
\label{fig:teaser_2}
\end{figure}

\myparagraph{Label-derived description.}
As illustrated in~\cref{fig:teaser_2} (a), we use the image-level lesion labels (e.g., normal, SRF, PED) to define fixed, human-readable textual descriptions for each category (e.g., SRF → “an optical coherence tomography b-scan with subretinal fluid”). These category-level descriptions are applied consistently to all images within the same class. We encode them using the text encoder of a pretrained VLM (CLIP) to obtain local semantic features. These textual features are aligned with visual features through similarity operations to support classification, guiding the model to associate lesions with their corresponding categories. Although CLIP was pretrained on general data rather than OCT images, its broad semantic understanding enables it to infer spatial (e.g., `sub-') and morphological (e.g., `fluid') lesion characteristics even in unfamiliar medical contexts. Aligning CLIP’s text embeddings with OCT visual features guides the model’s attention to relevant regions, refining the representations of corresponding lesions and improving the model's ability to localize them.

\myparagraph{Domain-agnostic synthetic description.}
However, domain-specific label-derived descriptions are limited in quantity. To fully exploit the potential of large foundation models, we propose a complementary approach: using domain-agnostic synthetic description. This method leverages pretrained captioning models from other domains (e.g., natural images) to generate text descriptions for OCT images. These synthetic descriptions take a more global perspective, describing the overall appearance of the image and the relationships between objects. As illustrated by the example in~\cref{fig:teaser_2} (b), the description does not accurately describe the OCT image, instead interpreting it as a scene within the natural image domain. However, as we will show empirically, these seemingly irrelevant descriptions do provide valuable information. 
This approach is akin to a child describing unfamiliar scenes using familiar objects (e.g., toys and pets). While the descriptions may lack medical relevance, they encode consistent spatial information across similar lesions and offer a supplementary signal that surprisingly strengthens the model. 
By integrating synthetic texts with image-level supervision, the model better identifies subtle semantic features and corresponding regions. 

\myparagraph{Structural guidance.}
In addition to text-driven guidance, structural details (i.e., retinal layers) within OCT images provide reliable guidance for lesion localization, drawing inspiration from how doctors assess the positional relationship between objects and retinal layers. Lesions typically align with specific retinal layers, making their positional patterns crucial for distinguishing true lesions from shadows or artifacts. To complement our text-driven strategies, we incorporate structural features into the WSSS framework, leveraging anomalous signals that consistently appear at specific retinal layer positions (\cref{fig:teaser_2} (c)). We introduce cross-attention modules to effectively share structural features across the network and integrate semantic context from textual insights, enhancing weakly supervised segmentation accuracy.

In summary, our proposed method captures valuable signals beyond image labels and transforms them into meaningful representations. Our main contributions are:
(1) We introduce two complementary text-driven strategies to strengthen the model: label-derived descriptions encoding local semantics and domain-agnostic synthetic descriptions for robust lesion representation, which are consistent but medically unrelated texts generated by pretrained natural image captioning models.
(2) We incorporate structural information to guide learning toward retinal layer positions associated with lesion occurrence, improving segmentation precision.
(3) The proposed multi-label WSSS method enriches spatial and contextual guidance for OCT segmentation. To our knowledge, this is the first work to adopt VLMs for WSSS on OCT lesions. Comprehensive experiments on three OCT datasets show that our approach outperforms state-of-the-art image-level WSSS approaches.

\section{Related Works}
\subsection{Weakly Supervised Semantic Segmentation}
The WSSS has gained significant attention due to the scarcity of expert-annotated pixel-level labels, especially in domains such as medical imaging. Most of the methods exploit CAMs to segment class objects~\citep{ZhouKLOT15, gradcam, scorecam, ablationcam, abs-2103-07246, seam, lixiang2022, recam, zoomcam}. WSSS typically follows a three-stage learning process. First, a classification model is supervised using image-level labels to generate initial CAMs, which highlight the discriminative regions that influence the classification decision. C2AM~\citep{xie2022c2am} proposes to apply contrastive learning for foreground and background discrimination to reduce over-activation. Next, these initial CAMs are refined using various techniques, such as dense CRF, affinity-based methods, or saliency maps~\citep{lee2021railroad}, to enhance the quality of the pseudo labels. For instance, methods like~\citep{9577478} embed attention mechanisms to capture class-specific affinities, while approaches such as~\citep{9105077} and~\citep{jo2021puzzle} expand object regions by dropping out the most discriminative parts or removing patches from the images. Finally, the refined pseudo labels serve as ground-truth masks to train a dedicated segmentation network, often using models from the DeepLab~\citep{chen2017deeplab, chen2017rethinking} series, which is then employed for inference, closely mimicking a fully supervised setup.

However, the direct application of these methods to medical images presents significant challenges due to the finer and more complex anatomical structures inherent in medical images. Unlike natural images, where objects are typically larger and more distinct, medical images often feature small, detailed regions, such as blood vessel exudates, with poorly defined boundaries between the foreground objects and the background.

\myparagraph{WSSS for Medical Image.}
In the medical imaging domain, several CAM-based WSSS methods have been developed to refine initial CAMs and improve segmentation across various modalities~\citep{roth2020, Wang:21, patel2022weakly}. However, compared to the computer vision field, research on medical image segmentation using only image-level labels~\citep{ouyang2019weakly, wsmis, belharbi2021deep} remains limited and often relies on domain-specific knowledge. Models from one imaging modality struggle to transfer to another due to differences in image characteristics and anatomical structures. For instance, OEEM~\citep{li2022online} enhances gland segmentation in histology images by using patch-level labels to refine the initial CAM seed, a method less applicable to OCT images with fewer target regions. Similarly, Patel et al.\citep{patel2022weakly} introduce a cross-modality invariant constraint that requires the availability of multiple modalities within the domain. Additionally, most existing methods in the medical image domain, particularly in OCT segmentation, focus on single biomarkers rather than multi-label tasks~\citep{zhang2022transws, tssk, Xing:21, wang2020weakly}, addressing only one lesion class per image. Even though some works~\citep{chen2022c, yang2024anomaly} propose advanced strategies for multi-label tasks in medical imaging, they still limit their exploration to the image modality alone.

Beyond the common challenges mentioned above, medical WSSS rarely addresses the inherent limitations of image-level supervision, which carries limited information. These approaches often overlook the potential of integrating richer information beyond the visual context. In contrast, our approach leverages additional sources of information, such as structural details and multimodal data, including text. This integration enables the model to more accurately localize lesions by utilizing enriched global contextual cues.

\subsection{Vision-Language Models} 
Recent advancements in pretrained large-scale vision-language models have significantly enhanced the integration of visual and textual modalities, leading to improved performance in downstream tasks. One notable model is Contrastive Language-Image Pre-training (CLIP)~\citep{clip}, which consists of an image encoder and a text encoder. CLIP is pretrained on a vast dataset of 400 million image-text pairs, automatically collected from the Internet without manual annotation. This extensive pretraining enables CLIP to effectively map a wide range of visual concepts to their corresponding text labels, demonstrating significant success and potential, particularly in zero-shot settings. To leverage the advantages of CLIP, subsequent works have developed MedCLIP~\footnote{\url{https://github.com/Kaushalya/medclip}}, a model pretrained on the ROCO medical dataset~\citep{pelka2018radiology} using the CLIP framework. Similarly, another version of MedCLIP~\citep{wang2022medclip} has been pretrained on four x-ray datasets, maintaining the core design of CLIP to enhance its performance in the medical domain.
In parallel with these developments, other vision-language models like BLIP~\citep{blip} have emerged, focusing on tasks such as image captioning. BLIP demonstrates strong capabilities in generating natural language descriptions of visual content. It accomplishes this by utilizing a text decoder that interprets the visual input as a question and the text generation task as its corresponding answer.

\myparagraph{CLIP in WSSS.} In the context of WSSS, CLIMS~\citep{xie2022clims} is the first to incorporate CLIP, utilizing the similarity scores generated by the frozen CLIP model as supervision to improve the initial CAMs gradually. Similarly, CLIP-ES~\citep{clip-es} proposes enhancements across all three stages of the WSSS process with a special design for CLIP usage. Both CLIMS and CLIP-ES rely on the zero-shot capabilities of a frozen CLIP model, which is crucial for generating reliable textual features or similarity scores that can effectively guide segmentation in natural images. Since these methods depend on the pretrained CLIP model's outputs, they are not directly adaptable to the OCT image domain, where the visual characteristics differ significantly. Unlike these works in natural images, we leverage CLIP's strong generalization capabilities to adapt it for the retinal OCT domain, aligning image-text similarity through CLIP-generated textual features without relying solely on its zero-shot performance.

In the context of applying CLIP to WSSS in medical imaging, TPRO~\citep{zhang2023tpro} introduces a text-prompting-based WSSS method specifically designed for histopathological images, utilizing MedCLIP embeddings for label texts and ClinicalBert embeddings for knowledge texts.
While TPRO adapts CLIP to medical images using domain-specific embeddings, it applies the same knowledge texts across the dataset, which may miss the nuanced variations between images. In contrast, our approach generates image-specific synthetic captions for each OCT image, enabling more precise differentiation of target objects through consistent semantic descriptions for images within the same lesion category.

\section{Method}

\begin{figure*}[ht!]
\centering
    \includegraphics[width=0.90\linewidth]{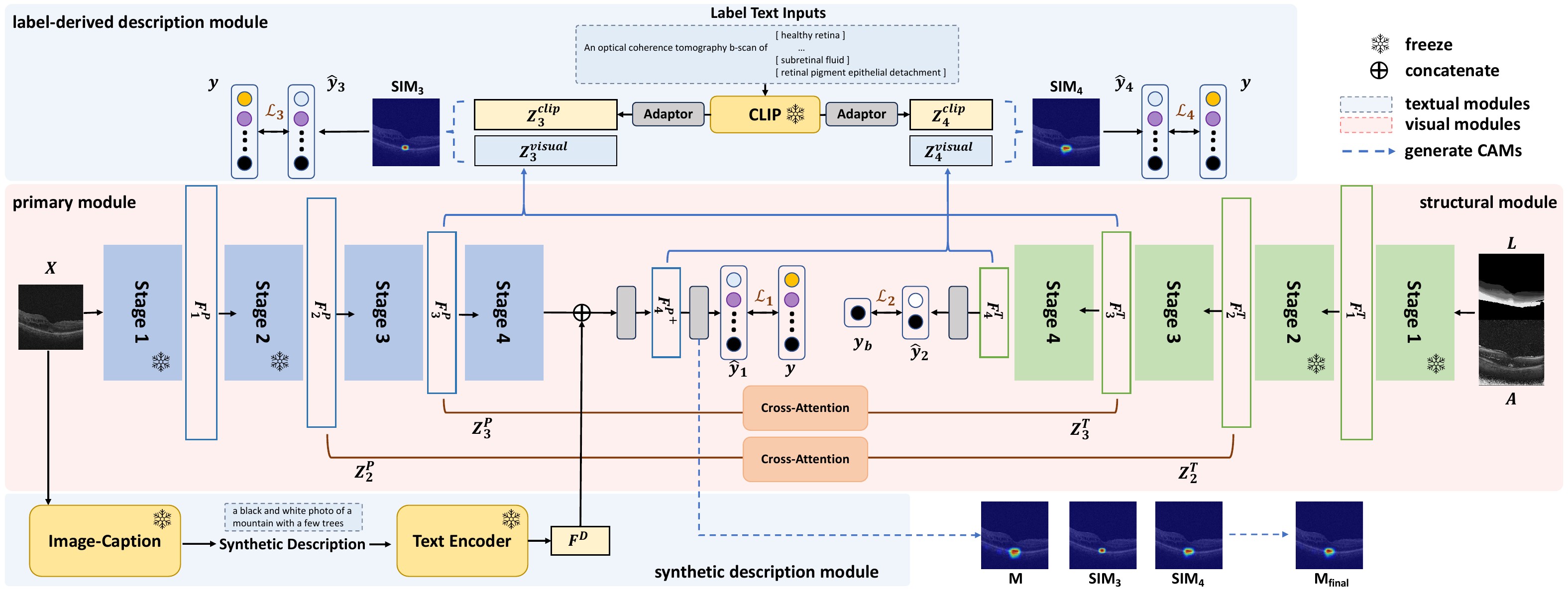}
\caption{Architecture of the proposed approach for pseudo-label generation. A classification network is trained using image-level labels $y$, and the framework integrates two visual modules and two textual modules. The pink blocks indicate feature exchange between the visual modules, where one processes the original OCT image (primary) and the other incorporates layer and anomaly representations (structural). The blue blocks represent textual modules, with one encoding label texts and the other encoding a synthetic description for each image. Aggregated activation maps from the visual modules ($M$) and the textual modules ($SIM_3$ and $SIM_4$) are combined to improve pseudo-label quality.}
\label{fig:arch}
\end{figure*}

As shown in~\cref{fig:arch}, the proposed model is trained using image-level supervision for multi-label classification and then generates high-quality pseudo labels, which are used to train a fully supervised segmentation network in the downstream task.
The architecture comprises two visual processing modules and two textual processing modules. The visual modules process original OCT images with structural information (layer and anomaly inputs). The textual modules use label-derived and domain-agnostic synthetic descriptions via CLIP, fused with visual features to enhance the model's learning process.
The model is trained on a weakly labeled dataset $\mathcal{D}=\{(X_i, y_i)\}_{i=1}^N$, containing healthy and diseased samples. Each input image $X_i \in \mathbb{R}^{C \times H \times W}$ has a corresponding multi-hot image-level label $y_i \in \{0,1\}^K$, where $K$ represents the total number of categories, including $(K-1)$ lesion classes and one non-lesion/background class.

\subsection{Visual Processing Modules}\label{visual_module}
\myparagraph{Primary Feature Module.}
The primary feature module encodes the original OCT images $X_i \in \mathbb{R}^{C \times H \times W}$ to extract relevant features. Specifically, we adopt the Mix Transformer encoder (MiT) as the backbone, known for its efficient multi-scale feature representation, to capture both local and global details of the retinal contexts. The encoder outputs a set of feature maps that encode the essential characteristics of the input images, which are then utilized in subsequent model components. This process is illustrated by the blue flow on the left side of~\cref{fig:arch}.
To preserve the pretrained image context extraction, we freeze the projection layer and the first two stages of the encoder. This approach results in more stable training on our OCT images. The feature set generated by different stages of the primary encoder is denoted as:
$F^P_s \in \mathbb{R}^{H_s \times W_s \times C_s}$
where $s$ indicates the stage, $H_s$ and $W_s$ are the spatial dimensions, and $C_s$ is the feature dimension at stage $s\in[1,2,3,4]$.

\myparagraph{Structural Feature Module.}
The location of a lesion within the retinal structure is crucial, as it directly influences how lesions are identified and differentiated from other abnormalities, such as shadows. However, manually annotating structures like layer segmentation in OCT images is highly time-consuming. To streamline this process and enhance our model's training, we explore pretrained models from existing research~\cite{pnet, li2021mgunet}. It’s important to note that these models, trained on diverse datasets from various devices, may introduce noise and data distribution variations, leading to artifacts that may not align well with our specific datasets.

\begin{figure}[ht!]
\centering
    \includegraphics[width=0.8\linewidth]{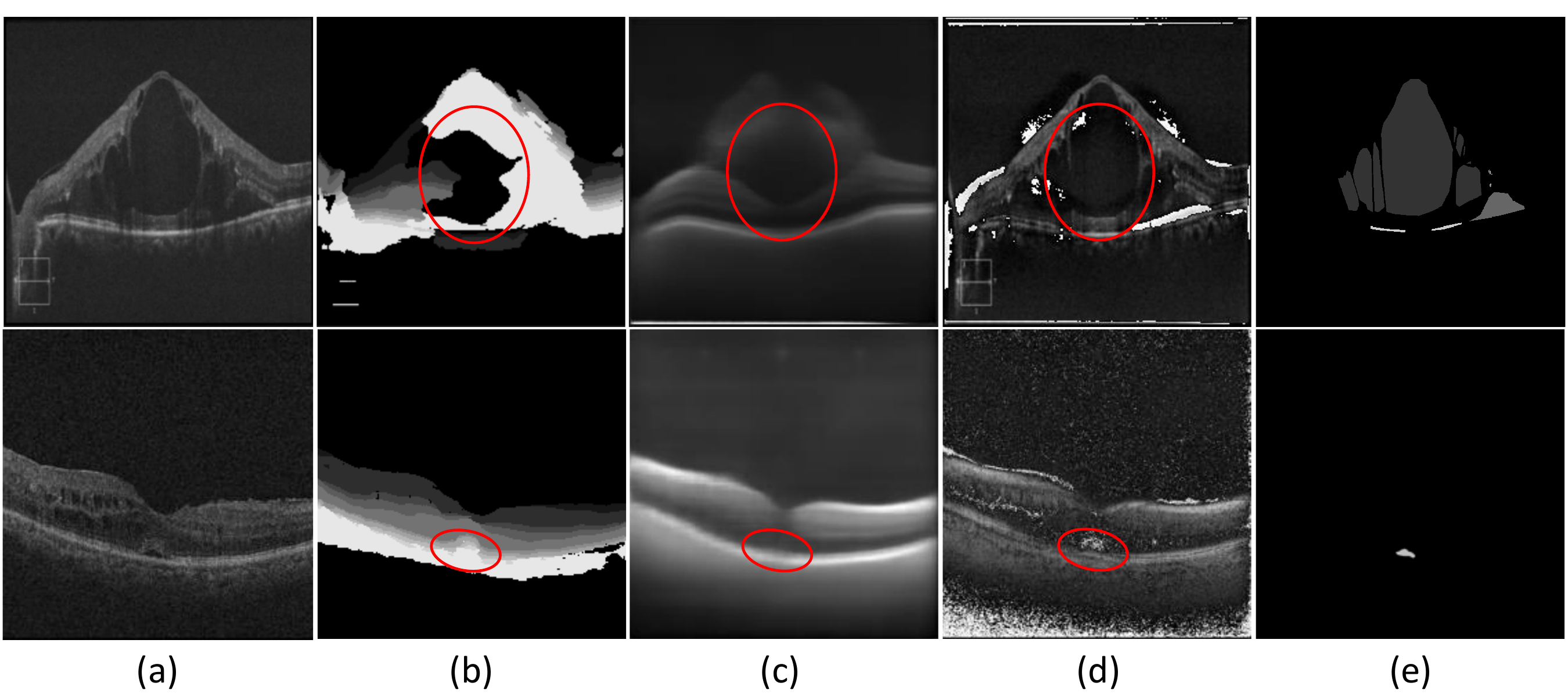}
\caption{Visualization of pretrained models' outputs and their relationship to lesion localization.
(a) Original OCT image;
(b) Layer segmentation;
(c) GAN-generated healthy counterpart;
(d) Anomalous representation;
(e) Ground truth.
Red circles highlight the relationship between lesions and retinal layers, demonstrating how noisy structural information aids lesion localization.}
\label{fig:structure_inp}
\end{figure}

As shown by the green flow on the right side of visual modules in~\cref{fig:arch}, we adopt the pretrained model~\cite{pnet} for layer segmentation and enhance it with anomaly-discriminative representations, derived from the absolute difference between the original input $\mathcal{D}$ and its healthy counterpart generated by GANomaly, from~\cite{yang2024anomaly}. These inputs are denoted as $\mathcal{L}={(L_i, y_i)}^N_{i=1}$ and $\mathcal{A}={(A_i, y_i)}^N_{i=1}$ respectively. Although pretrained models introduce noise and artifacts, they still offer valuable guidance by capturing the positional relationships between lesions and retinal layers and identifying anomalous regions. 
As illustrated in~\cref{fig:structure_inp} (b, d), their outputs are imperfect yet effectively highlight where lesions occur in relation to specific retinal layers, providing critical insights for accurate lesion localization.

Similar to the primary feature module, we employ the MiT encoder as the backbone. To enable effective feature extraction, we input the sum of the layer and anomaly images. Thus, the input of the structural encoder is designed as:
\begin{align}
    &X'_i = \frac{(L_i + A_i)  - (L_i + A_i)_{\text{min}}}{(L_i + A_i)_{\text{max}} - (L_i + A_i)_{\text{min}}} \\
    &\mathcal{D}'=\{(X'_i, y_i)\}_{i=1}^N, \text{ where } X'_i \in \mathbb{R}^{C \times H \times W}
\end{align}
where $X'_i$ denotes the input to the structural module and $\mathcal{D}'$ represents the dataset. The final output of the structural module is a set of feature maps, denoted as $F^T_s \in \mathbb{R}^{H_s \times W_s \times C_s}$. These feature maps are then integrated with those from the primary feature module to provide richer structural information.

\myparagraph{Cross-Attention Feature Exchange.}
We introduce a cross-attention module to facilitate the exchange of information between the primary and structural modules. This mechanism enables feature interaction across visual encoders before progressing to the next encoding stage.

The structure of this module is based on the core components of the transformer block, incorporating layer normalization, multi-head attention, multi-layer perceptron (MLP), and skip connections. We utilize features from both modules to create the \textit{Affinity} matrix, which then enhances the representation in the target module. Taking the primary module as an example: 
\begin{equation}
    \begin{aligned}
        \mathsf{Q} = \mathbf{W}^q_sF^T_s, \; \mathsf{K} = \mathbf{W}^k_sF^P_s, \;  \mathsf{V} = \mathbf{W}^v_sF^P_s \\
        \text{Affinity}^P_s = \text{softmax}(\mathsf{QK}^\intercal / \sqrt{C_s/h}) \\
        Z^P_s =  \text{Affinity}^P_s \mathsf{V}, \; s \in [2,3]
    \end{aligned}
\end{equation}
where $\mathbf{W}^q_s, \mathbf{W}^k_s, \mathbf{W}^v_s$ are learnable parameters at encoder stage $s$. $h$ is the number of heads, and $Z^P_s \in \mathbb{R}^{H_s \times W_s \times C_s}$ represents the features after the interaction between $F^T_s$ and $F^P_s$ at stage $s$. 
Similarly, for the structural branch at stage $s$, we obtain the cross-attention enhanced feature $Z^T_s$. In this design, we ensure comprehensive feature exchange between both modules, effectively enhancing the primary features with structural information that carries positional and weak anomaly signals. This enriched contextual information improves the guidance for subsequent feature processing during training.

\subsection{Textual Processing Modules}\label{text_module}
In this section, we introduce the textual modules that leverage large VLMs for medical imaging, transforming `imperfections' into valuable signals for the WSSS task. 

\myparagraph{Synthetic Description Module.} For each OCT image $X_i$, we use BLIP~\cite{blip} which is trained on natural images to generate a domain-agnostic description. The generated description is encoded into a textual feature $F^D \in \mathbb{R}^{1 \times C^{desp}}$ which is then spatially broadcasted to $F^D \in \mathbb{R}^{H_s \times W_s \times C^{desp}}$ so that its resolution aligns with the visual feature map. As illustrated in the bottom left of the textual modules in~\cref{fig:arch}, the visual and textual features are then concatenated and passed through learnable layers for joint optimization:
\begin{equation}
    F^{P\textsuperscript{+}}_s = (F^P_s \oplus F^D)\mathbf{W}_s^{fuse}
\end{equation}
To avoid disrupting the cross-attention module and to stabilize the training process, we fuse the synthetic descriptive feature with the visual feature only at the last stage, $s=4$, of the primary module. In this integration, the model leverages the semantic consistency of similar lesions to enhance the visual features in regions relevant to the descriptions.

\myparagraph{Label-Derived Description Module.}
We utilize the pretrained CLIP model~\cite{clip} to encode the predefined text labels for each OCT dataset, as the example shown in~\cref{fig:teaser_2} (a). The encoded features are denoted as $F^{clip} \in \mathbb{R}^{K \times C^{clip}}$, where $K$ represents the number of categories, including \textit{background}, and $C^{clip}$ is the dimension of the encoded features. The CLIP-encoded features are used to create similarity maps with the features generated by the visual modules. To achieve this, we need an adaptor to learn the characteristics of each lesion and align the vector dimensions.

Firstly, we generate the unified visual feature at stage $s$ by simply adding $Z^{visual}_s = F^P_s + F^T_s $, and reshaping it to $Z^{visual}_s\in \mathbb{R}^{H_sW_s \times C_s}$. Note that the $F^P_4$ is the $F^{P\textsuperscript{+}}_4$, which is the enhanced version by synthetic text features at stage 4. Then, we introduce MLP layers to transform the embeddings in the label-derived text space: 
\begin{equation}
    Z^{clip}_s = ReLU(F^{clip} \mathbf{W}^1_s)\mathbf{W}^2_s
\end{equation}
where $\mathbf{W}$ represents the learnable parameters, and $Z^{clip}_s \in \mathbb{R}^{K \times C_s}$. Next, we transpose the transformed CLIP feature into $Z^{clip}_s \in \mathbb{R}^{C_s \times K}$ to align the feature dimensions for subsequent matrix operations and facilitate efficient fusion with visual feature maps. Now, we can generate the similarity map by:
\begin{equation}
    \text{SIM}_s = r_s \cdot (Z^{visual}_sZ^{clip}_s), \; s \in [3,4]
\end{equation}
where $r_s$ is the learnable scaling factor at stage $s$. The vector $\text{SIM}_s \in \mathbb{R}^{H_sW_s \times K} \mapsto \mathbb{R}^{K \times H_s \times W_s} $ represents the similarity score between the visual features and each of the label-derived text embedding.

\subsection{Learning Objective and Pseudo Label Generation}
\myparagraph{Weakly Supervised Objectives.}
To train the model with weak supervision and facilitate the learning of correlations between textual and visual features, we design a loss function with three aspects: image-level multi-label classification, structural binary classification, and label-derived multi-label classification.

We employ the global max pooling (GMP) layer at the end of the primary module, followed by a convolutional layer $\mathbf{W} \in \mathbb{R}^{C_4 \times K}$, to generate the class prediction, $\hat{y}_1 = \text{GMP}(F^{P\textsuperscript{+}}_4) \mathbf{W}$, and then we use the multi-label binary cross-entropy loss for training as follows: 
\begin{equation}
    \mathcal{L}_1 = -\frac{1}{K}\sum_{k=1}^K y^k\log \sigma (\hat{y}_1^k) + (1-y^k)\log [1-\sigma(\hat{y}_1^k)]
\end{equation}
where $\sigma$ is the sigmoid activation function.

For the structural module, which is responsible for capturing features indicating structural disruption and supporting the primary branch, we calculate the binary classification loss to distinguish between the healthy and non-healthy. For example, a multi-label target $y$ indicating different lesions is converted to a binary label $y_b$ (0 for healthy, 1 for lesion present). We then apply GMP followed by a convolutional layer, $\mathbf{W} \in \mathbb{R}^{C_4 \times 2}$, at the end of the structural module to obtain the binary prediction $\hat{y}_2 = \text{GMP}(F^{T}_4)\mathbf{W}$. The loss function is defined as follows:
\begin{equation}
    \mathcal{L}_2 = -\sum_{k=1}^2 y^k_b\log(\hat{y}_2^k)
\end{equation}

Moreover, to further enhance the model with textual guidance, we apply GMP on the obtained similarity maps to generate multi-label predictions as well. Since the SIM maps are derived from the last two stages ($s=3,4$), the final label-derived class predictions are obtained as $\hat{y}_{3} = \text{GMP}(\text{SIM}_3)$ and $\hat{y}_{4} = \text{GMP}(\text{SIM}_4)$. Similar to $\mathcal{L}_1(\hat{y}_1, y)$, we calculate the binary cross-entropy loss between the labels $y$ and the predictions $\hat{y}_3, \hat{y}_4$ for training, denoted as $\mathcal{L}_3(\hat{y}_3, y)$ and $\mathcal{L}_4(\hat{y}_4, y)$.

Finally, we compute the weighted sum of the losses from different components of the network to define the final loss of our proposed model as follows:
\begin{equation}
    \mathcal{L} = \lambda_1\mathcal{L}_1 + \lambda_2\mathcal{L}_2 + \lambda_3\mathcal{L}_3 + \lambda_4\mathcal{L}_4\label{loss_func}
\end{equation}

\myparagraph{Pseudo Label Generation.}
Once the model training is completed, we generate CAMs from different modules within the network. First, lesion CAMs are obtained from the primary module using the learned features from the encoder's last layer, denoted as $\mathbf{M} \in \mathbb{R}^{(K-1) \times H_4 \times W_4}$. Additionally, similarity maps between visual and textual features from the last two stages are directly used as additional heatmaps, retaining only the lesion channels, denoted as $\text{SIM}_s \in \mathbb{R}^{(K-1) \times H_s \times W_s}$.

We apply the $ReLU$ activation to the heatmaps, perform a weighted sum to emphasize localization, and then normalize them. 
Note that these maps need to be resized to the same dimensions before fusion.
\begin{equation}
    \begin{aligned}
    &\mathbf{M}_{fg} = \gamma_1ReLU(\mathbf{M}) + \gamma_2ReLU(\text{SIM}_3) + \gamma_3ReLU(\text{SIM}_4) \\
    &\mathbf{M}_{fg}^k = \frac{\mathbf{M}_{fg}^k-\text{min}(\mathbf{M}_{fg}^k)}{\text{max}(\mathbf{M}_{fg}^k) - \text{min}(\mathbf{M}_{fg}^k)}
    \end{aligned}\label{heatmaps}
\end{equation}
where $k$ represents the class.
Then, following common practice, we add a background map, $\mathbf{M}_{bg}$, by applying a threshold $\lambda$ in the range $[0, 1]$. The final localization map is $\mathbf{M}_{final}=\mathbf{M}_{bg} \oplus \mathbf{M}_{fg}$. Finally, we obtain the pseudo labels for segmentation by applying \textit{argmax} across the classes.

\section{Dataset and Experimental Settings}
\subsection{Datasets and Evaluation Metrics}
Following the AGM study~\citep{yang2024anomaly}, we evaluate our proposed method on the same three OCT B-scan datasets: RESC, Duke, and a private dataset. We employ standard evaluation metrics to assess our method's performance, focusing on pixel-level prediction through micro-averaging mean Intersection over Union (mIoU).

\subsection{Baselines and Evaluation Metrics}
To evaluate the performance of our proposed method, we compare the quality of pseudo labels with existing WSSS methods that utilize image-level supervision. These methods include IRNet~\citep{irnet}, ReCAM~\citep{recam}, SEAM~\citep{seam}, WSMIS~\citep{wsmis}, MSCAM~\citep{mscam}, TransWS~\citep{zhang2022transws}, DFP~\citep{wang2020weakly}, TPRO~\citep{zhang2023tpro}, and AGM~\citep{yang2024anomaly}. In our experiments, we consider both the base versions, for SEAM and ReCAM, and their corresponding extensions, SEAM\textsuperscript{+} and ReCAM\textsuperscript{+} (see~\Cref{tab:main_metrics}).
Please note that, for each method, we traverse the background threshold $\lambda \in [0,1]$ in increments of 0.01 using the same strategy to report the best mIoU of pseudo labels on each dataset.

In addition to assessing the final pseudo labels' performance, we report segmentation results from networks trained on these pseudo labels in a fully supervised manner (see~\Cref{tab:seg}).

\subsection{Implementation Details}
The input images are resized to \(512 \times 512\) pixels. For visual processing, we use the pretrained MiT-b2 encoder with the first two stages frozen. The label-derived branch is encoded with CLIP-large, while the synthetic description branch generates text with BLIP and encodes it with CLIP-base. For all baselines and our proposed model, the threshold \(\lambda\) for final pseudo labels is selected by traversing values from 0 to 1, choosing the value that yields the best mIoU across classes on the validation set. We train the model using a learning rate of \(1 \times 10^{-4}\) with a batch size of 8 for 30 epochs, applying binary cross-entropy loss and cross-entropy loss with the Adam optimizer. Data augmentation includes random horizontal flipping, rotation, and color jittering. For semantic segmentation, we employ DeepLabV3+~\citep{chen2018encoder} with a ResNet-101 backbone and Segformer with an MiT-b5 backbone, both pretrained on ImageNet~\citep{deng2009imagenet}. All experiments are implemented with PyTorch on a single RTX A6000 GPU. The code and pretrained models are publicly available at \url{https://github.com/yangjiaqidig/WSSS-AGM}.

\section{Results}
\subsection{Comparison on Pseudo Label} 
\begin{table*}[!ht]
  \centering
  \fontsize{7}{13}\selectfont
  \caption{Performance of pseudo labels on different datasets using our proposed method compared to previous methods. The table presents \textbf{mIoU} scores for each method. Rows indicate the performance of each method, while columns list the datasets used for evaluation along with their corresponding lesion classes, including background (bg). The best results are highlighted in bold.}
  \begin{tabular}{ll|cccc|ccc|cccccc}
\hline
Dataset &  & \multicolumn{4}{c|}{RESC} & \multicolumn{3}{c|}{Duke} & \multicolumn{6}{c}{Our Dataset}\\
\hline
Method & Backbone & bg & SRF & PED & \textbf{mIoU} & bg & Fluid & \textbf{mIoU} & bg & SRF & IRF & EZ & HRD & \textbf{mIoU} \\
\hline

\multicolumn{5}{l}{\textbf{Image-level supervision only - natural image.}} & &&&&& \\
IRNet                   & R50   & 97.78 & 33.75 & 14.66 & 48.73   & 98.10 & 20.45 & 59.27   & 98.18 & 0.01  & 11.29 & 0.07 & 1.17 & 22.14 \\
SEAM                    & WR38  & 97.43 & 34.13 & 10.71 & 47.42   & 97.03 & 17.87 & 57.45   & 97.87 & 13.52 & 14.45 & 1.61 & 2.25 & 25.94 \\
SEAM\textsuperscript{+} & WR38  & 97.65 & \textbf{47.29} & 3.49  & 49.48   & 96.75 & 18.64 & 57.69   & 97.95 & 13.30 & 14.68 & 1.71 & 2.29 & 25.99 \\  
ReCAM                   & R101  & 97.66 & 14.23 & 19.11 & 43.67   & 96.41 & 11.67 & 54.04   & 97.13 & 0.10  & 9.42  & 1.02 & 1.54 & 21.84 \\ 
ReCAM\textsuperscript{+}& R101  & 97.96 & 12.71 & 36.99 & 49.22   & 96.87 & 13.86 & 55.37   & 97.47 & 1.01  & 9.21  & 0.68 & 0.72 & 21.82 \\ 
\hline
\multicolumn{5}{l}{\textbf{Image-level supervision only - medical image.}} & &&&&&\\
WSMIS                   & R101  & 95.64 & 24.64 & 2.96  & 41.08   & 96.41 & 0.42  & 48.41   & \textbf{98.57} & 26.71 & 8.93  & 0.08 & 0.17 & 26.89 \\
MSCAM                   & R101  & 97.25 & 10.14 & 11.97 & 39.79   & 98.00 & 17.98 & 57.99   & 97.36 & 22.50 & 1.47  & \textbf{3.25} & 0.14 & 24.94\\
TransWS                 & MiT     & 98.18 & 34.88 & 17.22 & 50.09   & 98.15 & 27.01 & 62.58   & 97.52 & 9.41  & 11.21 & 0.11 & 1.96 & 24.04\\
DFP                     & R101  & 97.72 & 6.40  & 15.64 & 39.92   & 98.24 & 15.14 & 56.69   & 98.31 & 18.05 & 10.06 & 0.67 & \textbf{5.72} & 26.56\\
AGM                     & R50   & 98.34 & 43.94 & 22.33 & 54.87   & 98.29 & 30.06 & 64.17   & 97.09 & 31.76 & 11.17 & 1.74 & 0.57 & 28.46\\
\hline    
\multicolumn{5}{l}{\textbf{Image-level supervision + text - multimodal.}} & &&&&&\\
TPRO                    & MiT   & 97.10 & 30.41 & 8.27 &  45.26   & 97.72 & 27.80 & 62.76 & 95.80 & 11.03 & 13.37 & 0.18 & 0.85 & 24.25 \\
\rowcolor{gray!10}\textbf{Ours}              & MiT   & \textbf{98.60} & 43.50 & \textbf{41.35} & \textbf{61.15}   & \textbf{98.47} & \textbf{37.74} & \textbf{68.11} & 98.40  & \textbf{33.46} & \textbf{20.40} & 1.15 & 1.84 & \textbf{31.05}\\ 
\hline
\end{tabular}

  \label{tab:main_metrics}
\end{table*}

Since our model is designed to generate high-quality CAMs for better pseudo labels, we assess pseudo label performance in this section. The results in~\Cref{tab:main_metrics} compare the pseudo labels generated by our method and the baselines against the pixel-level ground truth on the validation set. Due to the limited availability of pixel-level annotations, we use the validation set as the final evaluation metric. To ensure a fair comparison, we report the highest scores achieved by each baseline and our method on the validation set.
Each dataset includes the background class (\textit{bg}) and various lesion classes: Subretinal Fluid (\textit{SRF}) and Pigment Epithelial Detachment (\textit{PED}) for RESC; \textit{Fluid} for Duke; and \textit{SRF}, Intraretinal Fluid (\textit{IRF}), Ellipsoid Zone disruption (\textit{EZ}), and Hyperreflective Dots (\textit{HRD}) for our dataset. The mIoU is calculated across all classes within each dataset. To facilitate analysis, the methods are organized into three categories based on the data domain and modality: (1) image-level supervision designed for natural images, (2) image-level supervision designed for medical images, and (3) image-level supervision combined with textual information (multimodal).

Our proposed method demonstrates superior performance across three datasets, achieving higher mIoU scores compared to other WSSS methods. However, as noted in AGM~\citep{yang2024anomaly}, applying the same threshold across different lesions' CAMs leads to an inherent trade-off, so high overall mIoU may not guarantee optimal performance for each lesion class.
On the RESC dataset, our method significantly outperforms the second-best method, AGM, achieving overall mIoU of \textbf{61.15\%} compared to 54.87\%. Specifically, for segmentation in the \textit{PED} lesion, our method attains an IoU of \textbf{41.35\%}, nearly double AGM’s 22.33\%. This substantial improvement indicates our method’s effectiveness in accurately relatively challenging lesions. Although AGM performs slightly better on the \textit{SRF} lesion (43.94\% vs. 43.50\%), our method demonstrates consistently strong and balanced performance across both lesion classes, resulting in a higher overall performance.

In the Duke dataset, which contains only the \textit{Fluid} lesion class, our method achieves a mIoU of \textbf{68.11\%}, outperforming AGM by 3.94\% and TPRO by 5.35\%. Specifically, our method attains an IoU of 37.74\% for the \textit{Fluid} class, surpassing AGM's 30.06\% by 7.68\%. 

Our private dataset presents the most challenging scenario due to the presence of four different lesion types, some with subtle features and sparse pixel representation. Despite these challenges, our method achieves the highest mIoU of \textbf{31.05\%}, outperforming AGM by 2.59\%. It excels in segmenting the \textit{SRF} and \textit{IRF} classes, achieving IoUs of 33.46\% and 20.40\%, respectively, surpassing AGM by 1.70\% and 9.23\%. Other methods like DFP and MSCAM achieve higher IoUs in certain classes, such as \textit{HRD} and \textit{EZ}, but they do so by sacrificing accuracy in other lesion types largely. For instance, MSCAM attains a higher IoU in \textit{EZ} (3.25\%), but its performance in \textit{IRF} drops significantly to only 1.47\%, whereas our method achieves 20.40\% in \textit{IRF}.

These findings indicate that our method not only achieves high overall mIoU scores but also maintains balanced performance across different lesion classes. Interestingly, TPRO, which also utilizes textual information, sometimes performs worse than methods relying solely on image data. One possible explanation is that TPRO generates CAMs using features from earlier network stages, which are less specific and struggle to capture the relatively small target regions in OCT images, unlike the larger regions in the histopathology images used by TPRO. Besides, while both TPRO and our method utilize textual information, TPRO uses only label text and shared knowledge text across the entire dataset, which may not provide sufficient independent, image-specific information. In contrast, our method leverages per-image independent text, allowing for a more tailored and detailed semantic understanding of each image, as detailed in~\cref{ablation_study_sec}.

\subsection{Comparison on Semantic Segmentation Results}
Pseudo labels generated with a WSSS model can be used as ground truth to train a separate semantic segmentation model. To provide a comprehensive comparison of the segmentation task, we select SEAM\textsuperscript{+}, AGM, and TPRO from different paradigms (see~\Cref{tab:main_metrics}) and compare their mIoU performance with our proposed method, all trained using their respective pseudo labels. The results are presented on the RESC and Duke datasets using DeepLabV3+ (ResNet-50/101) and Segformer architectures as shown in~\Cref{tab:seg}. 
The \textit{Upper Bound} represents the performance achieved when using pixel-level ground truth for training. Since the Duke dataset lacks pixel-level annotations, its \textit{Upper Bound} is marked as ‘-’. Notably, Segformer (b5) achieves the highest upper-bound performance at 77.1\%.

\begin{table}[ht]
  \centering
  \fontsize{8}{10}\selectfont
  \caption{Comparison of segmentation results (mIoU) between our proposed method, SEAM\textsuperscript{+} from the computer vision domain, AGM from the OCT domain, and TPRO from a CLIP-based strategy, on the RESC and Duke datasets. The comparison uses DeepLabV3+ (ResNet-50/101) and Segformer. The best results are in \textbf{bold}. For reference, the \textit{Upper Bound} represents a fully supervised segmentation method using pixel-level ground truth with various models.}
  \begin{tabular}{llcc}
    \toprule
    & Segmentation Model & RESC & Duke \\
    \midrule
    \textit{Upper Bound} & U-Net & 67.75 & - \\ 
    & DeepLabV3+ (ResNet-101) & \textit{71.65} & - \\ 
    & Segformer (b2) & 71.75 & - \\ 
    & Segformer (b5) & \textbf{77.10} & - \\ 
    
    \midrule
    SEAM\textsuperscript{+} & DeepLabV3+ (ResNet-50) & 49.24 & 58.28\\
    & DeepLabV3+ (ResNet-101) & 49.62 & 59.50\\
    AGM & DeepLabV3+ (ResNet-50) & 52.11 & 66.26\\  
    & DeepLabV3+ (ResNet-101) & 53.87  & 66.42\\
    TPRO & DeepLabV3+ (ResNet-101) & 47.86 & 64.23\\  
    & Segformer (b5) & 47.76  & 64.90\\
    \rowcolor{gray!10} \textbf{Ours} & DeepLabV3+ (ResNet-101) &59.68& \textbf{71.07}  \\
    \rowcolor{gray!10}& Segformer (b5) & \textbf{60.52} & 70.65\\
    \bottomrule
\end{tabular}
\label{tab:seg}

\end{table}

Compared to other WSSS approaches, our method consistently demonstrates superior performance on both datasets, largely due to the higher quality pseudo labels generated by our model (see~\Cref{tab:main_metrics}), leading to a more accurate segmentation model. Since DeepLabV3+ with ResNet-101 has shown stronger performance than ResNet-50, and Segformer (b5) performed exceptionally well in the \textit{Upper Bound} evaluation, we report results using ResNet-101 and Segformer (b5) for both TPRO and ours. On the RESC dataset, our method achieves 60.52\% mIoU with Segformer (b5) and 71.07\% mIoU with DeepLabV3+ (ResNet-101) on the Duke dataset, outperforming the other models by at least 6\% and 4\%, respectively. Interestingly, Segformer (b5), when trained on generated pseudo labels, does not significantly outperform DeepLabV3+ as observed in the \textit{Upper Bound} and even performs worse on the Duke dataset. This indicates that the generated pseudo labels still lack the quality of true ground truth, particularly in capturing finer, more precise details.

\subsection{Ablation Studies} \label{ablation_study_sec} 
In this section, we comprehensively evaluate our proposed model. All experiments are run five times to ensure reliability. We report either the highest mIoU or the full statistics (mean, std, highest, and lowest), depending on the table, as specified.

\subsubsection{Effects of Individual Modules}
In~\Cref{tab:ablation_submodule}, we assess the contribution of a textual or structural module when combined with the base module alone on two public datasets, focusing on the performance of pseudo labels measured by overall mIoU scores across lesions. The \textit{base} denotes the primary feature module that processes the original image as input, while the \textit{layer} and \textit{anomaly} refer to the inputs for the structural module. Additionally, the \textit{clip} and \textit{caption} represent the two textual modules, providing label-derived guidance and synthetic descriptions, respectively. Note that the \textit{clip} used is CLIP-large, and the \textit{caption} is generated by BLIP and encoded into vectors by CLIP-base.

\begin{table}[ht]
\centering
\fontsize{8}{10}\selectfont
\caption{Ablation study for our proposed method on public datasets (mIoU). Roman numerals in the first column correspond to different combinations of these components.}
  \begin{tabular}{lcccccccc}
    \toprule
    & base & layer & anomaly & clip & caption & RESC & Duke\\
    \midrule
    \RN{1} & \checkmark  & & &  & &   52.09 & 64.57  \\
    \RN{2} & \checkmark & \checkmark &  &  & &   55.38 & 65.66 \\ 
    \RN{3} & \checkmark &  & \checkmark &  & &  55.24 & 65.91 \\ 
    \RN{4} & \checkmark &  &  & \checkmark & &  54.18 & 66.88 \\ 
    \RN{5} & \checkmark &  &  &  & \checkmark &  51.58 & 66.28 \\ 
    \RN{6} & \checkmark & \checkmark & \checkmark & \checkmark & &  55.26 & 65.55    \\
    \RN{7} & \checkmark & \checkmark &  & \checkmark & \checkmark & 57.68 & 66.62   \\
    \RN{8} & \checkmark & &  \checkmark & \checkmark & \checkmark & 58.70 & 65.15   \\
    \rowcolor{gray!10}\RN{9} & \checkmark & \checkmark& \checkmark & \checkmark& \checkmark  &  \textbf{61.15} & \textbf{68.11}  \\ 
    \bottomrule
  \end{tabular}
  \label{tab:ablation_submodule}

\end{table}

As shown in~\Cref{tab:ablation_submodule}, performance generally improves as more sub-modules are added, peaking with the full architecture \RN{9} (61.15\% on RESC and 68.11\% on Duke). However, we observe some interesting performance drops when adding certain sub-modules, such as \RN{6} vs \RN{2} on both datasets, \RN{5} vs \RN{1} on RESC, and \RN{8} on Duke. These inconsistencies can be attributed to: (1) The full model (\RN{9}) benefits from a synergistic effect, where all sub-modules complement each other, resulting in more stable and higher performance. In contrast, partial combinations may disrupt this balance and fail to fully leverage the model's potential. (2) Since the table reports the highest values, simpler combinations may show greater performance variance, occasionally appearing better, even though configurations with more modules tend to be more consistent overall. We will explore this variance in the following section.

\subsubsection{CAM Variation Analysis}
As mentioned above, while we focus on the highest values for comparison, the inherent variability in CAM generation can lead to fluctuations in stability across runs. This section examines the variations to ensure that peak scores are not simply due to chance or “lucky outcomes”. In~\Cref{tab:ablation_variation}, we present the mean, standard deviation, and the highest and lowest performance across five runs for each experiment, referencing the configurations in~\Cref{tab:ablation_submodule} to provide a clearer view of stability on the RESC dataset. Notably, partial configurations, such as \RN{2}, show lower stability, with a minimum performance of 32.61\%, in contrast to \RN{6}, which achieves comparable top performance but exhibits less variation overall. The complete architecture (\RN{9}) consistently yields the best performance with relatively low variation, even in its lowest-scoring run.
\begin{table}[ht]
\caption{Comparison on RESC showing stability across sub-module combinations. The largest variation is in \color{red}red.}
  \centering
  \fontsize{8}{10}\selectfont
  \begin{tabular}{lcccc}
    \toprule
     & Mean (\%) & Std & High (\%) & Low (\%) \\
    \midrule
    \RN{1}  & 49.75 & $\pm$1.72 & 52.09 & 47.20 \\ 
    \RN{2}  & 46.91 & \color{red}$\pm$8.74 & 55.38 & \color{red}32.61\\
    \RN{5}  & 49.90 & $\pm$1.07 & 51.58 & 48.34\\
    \RN{6}  & 52.10 & $\pm$2.24 & 55.26 & 49.56\\ 
    \RN{8}  & 53.53 & $\pm$2.80 & 58.70 & 51.01\\ 
    \rowcolor{gray!10}\RN{9}  & \textbf{56.87} & $\pm$2.54 & \textbf{61.15} & \textbf{54.33}\\ 
    \bottomrule
  \end{tabular}
  \label{tab:ablation_variation}
\end{table}

\subsubsection{Label-Derived Features} 
\Cref{tab:ablation_clip} presents the performance of pseudo labels on the RESC dataset when using different label text encodings: random vectors, MedCLIP~\footnote{\url{https://github.com/Kaushalya/medclip}}, CLIP-base, and CLIP-large. For this study, we used a simplified model structure consisting of only the primary module with the CLIP label-derived module to isolate and examine the effect of each encoding method. The \textit{Random} serves as a baseline, allowing us to assess how much each specific label encoding improves performance over a non-informative input.
The results indicate that CLIP encodings outperform both MedCLIP and Random baselines. Interestingly, MedCLIP does not surpass Random, with a lower mean mIoU (47.46\%) and higher variability. This suggests that MedCLIP, fine-tuned on radiology images rather than a large variety of medical images, may lack the foundational text-image relationships needed for effective label alignment with OCT images. This finding highlights the value of general-purpose CLIP encodings, which retain a broad semantic understanding that proves advantageous even in OCT-specific applications.

\begin{table}[ht]
\caption{Performance comparison on RESC using different label text encodings: Random, MedCLIP, CLIP-base, and CLIP-large, highlighting the impact of each encoding on segmentation performance and stability.}
\centering
\fontsize{8}{10}\selectfont
\begin{tabular}{lcccc}
\toprule
 & Mean (\%) & Std & High (\%) & Low (\%)\\
\midrule
Random   & 49.94 & $\pm$1.23 & 51.26 & 47.94  \\ 
MedCLIP   & 47.46 & $\pm$2.36 & 52.11 & 45.55 \\ 
CLIP-base    & 52.36 & $\pm$2.36 & 54.63 & 48.38 \\ 
CLIP-large   & 51.32 & $\pm$1.98 & 54.18 & 49.44 \\ 
\bottomrule
\end{tabular}
\label{tab:ablation_clip}
\end{table}

We examine the impact of combining CLIP features with visual features from different encoder stages, as shown in \Cref{tab:ablation_clip_loss}. The stage selection for $\text{SIM}$ generation aligns with the included loss terms, as illustrated in \cref{fig:arch}; for example, if $\mathcal{L}_3$ is included, the $\text{SIM}$ map from Stage 3 is used. Loss terms $\mathcal{L}_1$, $\mathcal{L}_2$, $\mathcal{L}_3$, and $\mathcal{L}_4$ are defined in Eq.~\eqref{loss_func}. We can observe that raw CAM generated solely from the primary module is insufficient, yielding a mean mIoU of 49.75\%. However, adding structural information to visual features effectively improves the performance (see row 2). As shown in row 3 of \Cref{tab:ablation_clip_loss}, using only Stage 4 features with CLIP lowers mean performance to 45.85\% and reduces stability, suggesting that these abstract features lack fine-grained details required for precise lesion identification. By contrast, combining features from the last two stages ($\mathcal{L}_3 + \mathcal{L}_4$) in row 4 achieves the highest stability and performance, indicating that mid-level features from Stage 3 provide complementary spatial details that enhance the final output.

\begin{table}[ht]
\caption{Ablation study of different loss configurations on RESC. Loss terms $\mathcal{L}_1$, $\mathcal{L}_2$, $\mathcal{L}_3$, and $\mathcal{L}_4$ are defined in Eq.~\eqref{loss_func}.}
\centering
\fontsize{8}{10}\selectfont
    \begin{tabular}{lcccc}
    \toprule
    Controller & Mean(\%) & Std & High(\%) &  Low (\%)\\
    \midrule
    $\mathcal{L}_1$  & 49.75 & $\pm$1.72 & 52.09 & 47.20 \\ 
    $\mathcal{L}_1 + \mathcal{L}_2$   & 51.89 & $\pm$0.95 & 53.11 & 50.71\\ 
    $\mathcal{L}_1 + \mathcal{L}_2 + \mathcal{L}_4 $  & 45.85  & $\pm$5.88 & 53.40 & 39.14 \\ 
    $\mathcal{L}_1 + \mathcal{L}_2 + \mathcal{L}_3 + \mathcal{L}_4$   & \textbf{56.87} & $\pm$2.54 & \textbf{61.15} & \textbf{54.33}  \\ 
    \bottomrule
  \end{tabular}
  \label{tab:ablation_clip_loss}
\end{table}

\subsubsection{Structural Module Loss Analysis}
In our design, the structural module is trained as a binary classifier (healthy vs. non-healthy) to provide layer- and anomaly-aware cues that assist the primary module. We also evaluate a multi-label variant that differentiates lesion classes, mirroring the primary module. As shown in \Cref{tab:structural_loss}, the multi-label loss ($\mathcal{L}_2^M$) attains higher peak performance (55.27\% vs. 53.11\% for the binary $\mathcal{L}_2$) but exhibits significant variance across runs. This instability suggests competition with the primary module and noise amplification when enforcing fine-grained labels. In contrast, the binary objective reduces label complexity and acts as a regularizer, suppressing noise from layer- and anomaly-related signals and providing steadier structural guidance to the primary module.

\begin{table}[ht!]
\centering
\caption{Comparison of loss types for the structural module on the RESC dataset. $\mathcal{L}_1$, $\mathcal{L}_2$, and $\mathcal{L}_2^M$ denote the primary module's multi-label loss, and the structural module's binary and multi-label losses, respectively. Mean, std, highest, and lowest mIoU scores are reported over five runs.}
\begin{tabular}{lcccc}
\toprule
 & Mean(\%) & Std & High(\%) &  Low (\%)\\
\midrule
$\mathcal{L}_1 + \mathcal{L}_2$   & 51.89 & $\pm$0.95 & 53.11 & 50.71\\ 
$\mathcal{L}_1 + \mathcal{L}_2^M$   & 48.11 & $\pm$8.02 & 55.27 & 32.45\\ 
\bottomrule
\end{tabular}
\label{tab:structural_loss}
\end{table}

\subsubsection{Synthetic Description Analysis} 
\begin{figure}[ht]
\centering
\begin{subfigure}{0.25\linewidth}
\includegraphics[width=1\textwidth]{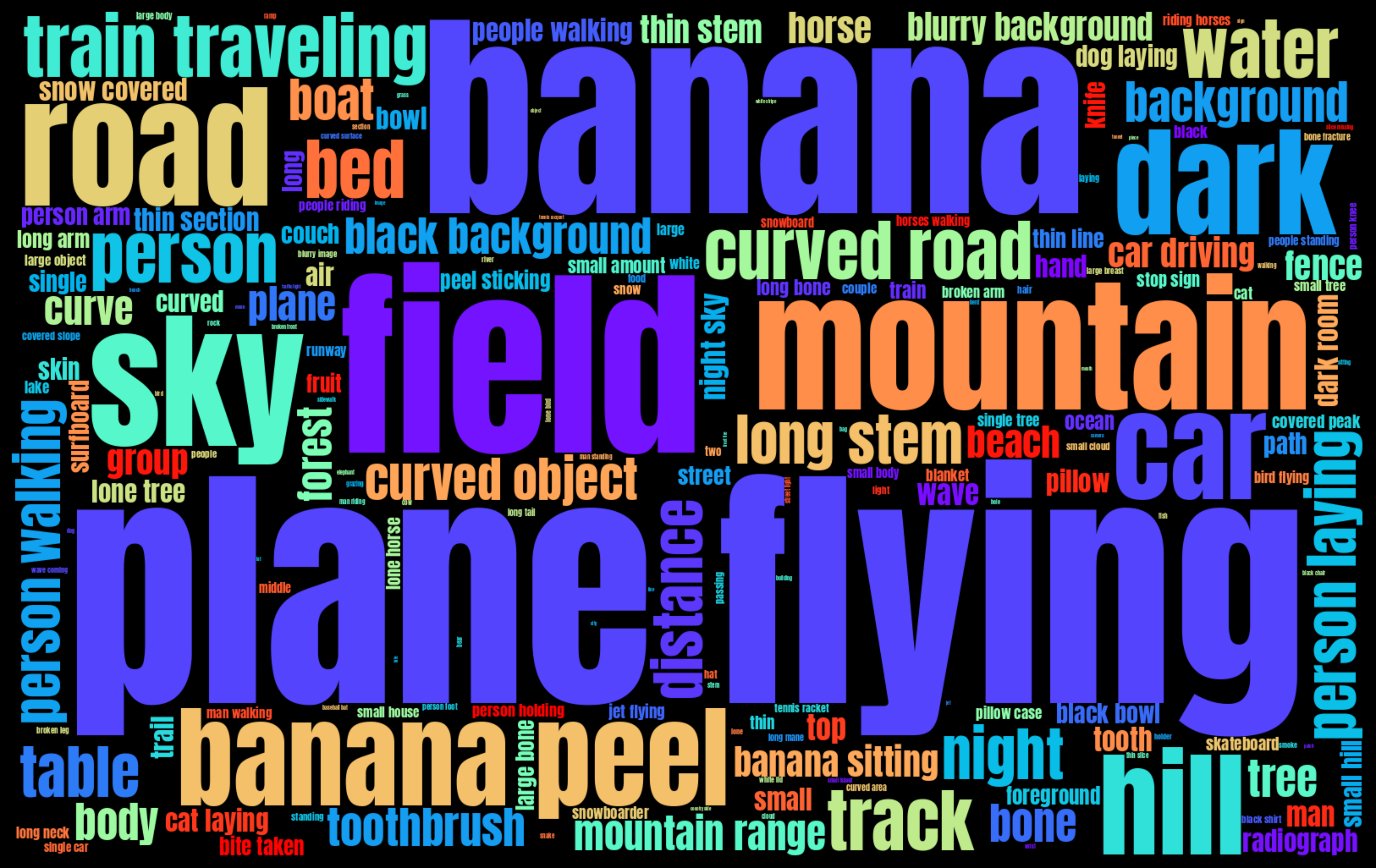}
\includegraphics[width=4cm, height=3.2cm]{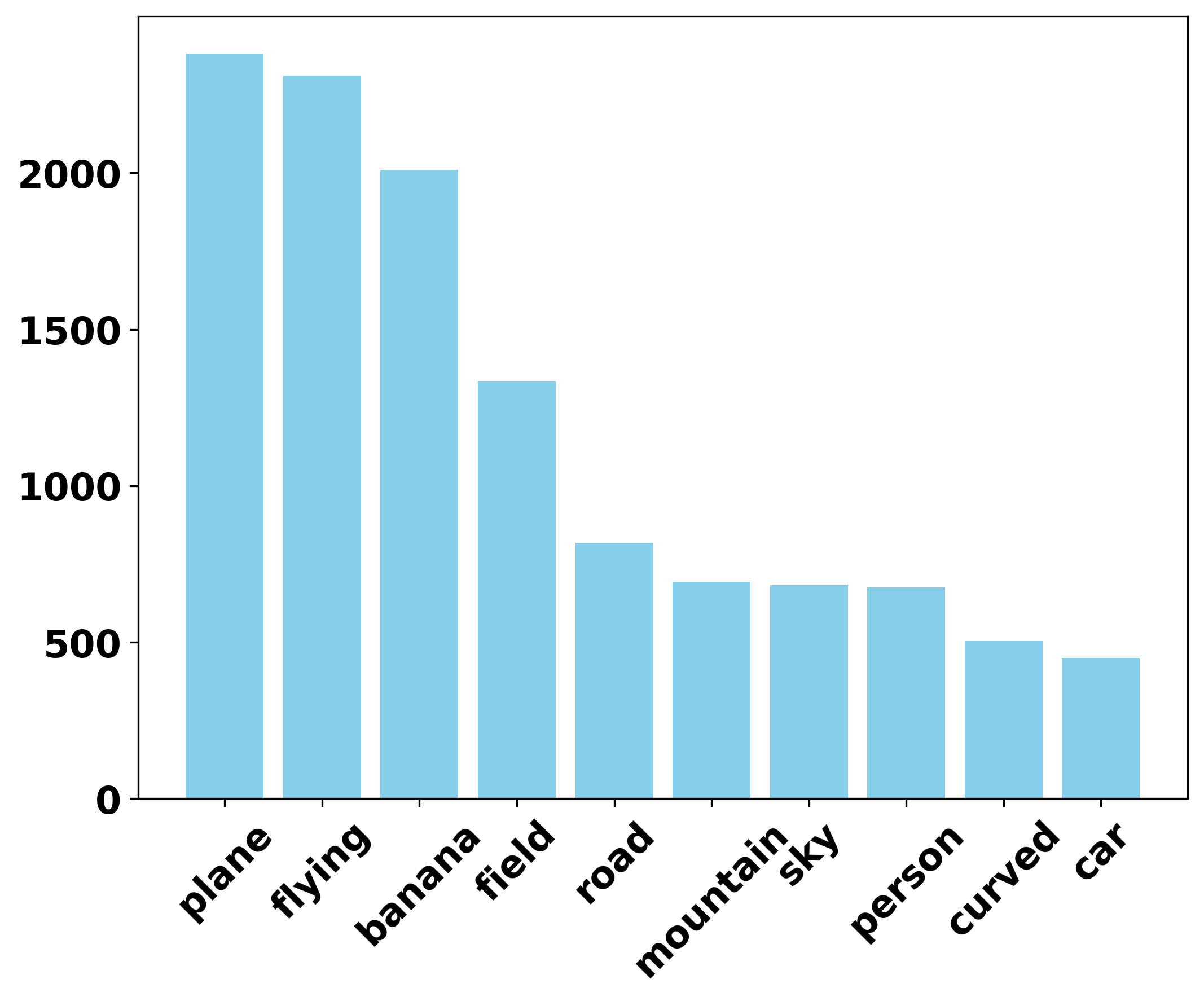}
\caption{BLIP - Healthy}
\includegraphics[width=1\textwidth]{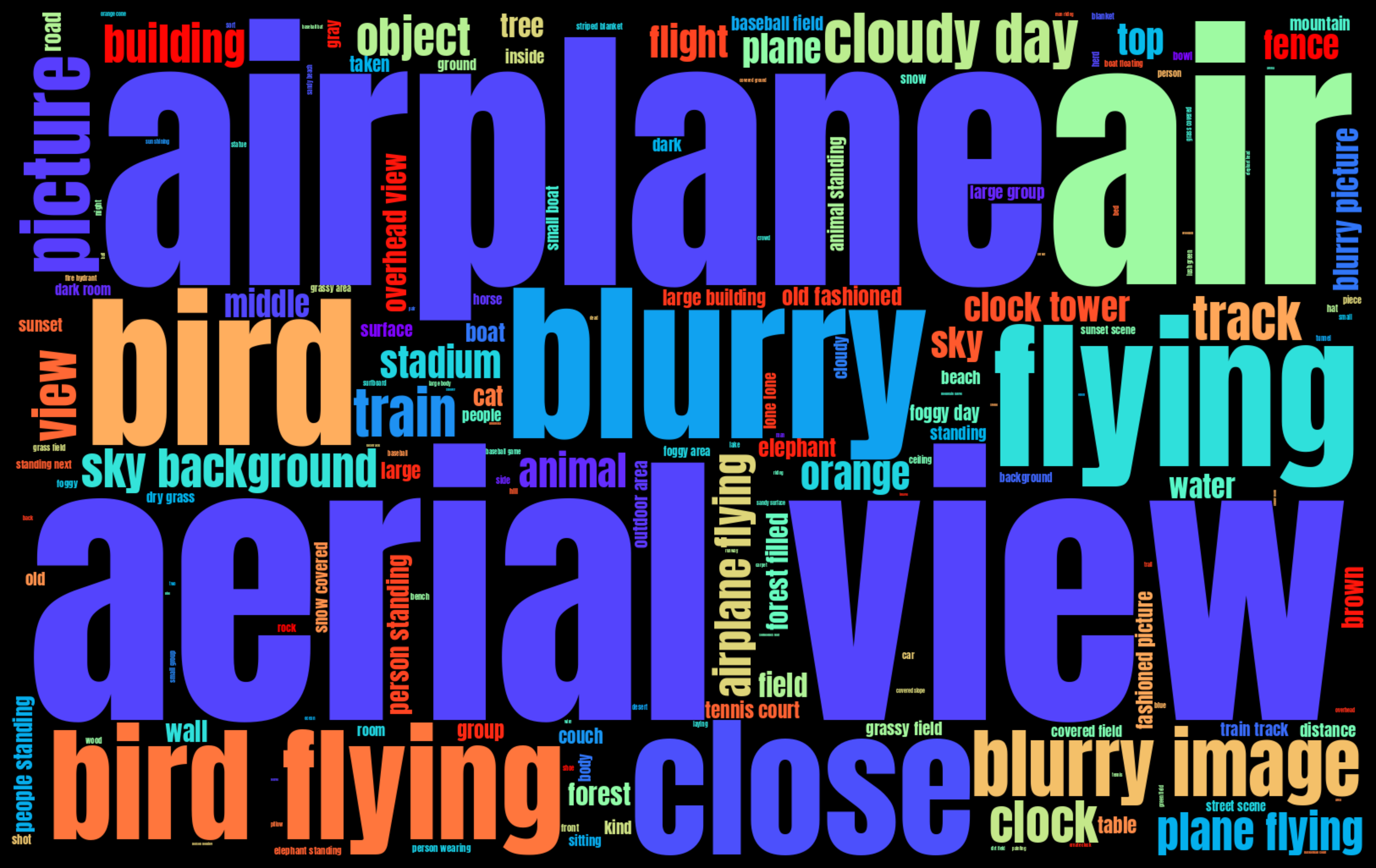}
\includegraphics[width=4cm, height=3.2cm]{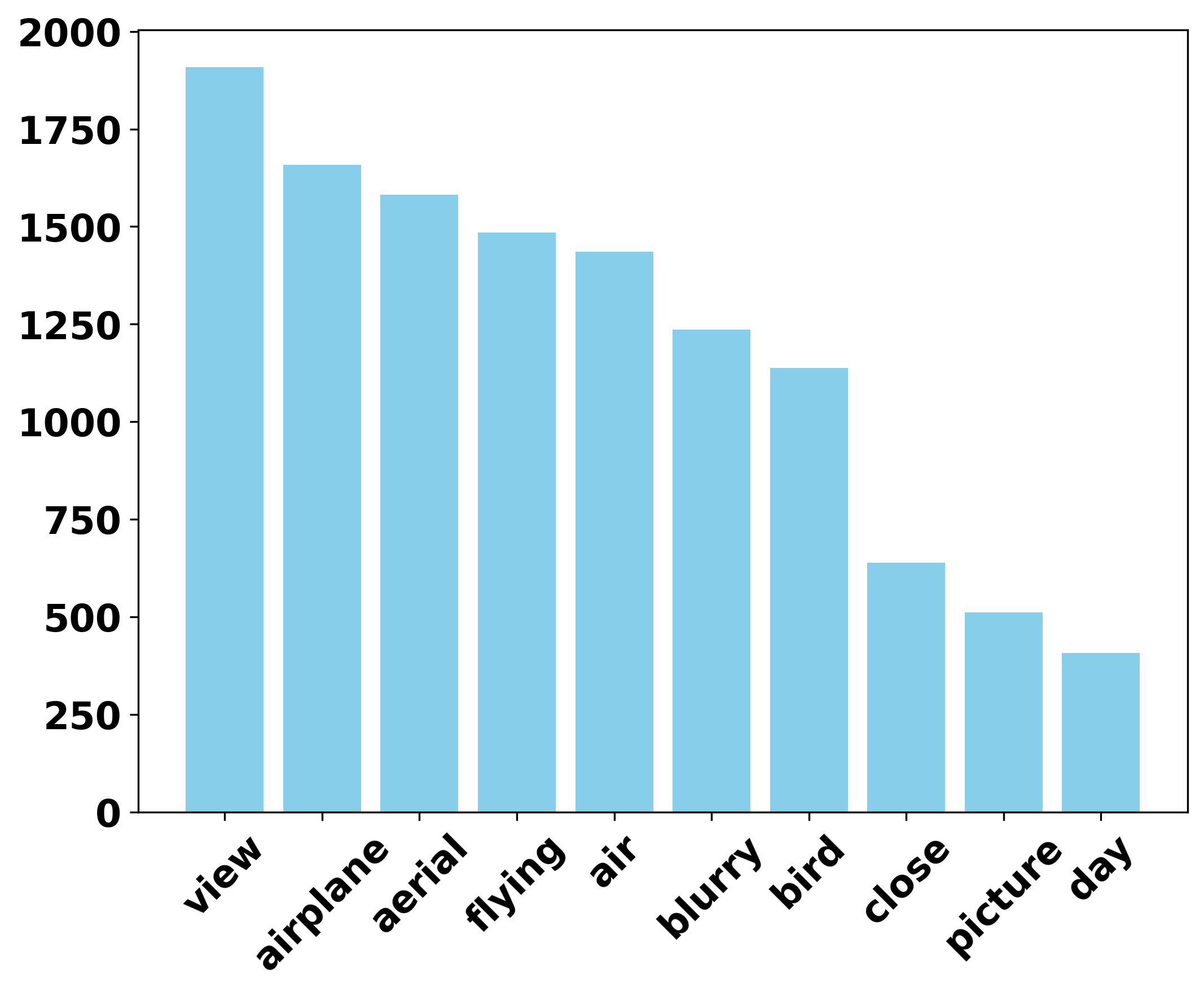}
\caption{ViT-GPT2 - Healthy}
\end{subfigure}
\begin{subfigure}{0.25\linewidth}
\includegraphics[width=1\textwidth]{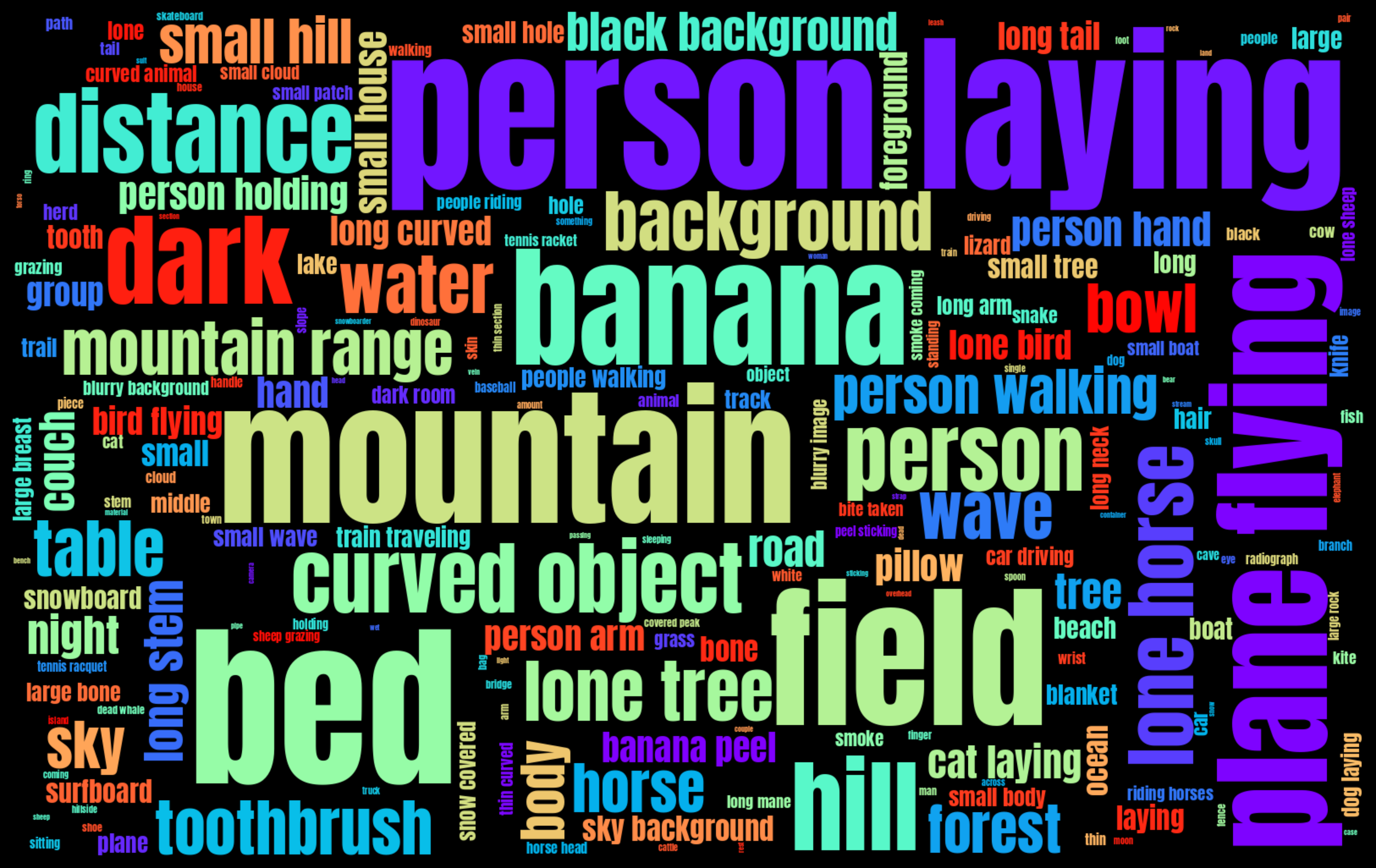}
\includegraphics[width=4cm, height=3.2cm]{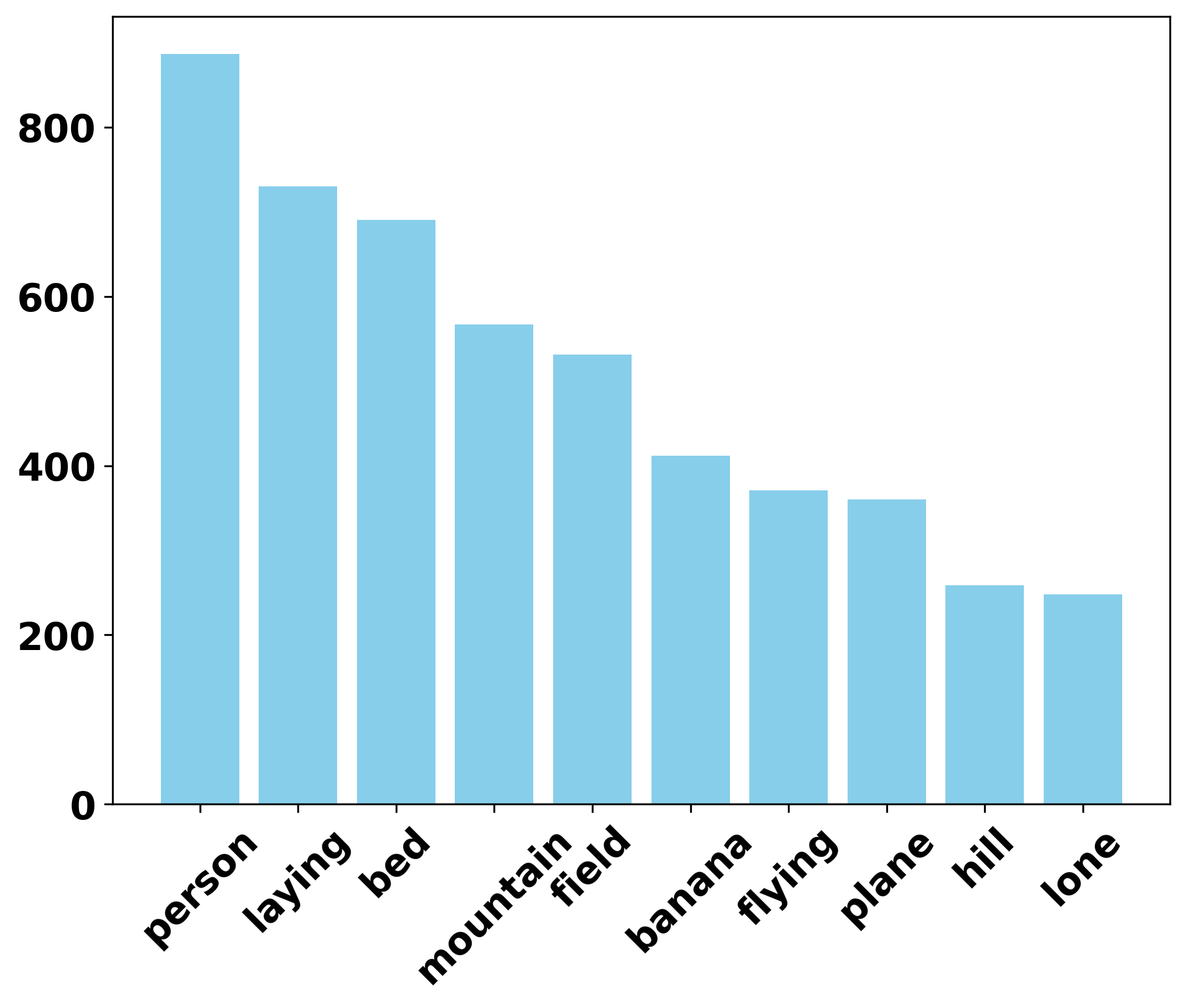}
\caption{BLIP - SRF}
\includegraphics[width=1\textwidth]{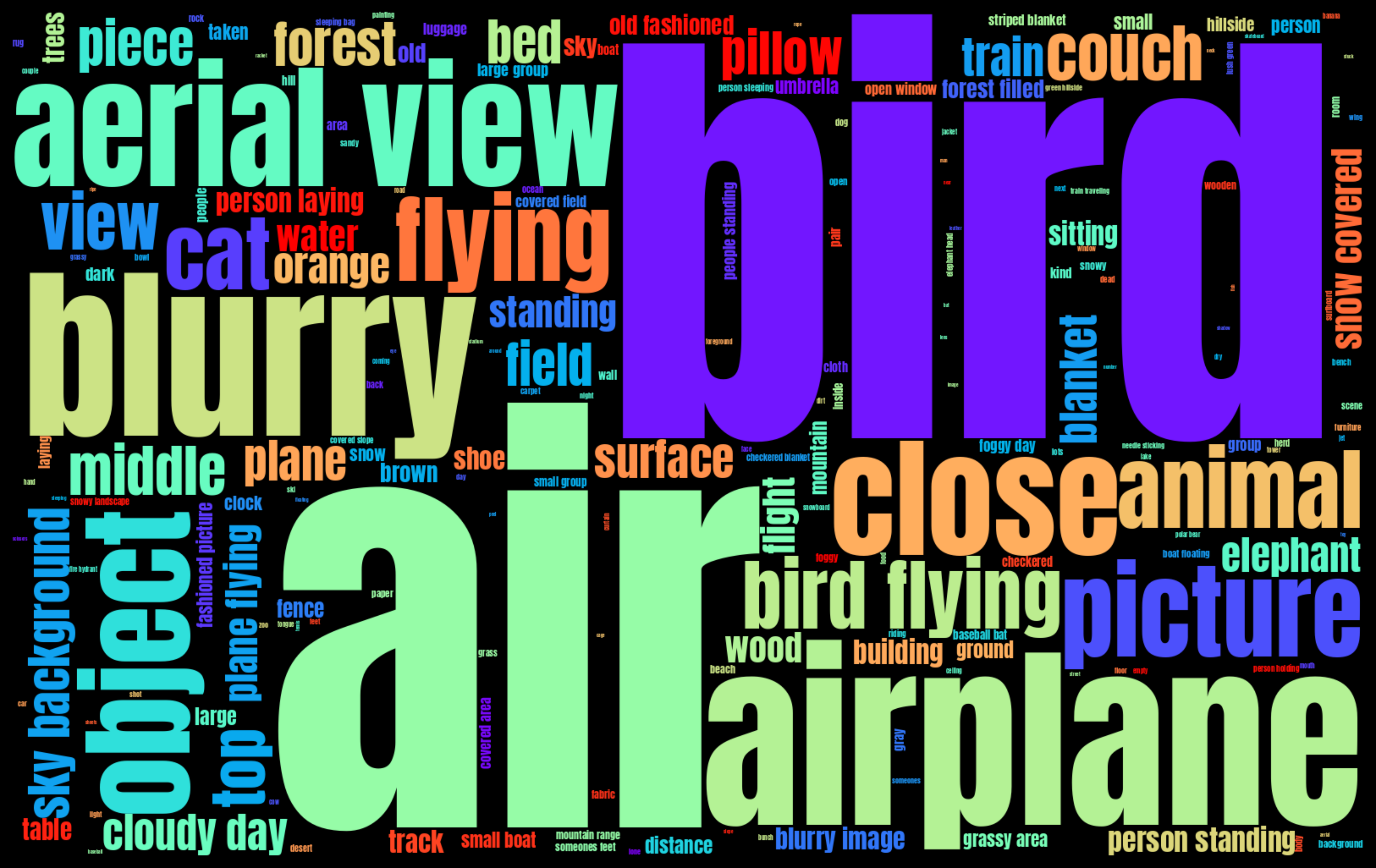}
\includegraphics[width=4cm, height=3.2cm]{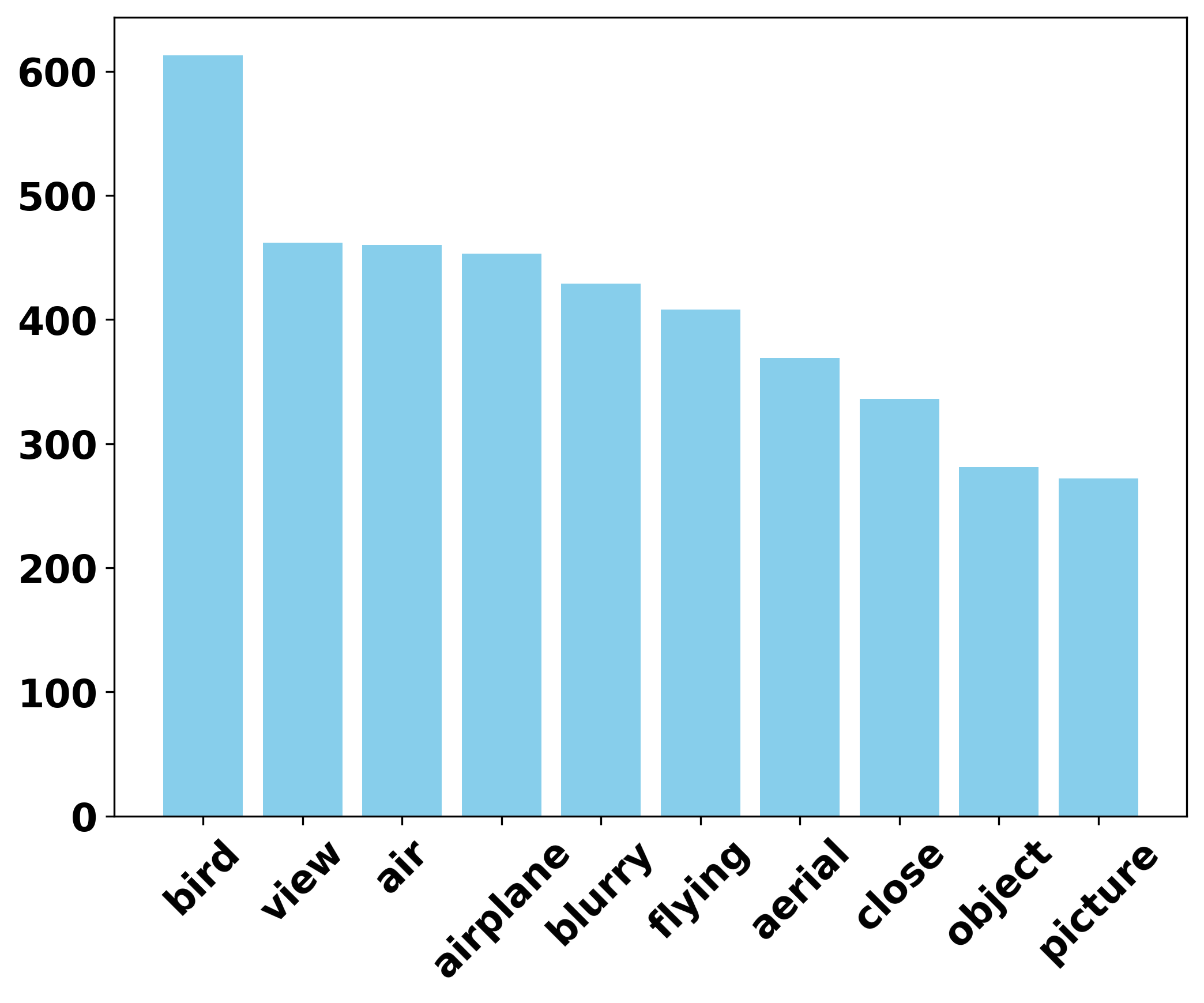}
\caption{ViT-GPT2 - SRF}
\end{subfigure}
\begin{subfigure}{0.25\linewidth}
\includegraphics[width=1\textwidth]{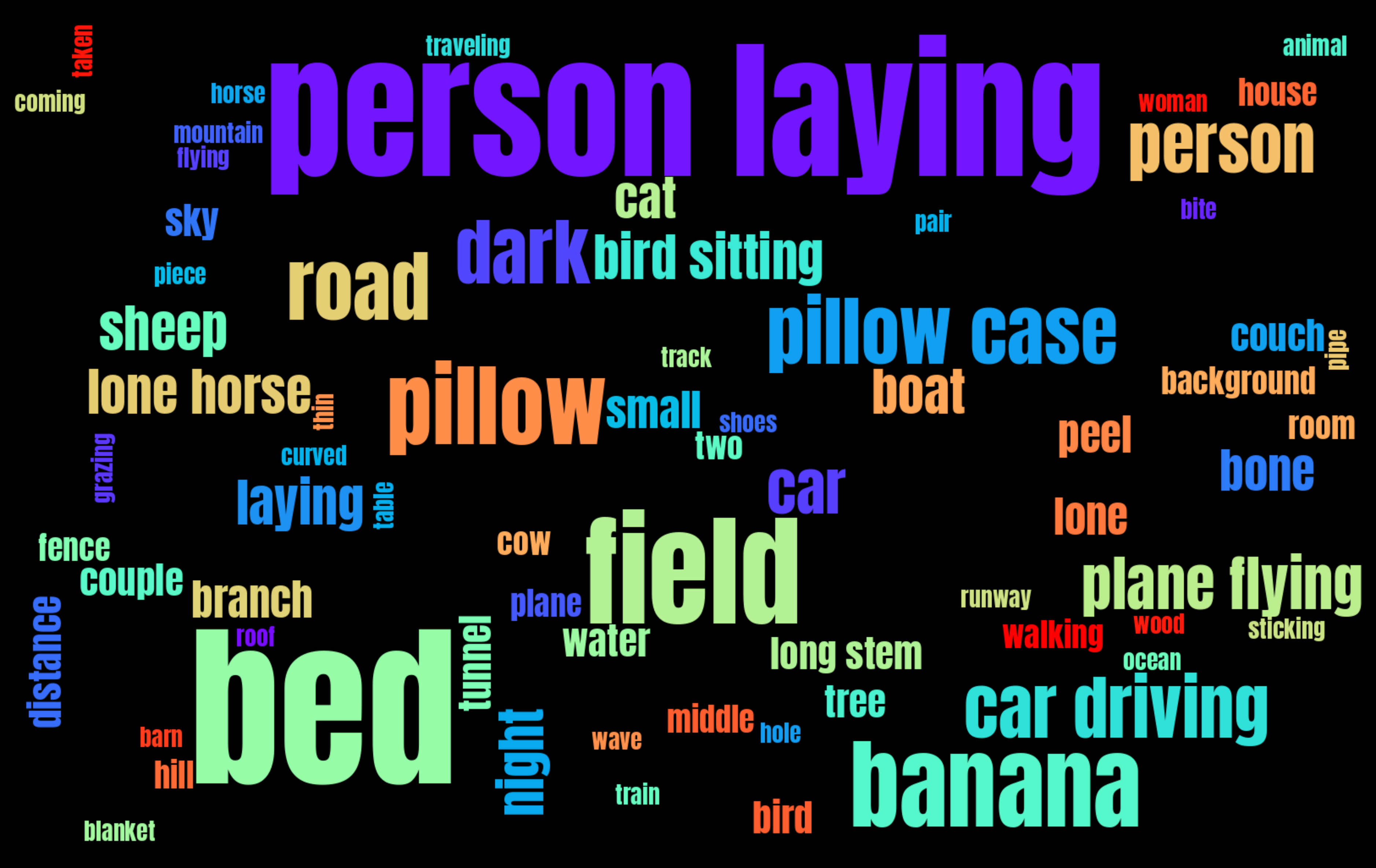}
\includegraphics[width=4cm, height=3.2cm]{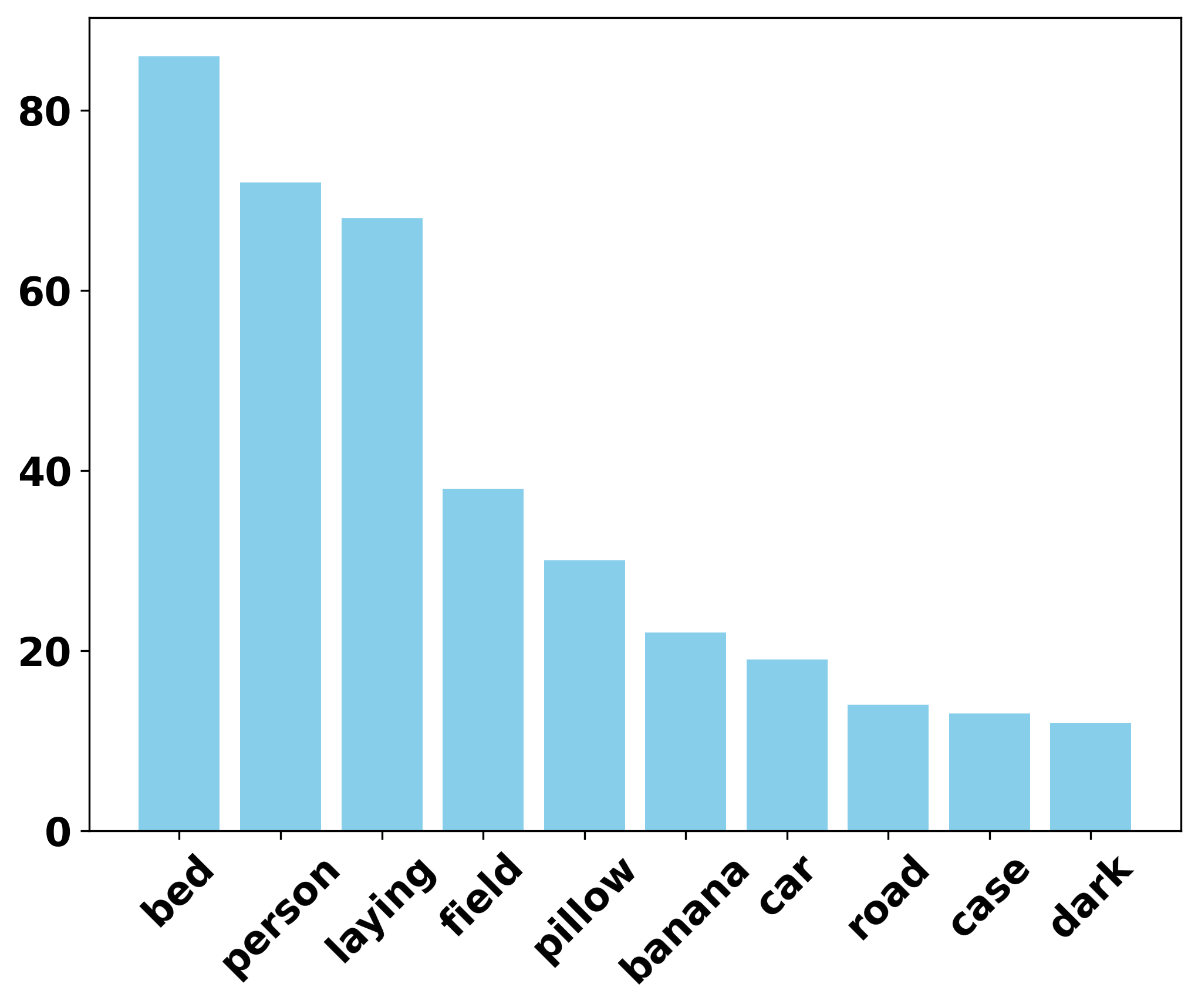}
\caption{BLIP - PED}
\includegraphics[width=1\textwidth]{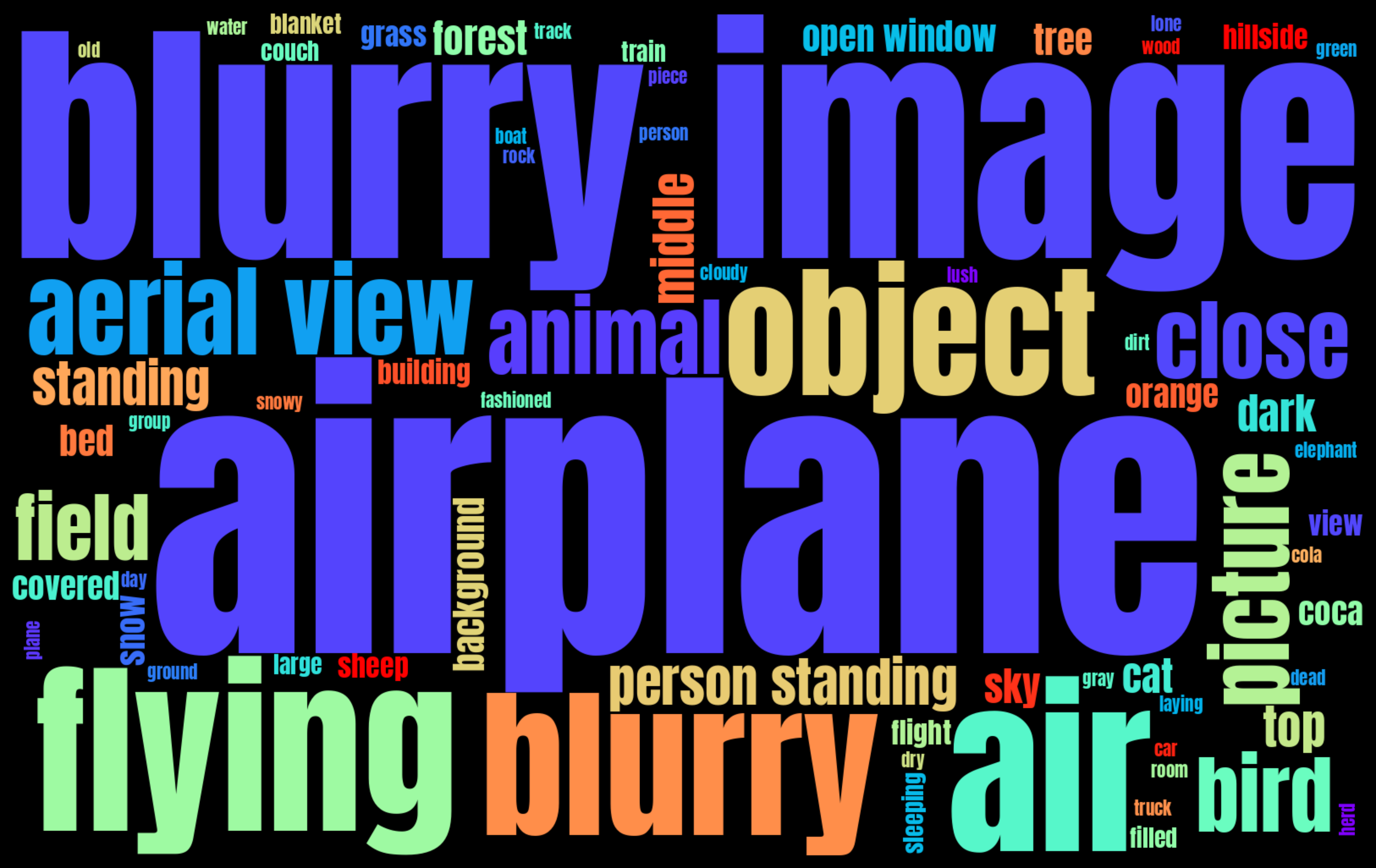}
\includegraphics[width=4cm, height=3.2cm]{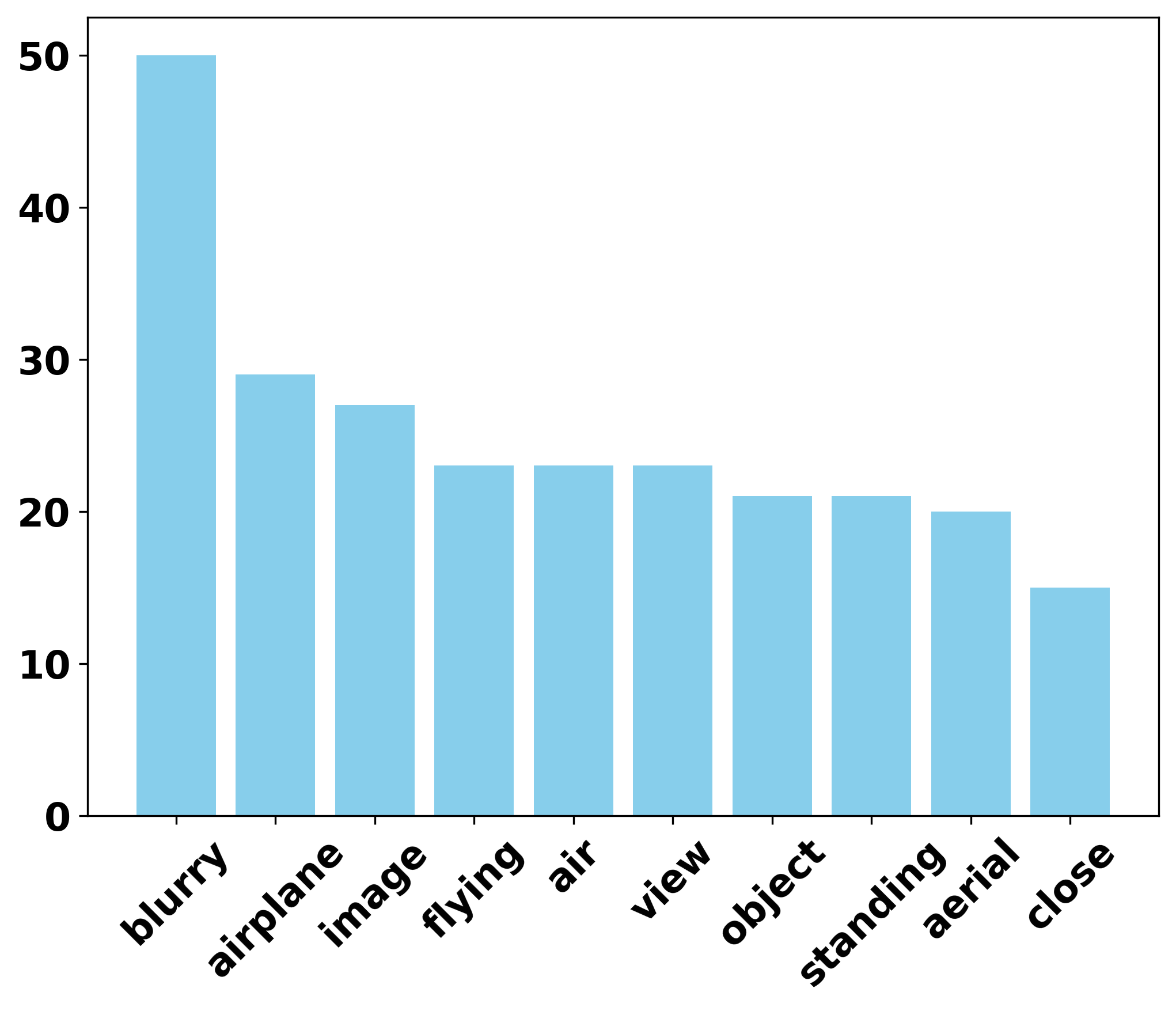}
\caption{ViT-GPT2 - PED}
\end{subfigure}
\caption{Visualization of common descriptions for each condition in the RESC dataset, displayed as word clouds and histograms of common word distributions. (a) and (b) show results for the healthy image collection using BLIP and ViT-GPT2 text generators, respectively. Similarly, (c) and (d) represent the SRF lesion collection, while (e) and (f) display the PED lesion collection for the two generators.}
\label{fig:blip_words}
\end{figure}

\begin{figure}[ht]
\centering
\subfloat[MiniLM-L12-v2]{
    \includegraphics[width=0.3\linewidth]{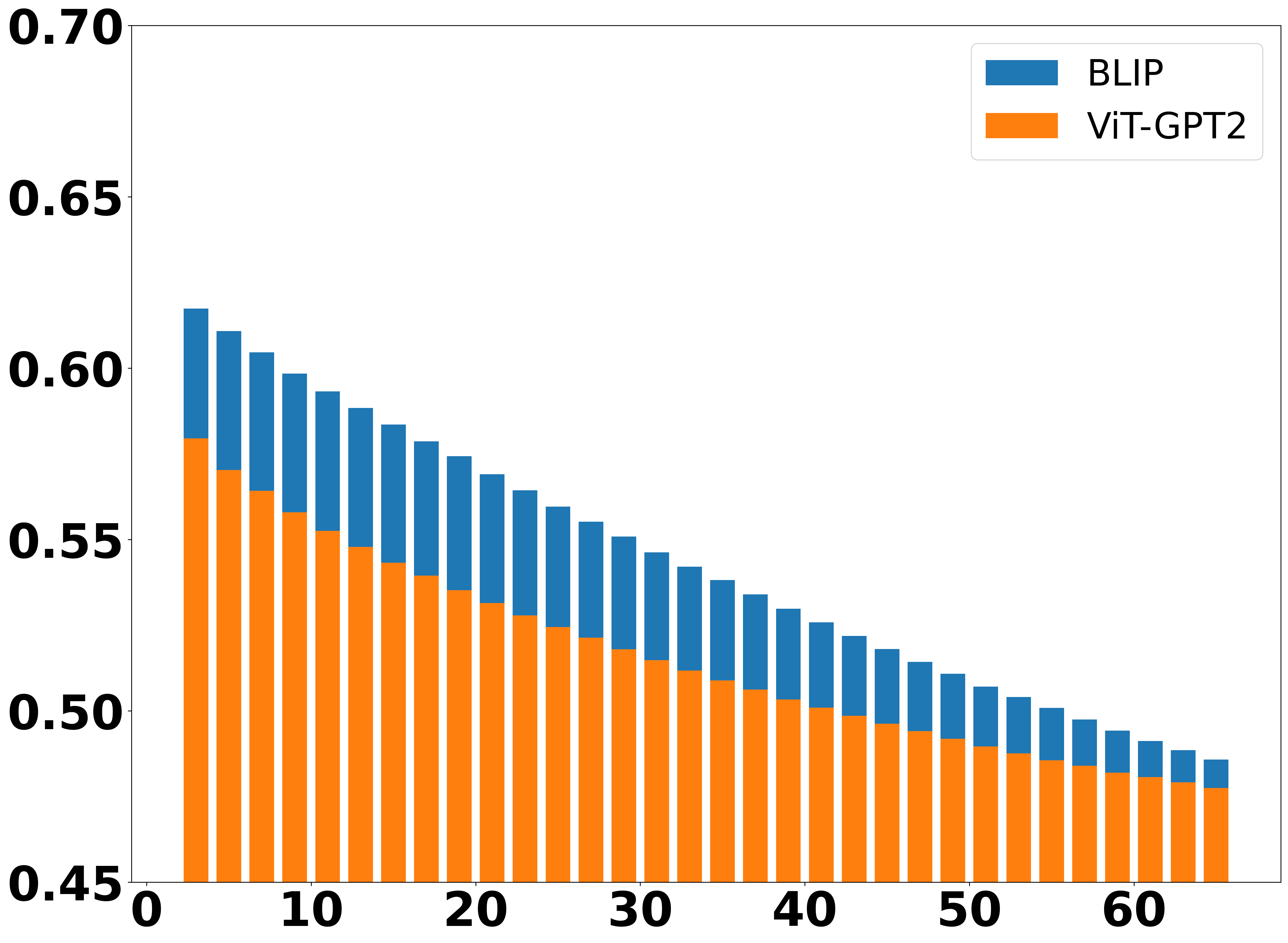}
}
\subfloat[CLIP-large]{
    \includegraphics[width=0.3\linewidth]{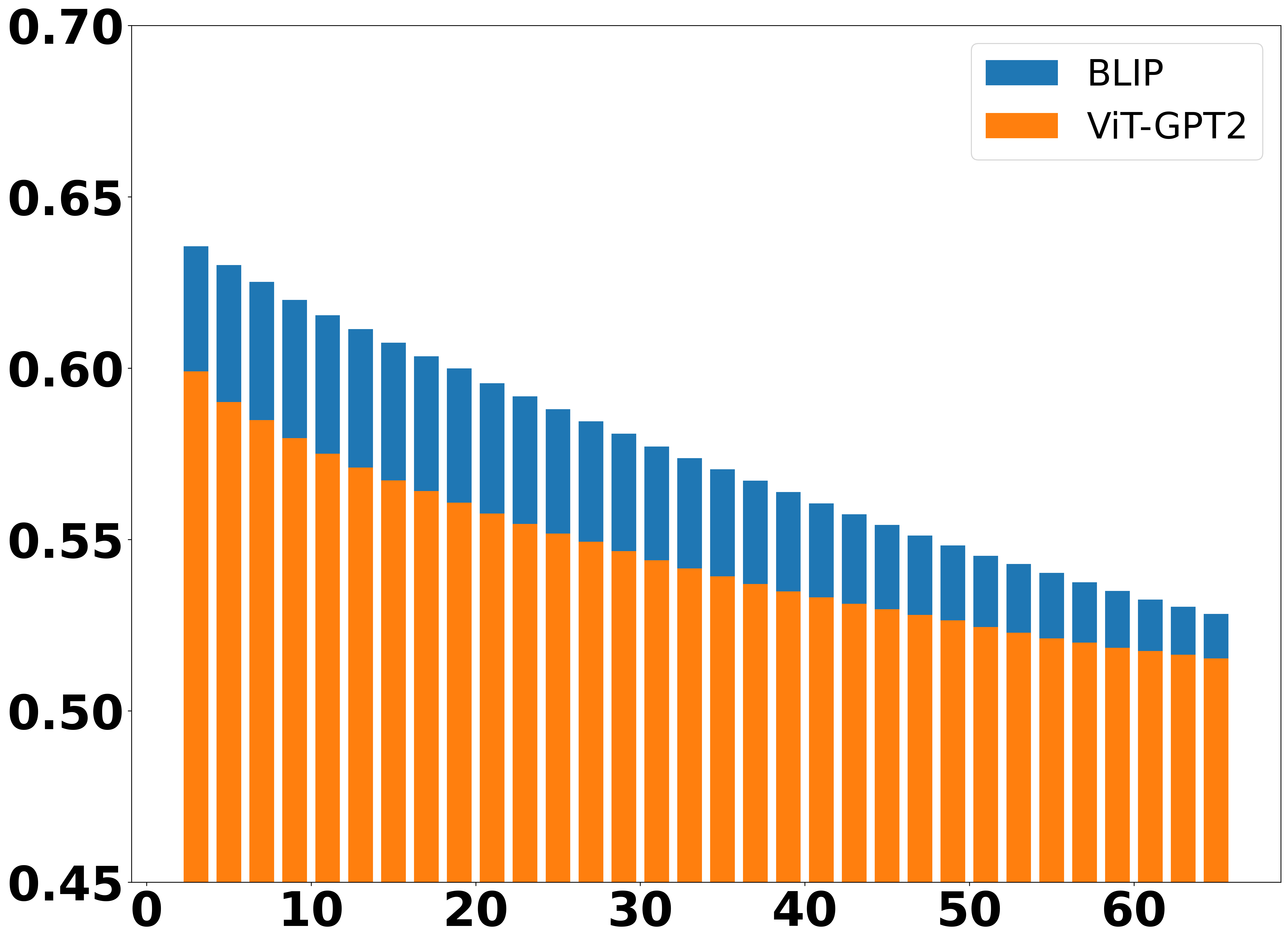}
}
\subfloat[CLIP-base]{
    \includegraphics[width=0.3\linewidth]{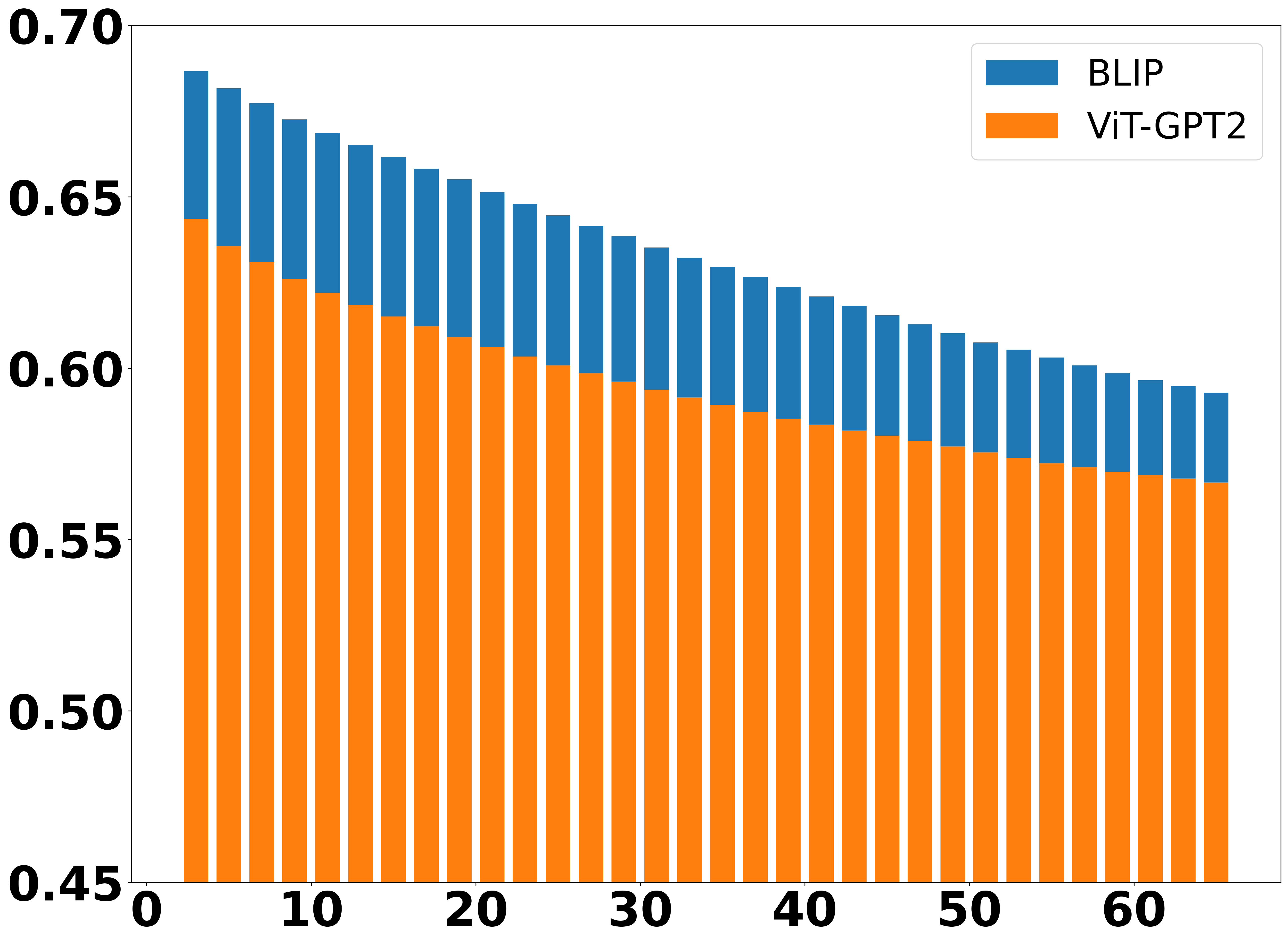}
}
\caption{Similarity score histograms of text embeddings from BLIP (blue) and ViT-GPT2 (orange) within OCT volumes on the RESC dataset, encoded by (a) MiniLM-L12-v2, (b) CLIP-large, and (c) CLIP-base. X-axis: sliding window size; Y-axis: cosine similarity. Note that blue bars represent only the visible portion extending above the orange bars.}
\label{fig:similarity}
\end{figure}

\Cref{tab:ablation_blip} presents the performance (mIoU) of pseudo labels generated using different synthetic description methods (ViT-GPT2~\footnote{https://huggingface.co/nlpconnect/vit-gpt2-image-captioning} and BLIP). We experiment with three encoders (MiniLM-L12-v2~\cite{reimers2019sentence}, CLIP-large, and CLIP-base) to convert the descriptions into vector space, with dense vector dimensions of 384, 512, and 768, respectively. We observe that BLIP with the CLIP-base encoder yields the best performance, while ViT-GPT2 generally underperforms compared to BLIP, and MiniLM-L12-v2 encoding also shows lower effectiveness than the CLIP encoders. Why do we observe this pattern, and how does non-medical description guide model learning? This question warrants further exploration, and \cref{fig:blip_words} and \cref{fig:similarity} provide insights into a potential underlying cause.

\begin{table}[ht!]
\caption{Ablation study on synthetic description methods, comparing mIoU (\%) of different text generators (ViT-GPT2, BLIP) with text encoders (CLIP-base, CLIP-large, and MiniLM-L12-v2) for embedding text into vector space.}
\centering
\fontsize{8}{10}\selectfont
\begin{tabular}{lccc}
\toprule
 &  \multicolumn{3}{c}{Text Encoders} \\
\cmidrule(){2-4}
& MiniLM-L12-v2 & CLIP-large & CLIP-base  \\
\midrule
ViT-GPT2  & 55.29 & 55.61 & 54.98\\ 
BLIP   &55.66  & 56.94 & \textbf{61.15}  \\ 
\bottomrule
\end{tabular}
\label{tab:ablation_blip}
\end{table}

From the text generator perspective, as shown in \cref{fig:blip_words}, BLIP produces more informative descriptions than ViT-GPT2. For example, BLIP outputs descriptive phrases like \textit{``person laying," ``pillow case,"} and \textit{``curved,"} while ViT-GPT2 tends to generate generic terms such as \textit{``aerial view," ``blurry,"} and \textit{``picture"}. 
This explains why BLIP is more effective than ViT-GPT2, as shown in \Cref{tab:ablation_blip}, due to BLIP’s more accurate and meaningful descriptions of the global context and relationships between objects in the images.
On the other hand, from the perspective of descriptions within the same generator across different lesion types, we observe variations in word choices: healthy images often include \textit{``plane"} and \textit{``flying,"} SRF images mention \textit{``person"} and \textit{``hill,"} and PED cases contain \textit{``pillow"} and \textit{``bed."} These high-frequency terms reflect general retinal structure and shape, supporting the value of non-medical descriptions in guiding the model to distinguish lesion types through descriptive variation.

To evaluate the consistency of synthetic descriptions, we analyze similarity scores between text embeddings of OCT slices within 3D volumes. Since adjacent slices are expected to share visual context, their generated descriptions should also be consistent. \cref{fig:similarity} shows cosine similarity scores across sliding window sizes (3 to 65) for BLIP and ViT-GPT2 text embeddings in the RESC dataset. Higher similarity scores are expected for smaller window sizes, decreasing as more distant slices are included. Each score represents the average similarity between neighboring slices and the central target slice. The results indicate that BLIP achieves higher similarity scores than ViT-GPT2, reflecting stronger contextual consistency. 
Among encoders, CLIP-base achieves the highest similarity, aligning with the superior mIoU observed for BLIP with CLIP-base in \Cref{tab:ablation_blip}.
These findings underscore the value of synthetic non-medical text, as its consistent descriptions of the OCT domain enhance model performance.

\subsubsection{Synthetic Description 2D Projection}
To further understand the impact of different generator-encoder combinations observed in Table 8, we visualize the 2D projections of synthetic descriptions using Linear Discriminant Analysis (LDA). While Fig. 5 illustrates the phrase distribution across classes and Fig. 6 examines semantic consistency across neighboring slices, Fig. 7 provides a deeper view of class separability in the text embedding space. Notably, the combination of BLIP-generated descriptions with CLIP encoders shows clearer inter-class boundaries and tighter intra-class clusters, indicating that the synthetic descriptions encode meaningful semantics rather than merely contributing generic textual diversity.
\begin{figure}[ht]
\centering
\subfloat[ViT-GPT2, MiniLM-L12-v2]{
    \includegraphics[width=0.42\linewidth]{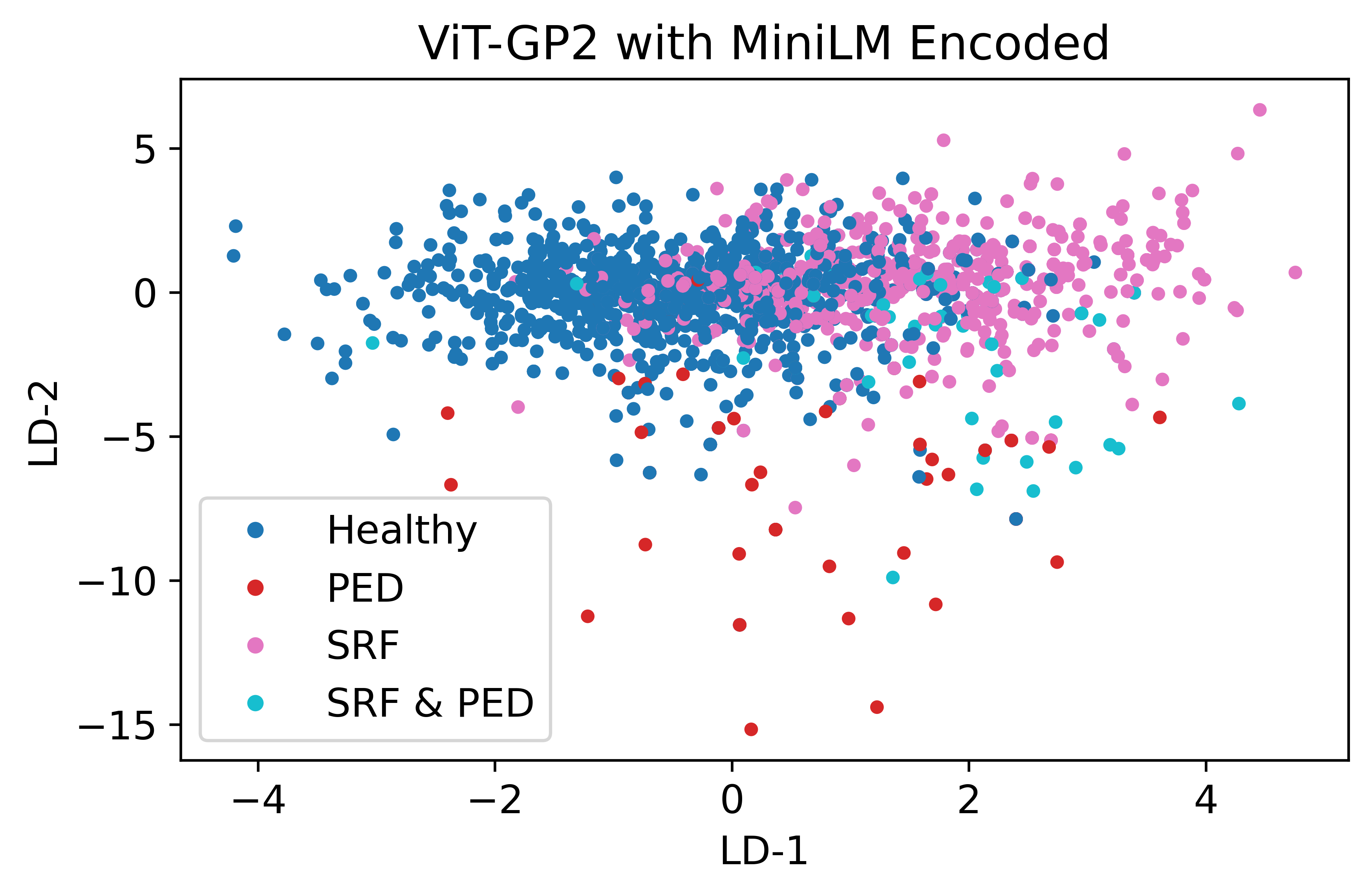}
}
\subfloat[ViT-GPT2, CLIP-large]{
    \includegraphics[width=0.42\linewidth]{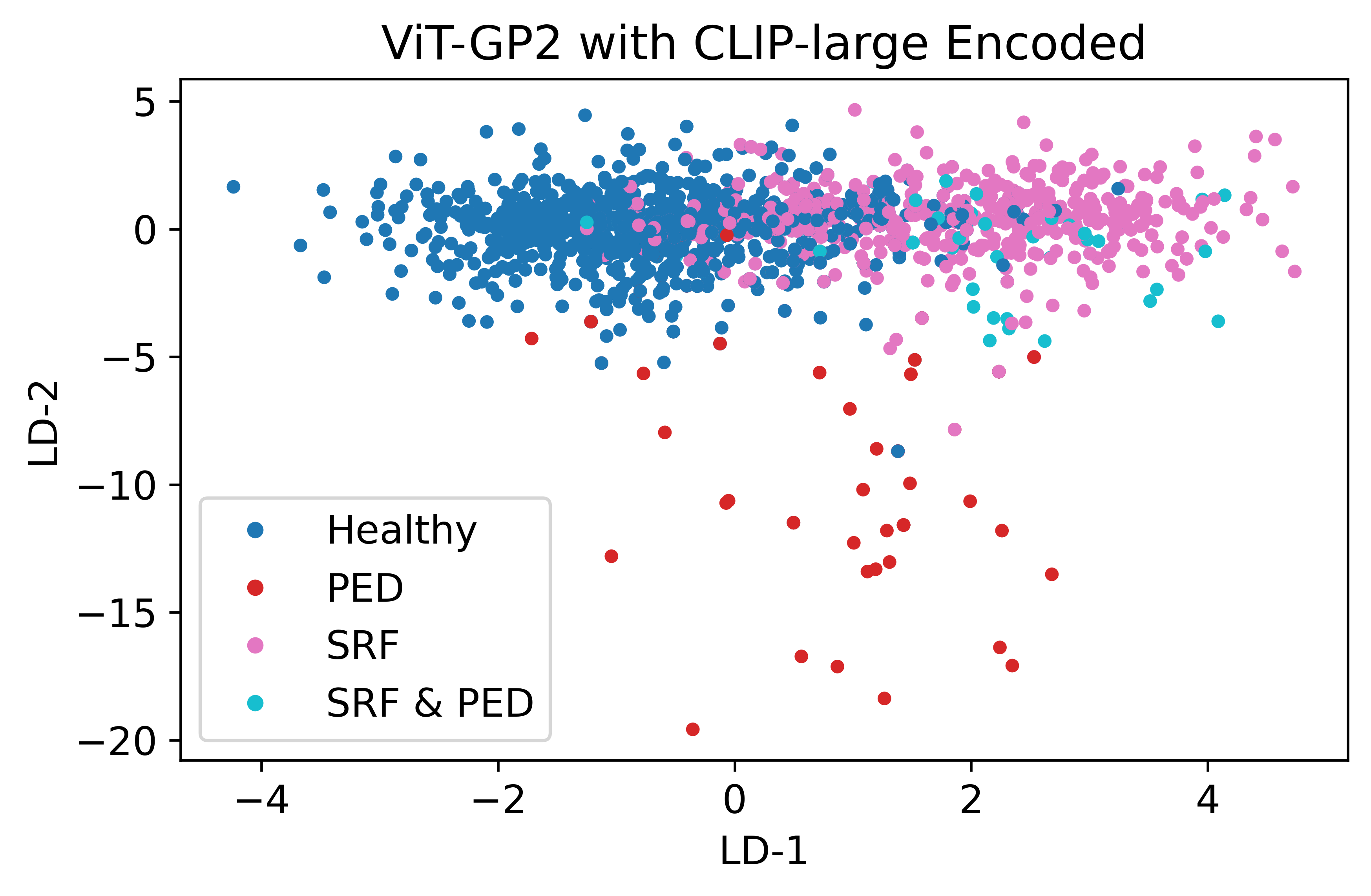}
}\hfill
\subfloat[BLIP, MiniLM-L12-v2]{
    \includegraphics[width=0.42\linewidth]{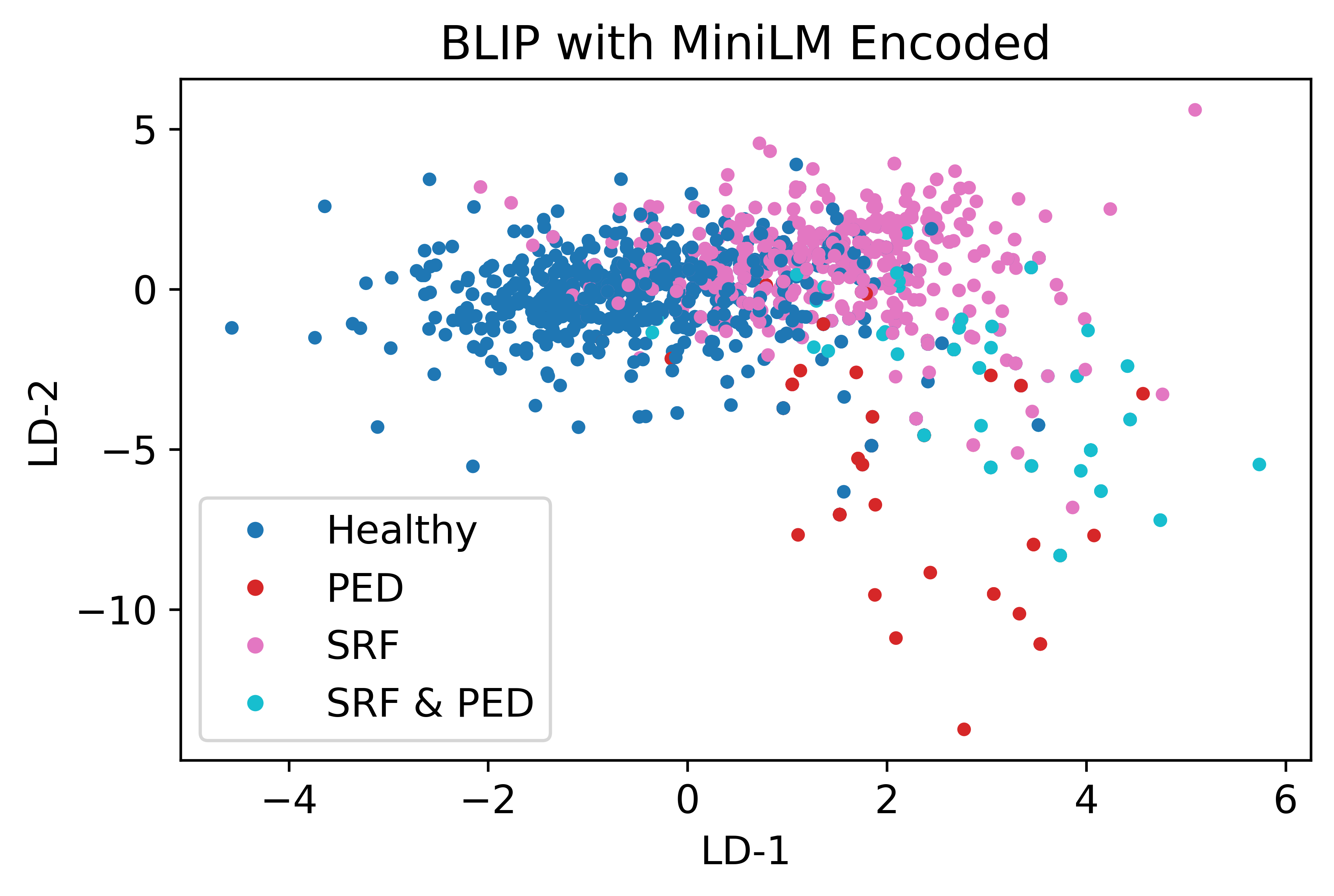}
}
\subfloat[BLIP, CLIP-large]{
    \includegraphics[width=0.42\linewidth]{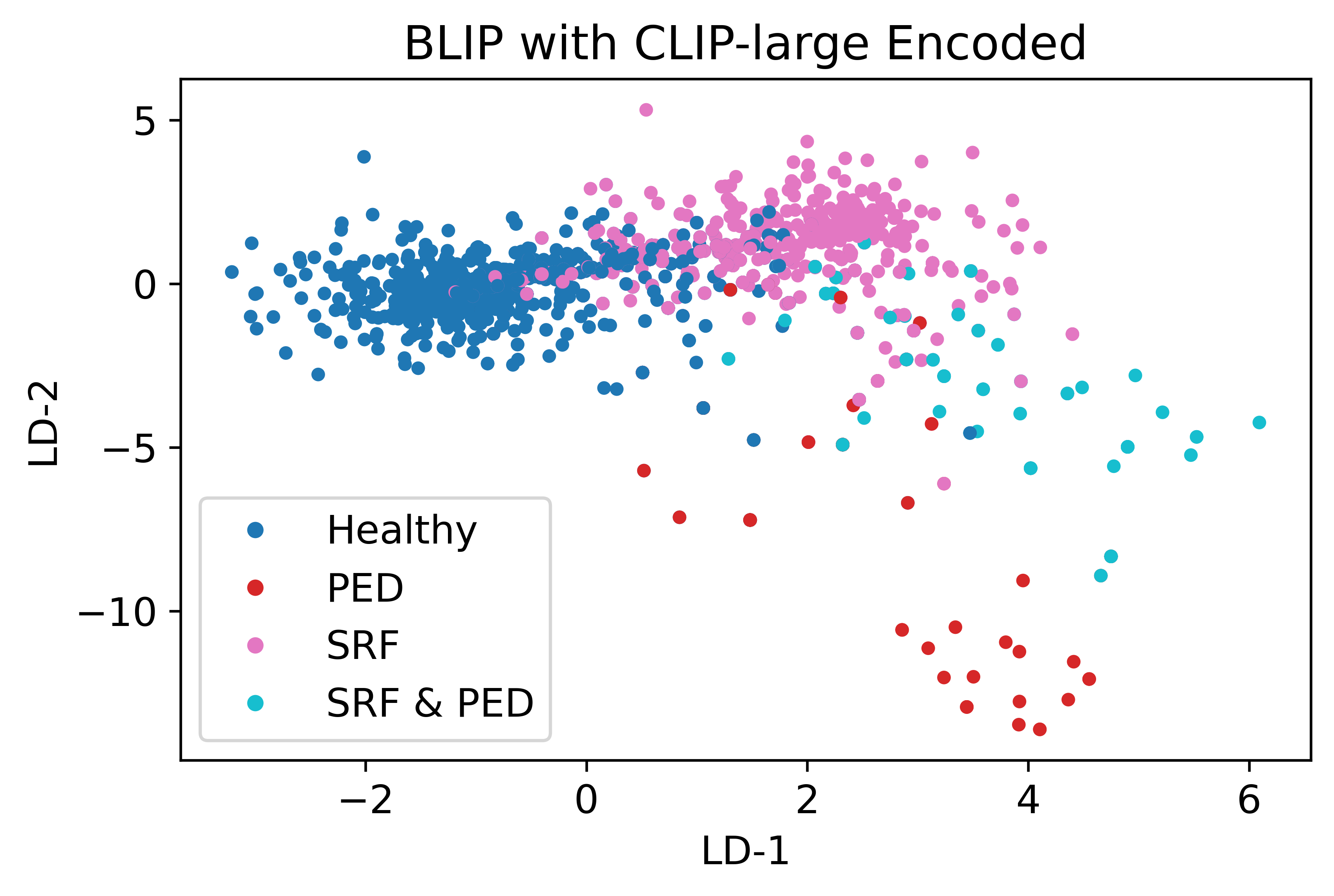}
}
\caption{LDA projections of synthetic descriptions. (a–b): ViT-GPT2 generated texts encoded by MiniLM and CLIP-Large; (c–d): BLIP-generated texts encoded by MiniLM and CLIP-Large.}
\label{fig:2d}
\end{figure}

\subsubsection{Validating the Semantic Value of Synthetic Description}
Furthermore, we investigate the necessity of the synthetic descriptions by comparing their embeddings with direct vision embeddings from pretrained BLIP. As shown in \Cref{tab:ablation_blip_vision_encode}, the results indicate that text-encoded vectors, which leverage both visual and textual modalities, significantly outperform direct vision embeddings. While the direct vision embeddings in BLIP represent a foundational visual encoding of the image that is well-aligned with text, they do not incorporate the advanced semantic and contextual interactions that occur during the later stages of BLIP's captioning process. These interactions involve iterative alignment between visual features and linguistic components, enabling the generation of accurate and contextually rich textual descriptions. This finding highlights the importance of utilizing the full vision-language processing pipeline, which integrates multi-modal interactions to capture nuanced semantic relationships. It also confirms that the performance gain is not merely from increased feature diversity, but from the discriminative semantic content embedded in the generated captions.

\begin{table}[ht]
\caption{Comparison of mIoU using BLIP vision encoder embeddings vs. BLIP-generated captions encoded with CLIP-base.}
\centering
\fontsize{8}{10}\selectfont
\begin{tabular}{lcccc}
    \toprule
    Encode Method & Mean(\%) & Std & High(\%) &  Low (\%)\\
    \midrule
    BLIP-vision-embedding   & 51.86 & $\pm$2.53 & 55.22 & 48.02\\ 
    BLIP + CLIP-base  & \textbf{56.87} & $\pm$2.54 & \textbf{61.15} & \textbf{54.33} \\
    \bottomrule
  \end{tabular}
  \label{tab:ablation_blip_vision_encode}
\end{table}

\subsubsection{Backbone Analysis}
To evaluate backbone size and stage-freezing strategy, we present the performance of different MiT sizes and frozen stages in \Cref{tab:backbones}. Fully fine-tuning all parameters leads to instability and poorer performance on OCT images, as seen with MiT-b2 in the first row, which shows high variance and a mean mIoU of only 52.33\%. This suggests freezing early stages effectively leverages pretrained feature representations and stabilizes training. MiT-b1 (\ding{100} S1+S2), being smaller, reaches only a peak mIoU of 55.76\%. In contrast, the large MiT-b5 (\ding{100} S1+S2), while powerful, is overly heavy and potentially limits the contribution of other modules. Although further tuning could benefit MiT-b5, it remains less efficient in our setup. Balancing efficiency and performance, we select MiT-b2 (\ding{100} S1+S2) for our final design.

\setlength{\tabcolsep}{2.5pt}
\begin{table}[ht]
\caption{Ablation on the RESC dataset using different MiT backbone sizes. S1 and S2 denote frozen stages (\ding{100}). We report mIoU statistics over five runs, along with trainable parameters (millions). The best results are in \textbf{bold}.}
  \centering
  \fontsize{8}{10}\selectfont
  \begin{tabular}{lccccc}
    \toprule
     Backbone & Mean (\%) & Std & High (\%) & Low (\%) & Params (M)\\
    \midrule
    MiT-b2 & 52.33 & $\pm$5.10 & 58.55 & 43.23 & 54.1\\ 
    MiT-b2 (\ding{100} S1) & 51.34 & $\pm$3.25 &  55.85 & 47.33 & 48.3\\
    \rowcolor{gray!10}MiT-b2 (\ding{100} S1+S2) & \textbf{56.87} & $\pm$2.54 & \textbf{61.15} & \textbf{54.33} & 44.6 \\
    MiT-b1 (\ding{100} S1+S2) & 54.00 & $\pm$2.16  & 55.76 & 49.83 & 25.0\\ 
    MiT-b5 (\ding{100} S1+S2) & 52.54 & $\pm$3.24 & 57.77 & 48.00 & 157.2\\ 
    \bottomrule
  \end{tabular}
  \label{tab:backbones}
\end{table}

\subsubsection{Pooling Layer Analysis}
We investigate pooling strategies within our weakly supervised framework, comparing GMP, GAP, Top-k (k=3,5), and an adaptive-k variant that retains activations above a channel-wise fraction. While recent ViT-based WSSS uses top-k/adaptive-k with patch-contrastive learning~\cite{wu2023tkp,wu2024apc}, here we isolate the effect of pooling alone in a CAM-based setting. On RESC, GMP yields the best pseudo-label mIoU (\Cref{tab:ablation_pooling}). This result is consistent with prior findings in AGM~\cite{yang2024anomaly}, where target lesions typically occupy only a small portion of the scan. By emphasizing the most discriminative activations, GMP enables the classification network to generate more localized and accurate CAMs.
\begin{table}[ht]
    \caption{Comparison of pseudo-label mIoU (\%) using different pooling strategies on the RESC dataset. Best result in bold.}
    \centering
    \small
    \begin{tabular}{ccccc}
    \toprule
      GAP & Top-5 & Top-3 & Adaptive-k & GMP \\
    \midrule
      34.55 & 54.36 & 57.15 & 55.37 & \textbf{61.15} \\ 
    \bottomrule
    \end{tabular}
    \label{tab:ablation_pooling}
\end{table}

\subsection{Qualitative Results}

\begin{figure*}[ht!]
\centering

\begin{subfigure}{0.13\linewidth}
\includegraphics[trim={0 13cm 0 6cm},clip,width=1\textwidth]{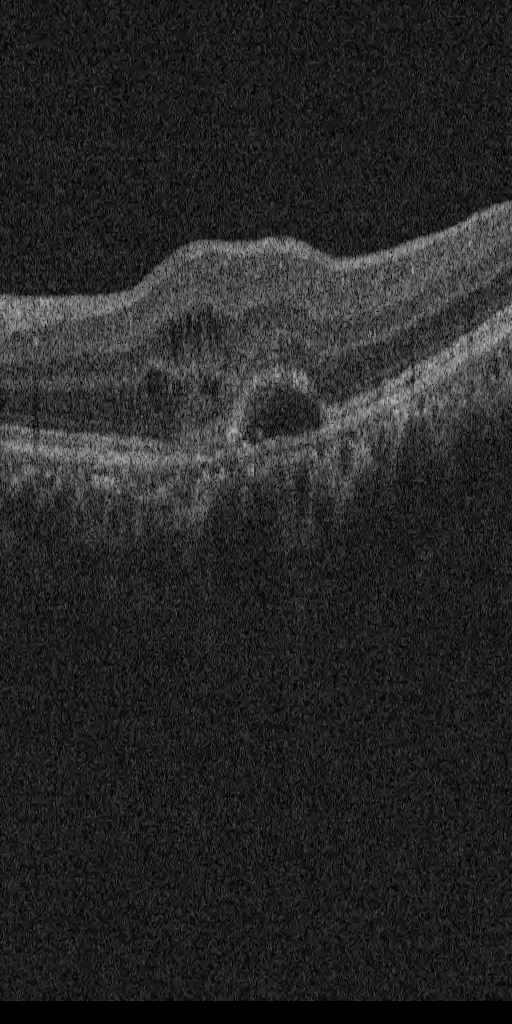}
\includegraphics[trim={0 3cm 0 14cm},clip,width=1\textwidth]{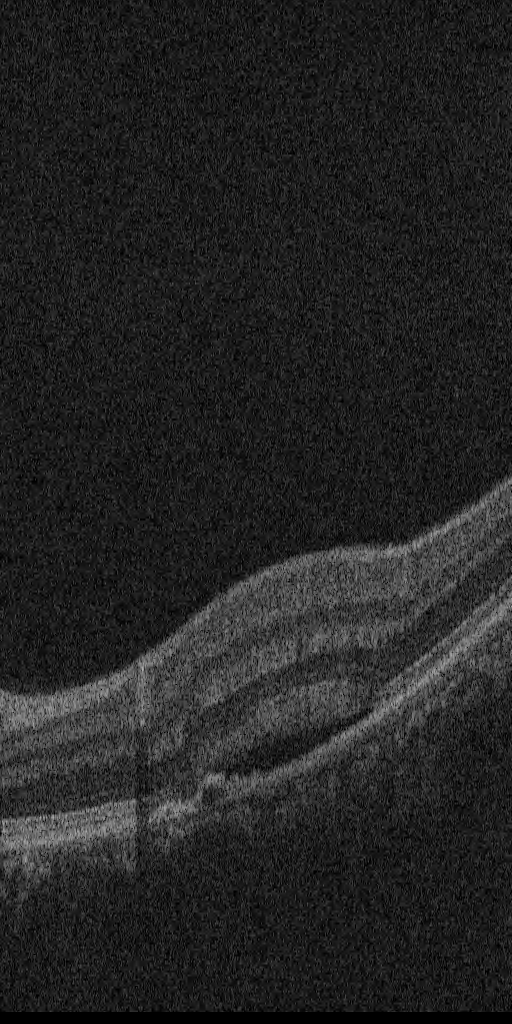}
\includegraphics[trim={0 9cm 0 12cm},clip,width=1\textwidth]{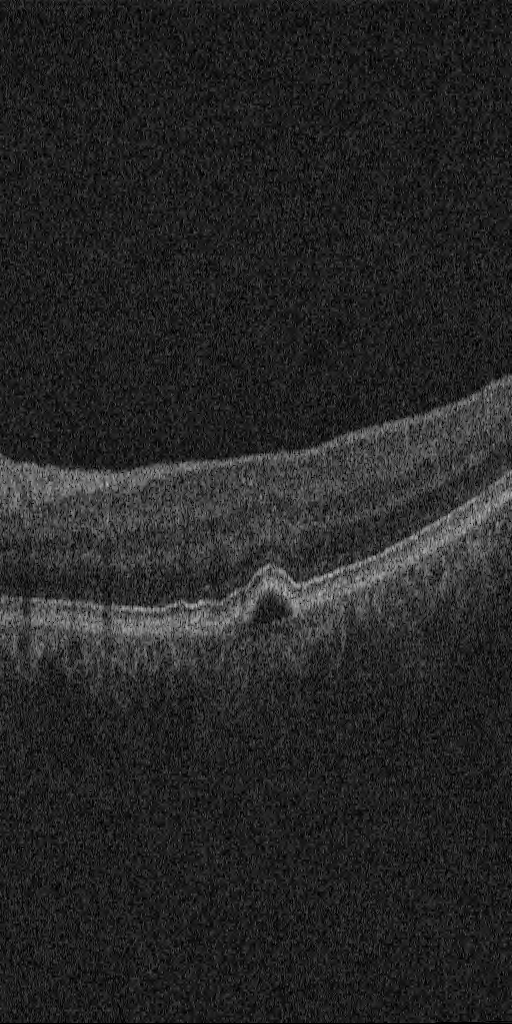}
\includegraphics[width=1\textwidth]{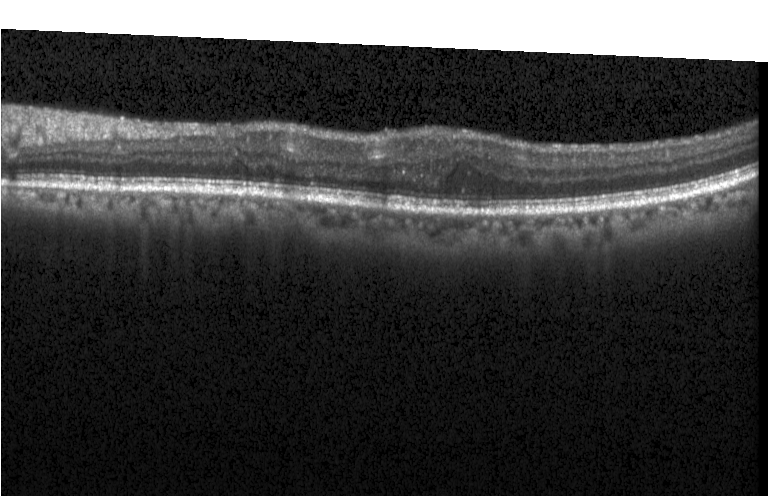}
\includegraphics[trim={0 0 0 3cm},clip,width=1\textwidth]{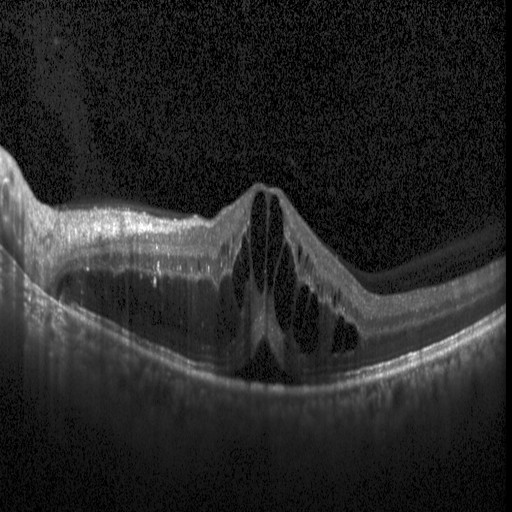}
\caption{Images}
\end{subfigure}
\begin{subfigure}{0.13\linewidth}
\includegraphics[trim={0 13cm 0 6cm},clip,width=1\textwidth]{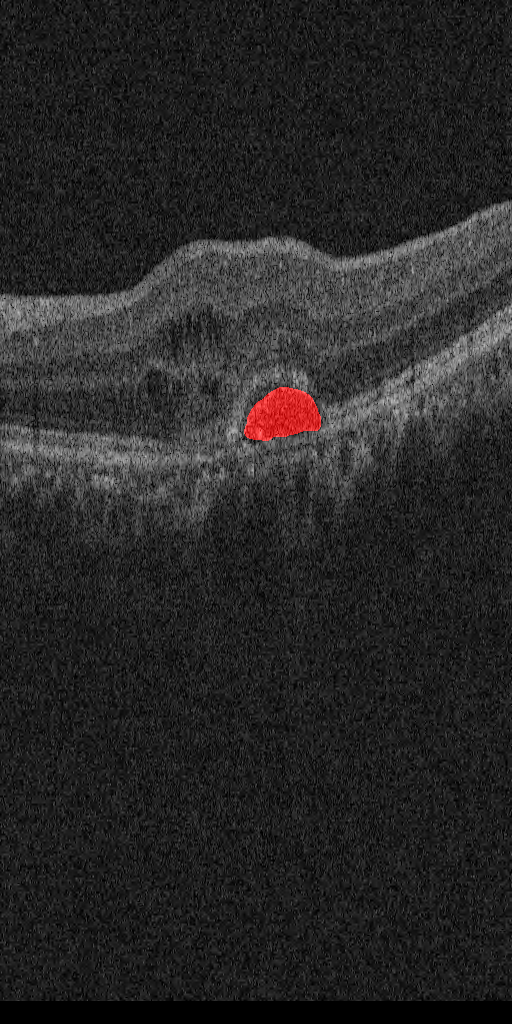}
\includegraphics[trim={0 3cm 0 14cm},clip,width=1\textwidth]{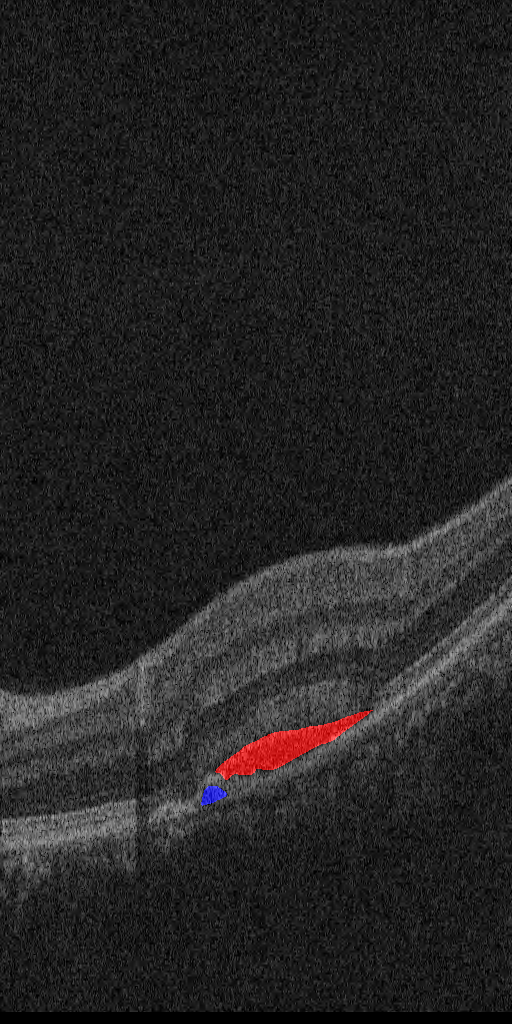}
\includegraphics[trim={0 9cm 0 12cm},clip,width=1\textwidth]{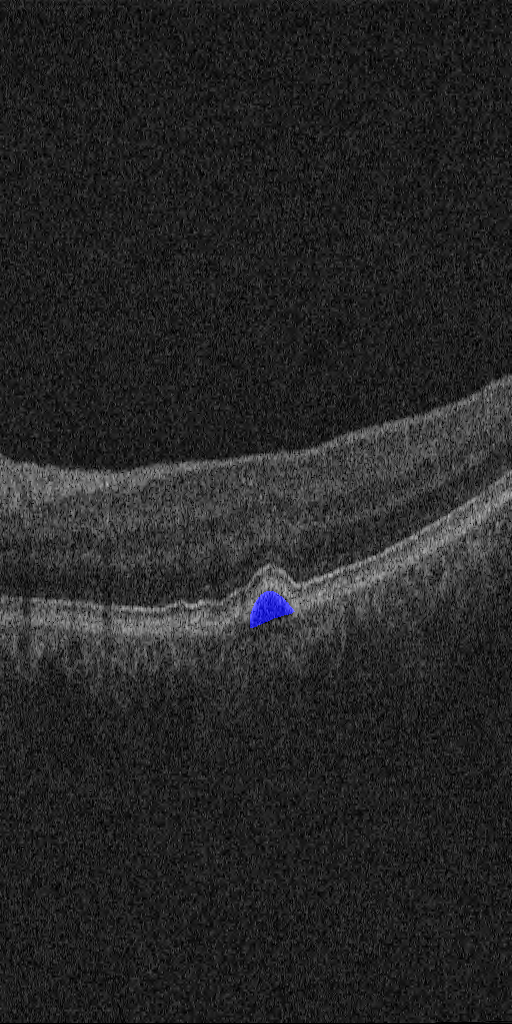}
\includegraphics[width=1\textwidth]{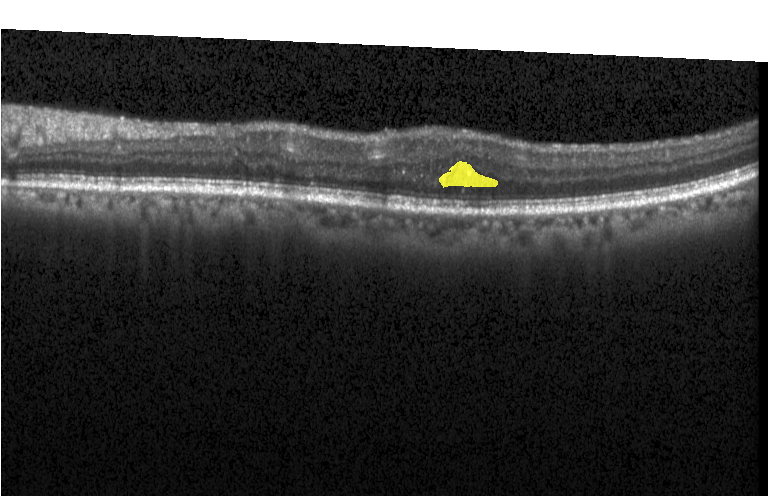}
\includegraphics[trim={0 0 0 3cm},clip,width=1\textwidth]{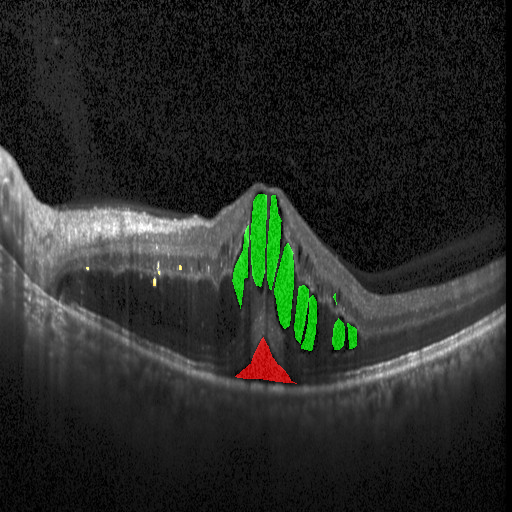}
\caption{GT}
\end{subfigure}
\begin{subfigure}{0.13\linewidth}
\includegraphics[trim={0 13cm 0 6cm},clip,width=1\textwidth]{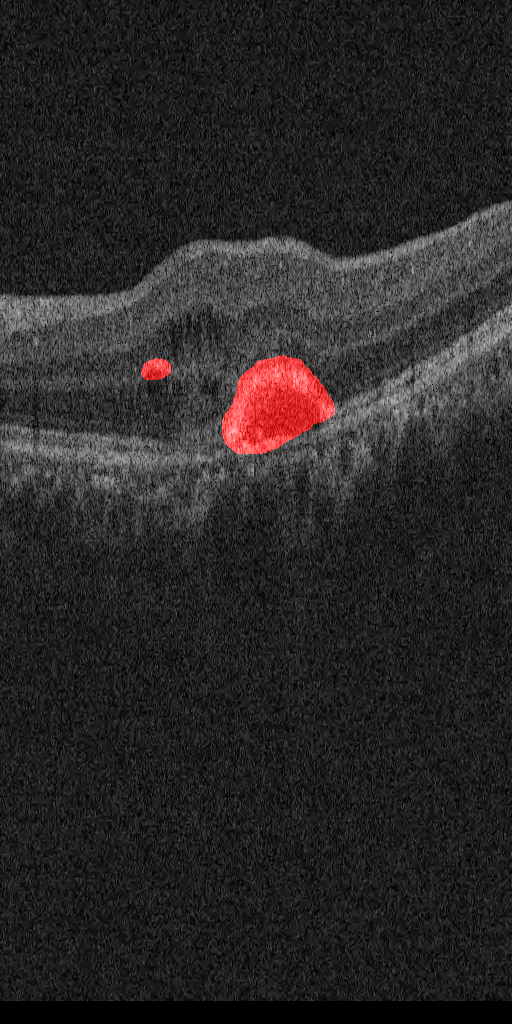}
\includegraphics[trim={0 3cm 0 14cm},clip,width=1\textwidth]{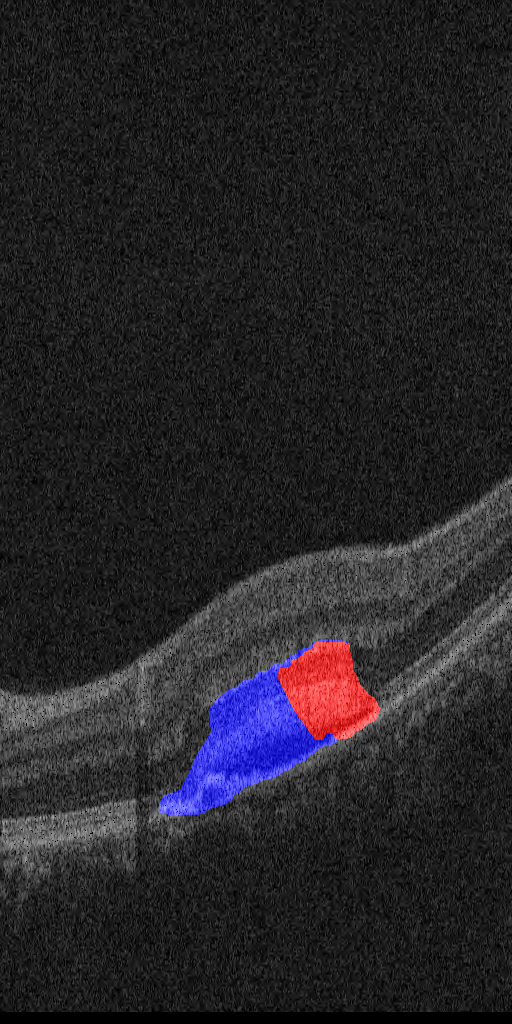}
\includegraphics[trim={0 9cm 0 12cm},clip,width=1\textwidth]{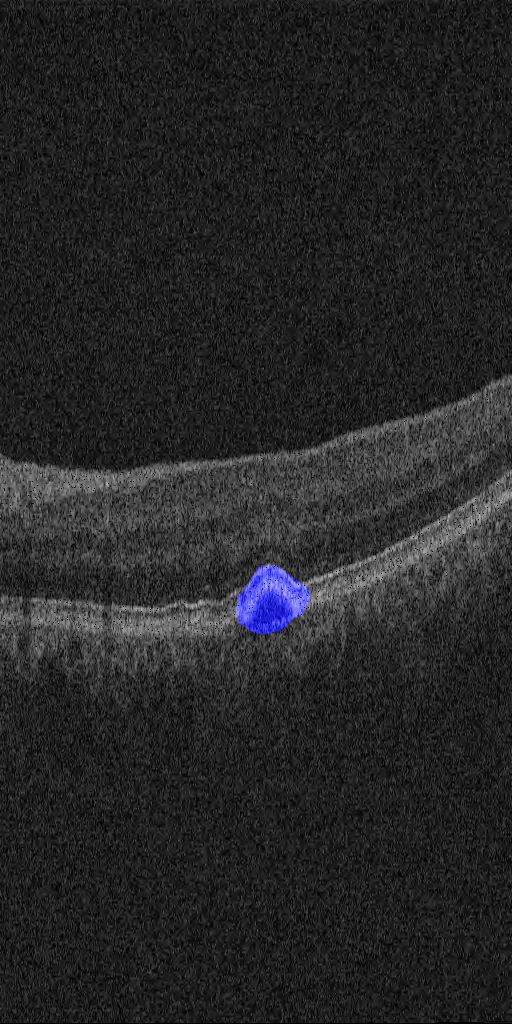}
\includegraphics[width=1\textwidth]{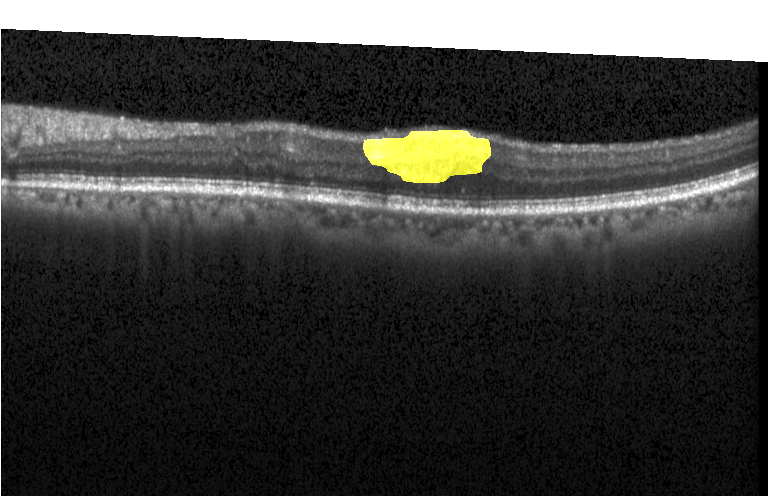}
\includegraphics[trim={0 0 0 3cm},clip,width=1\textwidth]{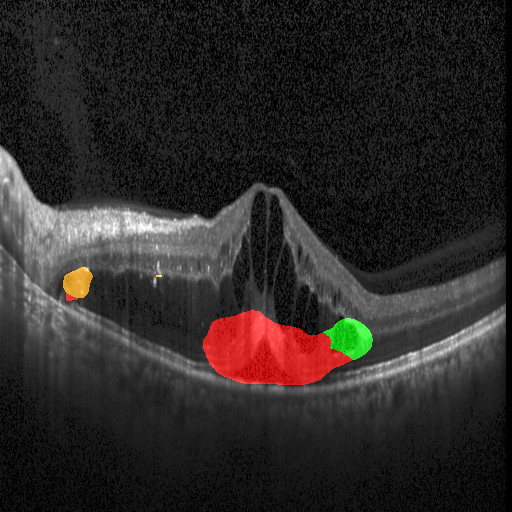}
\caption{SEAM}
\end{subfigure}
\begin{subfigure}{0.13\linewidth}
\includegraphics[trim={0 13cm 0 6cm},clip,width=1\textwidth]{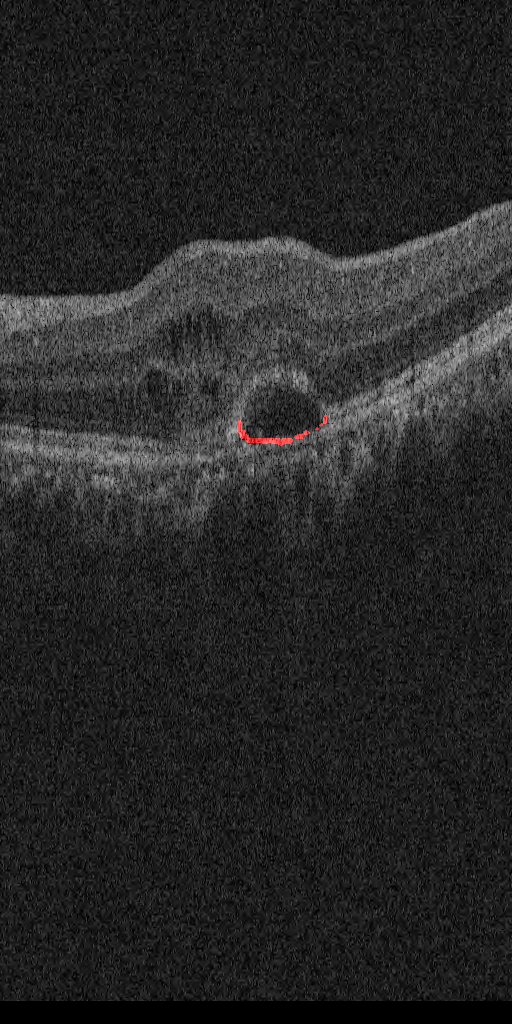}
\includegraphics[trim={0 3cm 0 14cm},clip,width=1\textwidth]{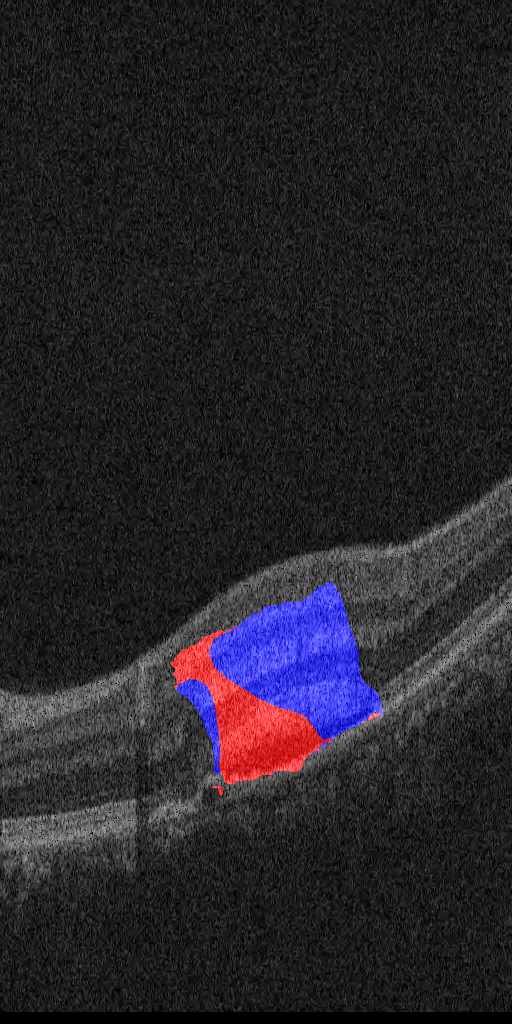}
\includegraphics[trim={0 9cm 0 12cm},clip,width=1\textwidth]{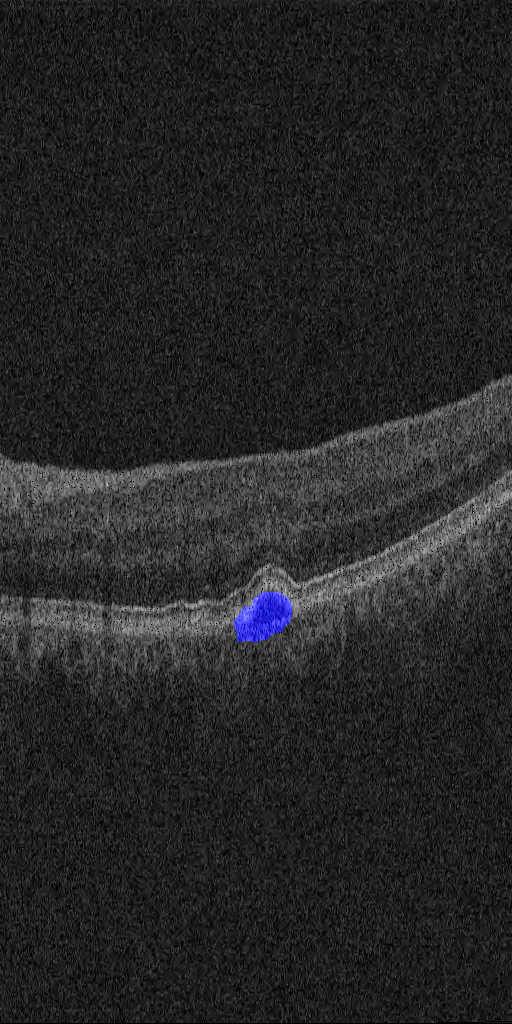}
\includegraphics[width=1\textwidth]{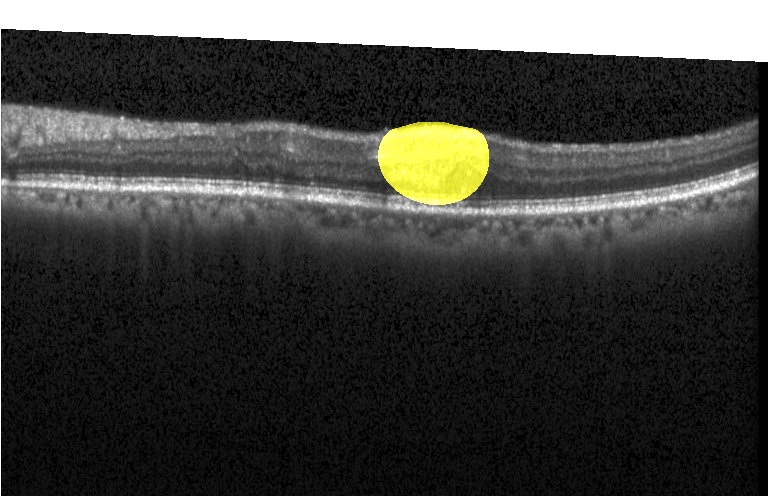}
\includegraphics[trim={0 0 0 3cm},clip,width=1\textwidth]{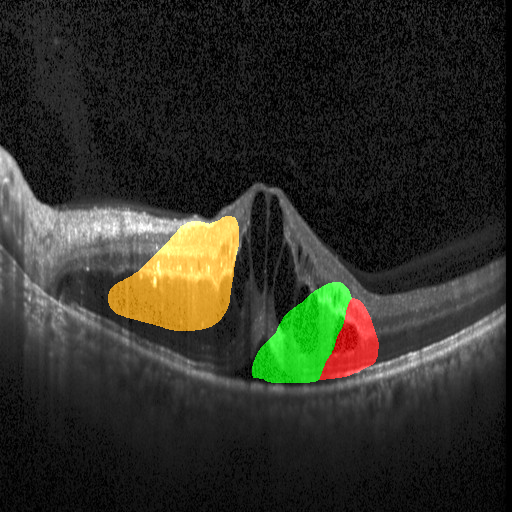}

\caption{ReCAM}
\end{subfigure}
\begin{subfigure}{0.13\linewidth}
\includegraphics[trim={0 13cm 0 6cm},clip,width=1\textwidth]{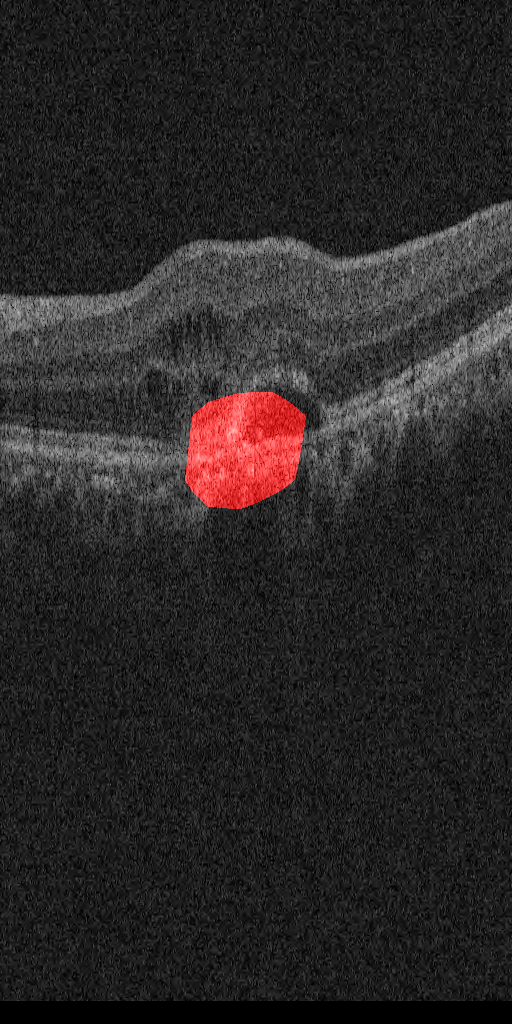}
\includegraphics[trim={0 3cm 0 14cm},clip,width=1\textwidth]{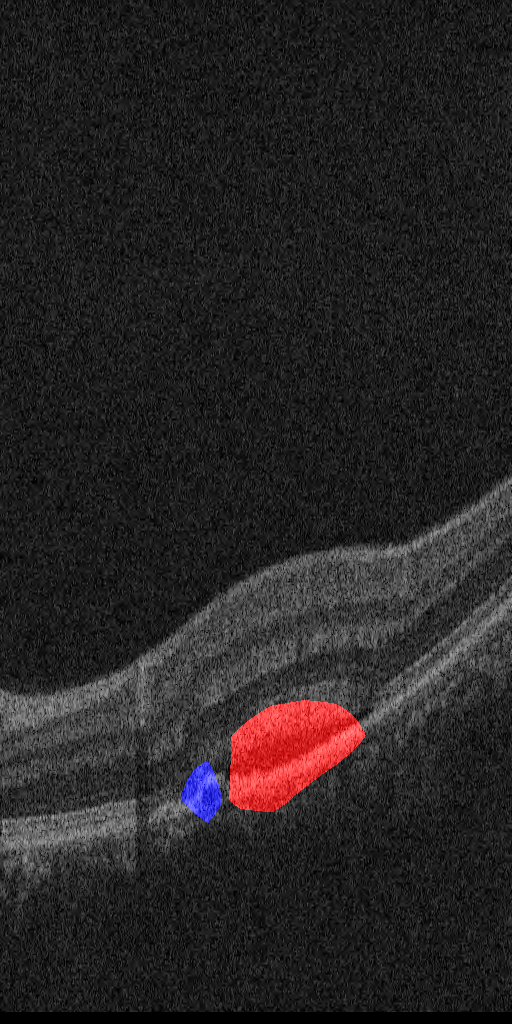}
\includegraphics[trim={0 9cm 0 12cm},clip,width=1\textwidth]{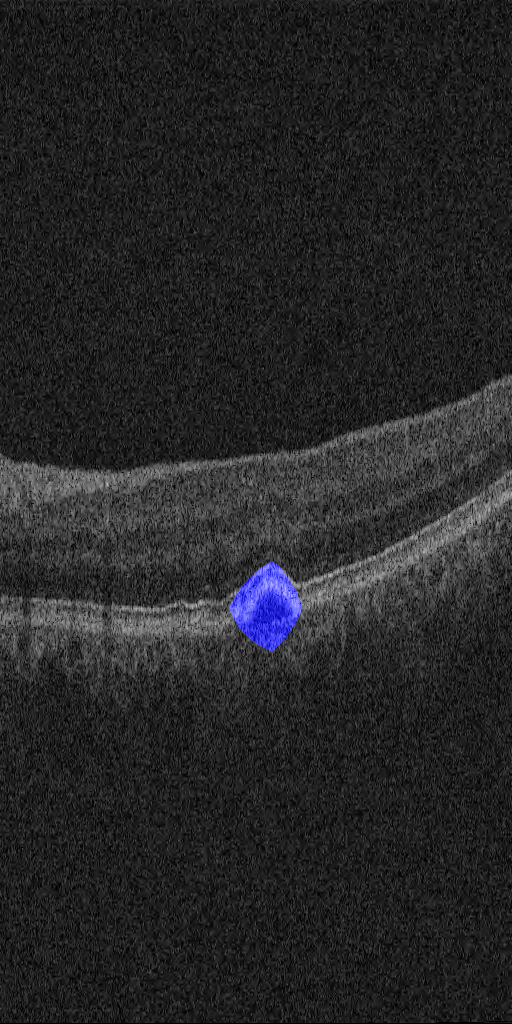}
\includegraphics[width=1\textwidth]{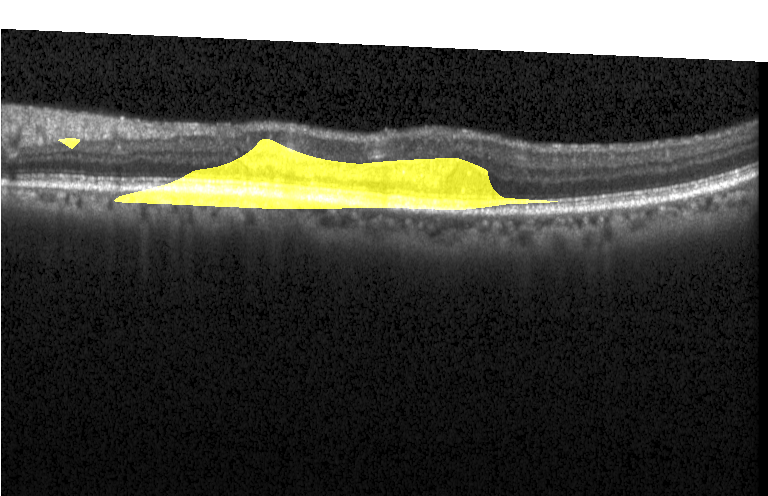}
\includegraphics[trim={0 0 0 3cm},clip,width=1\textwidth]{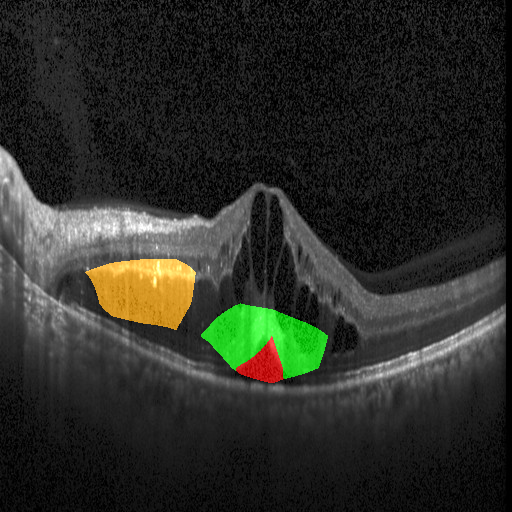}

\caption{AGM}
\end{subfigure}
\begin{subfigure}{0.13\linewidth}
\includegraphics[trim={0 13cm 0 6cm},clip,width=1\textwidth]{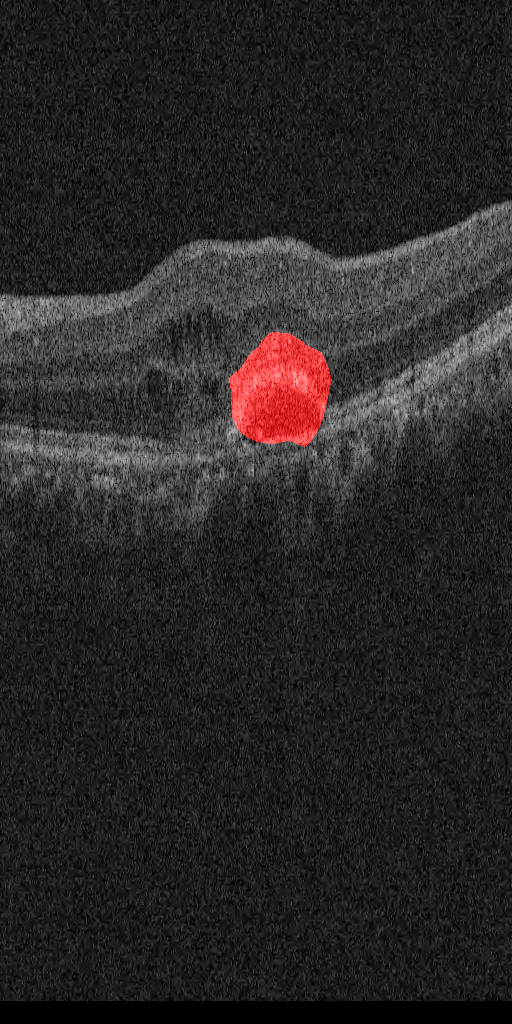}
\includegraphics[trim={0 3cm 0 14cm},clip,width=1\textwidth]{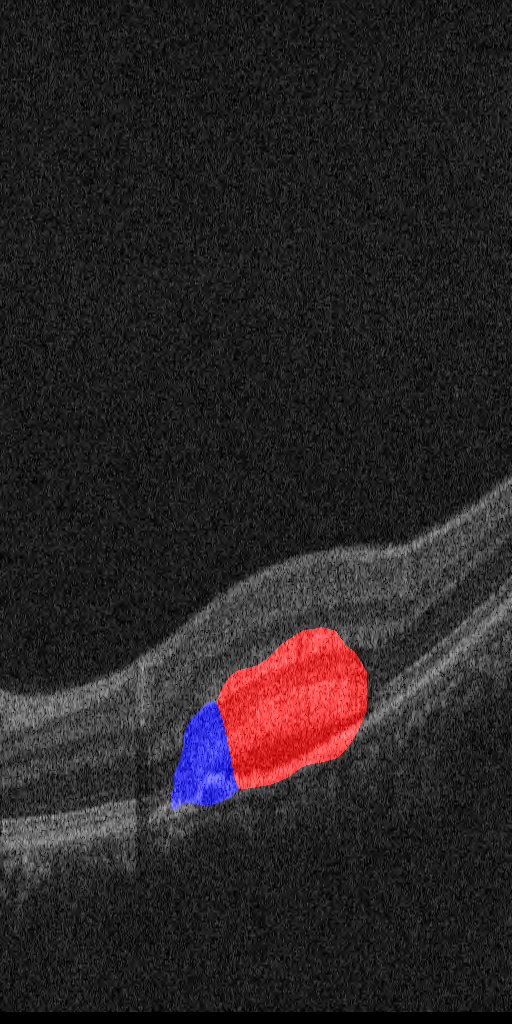}
\includegraphics[trim={0 9cm 0 12cm},clip,width=1\textwidth]{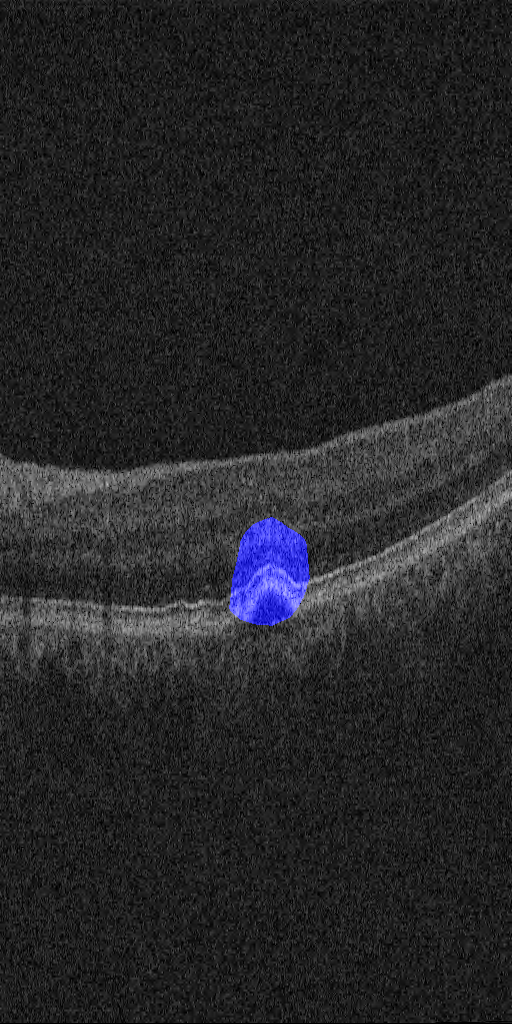}
\includegraphics[width=1\textwidth]{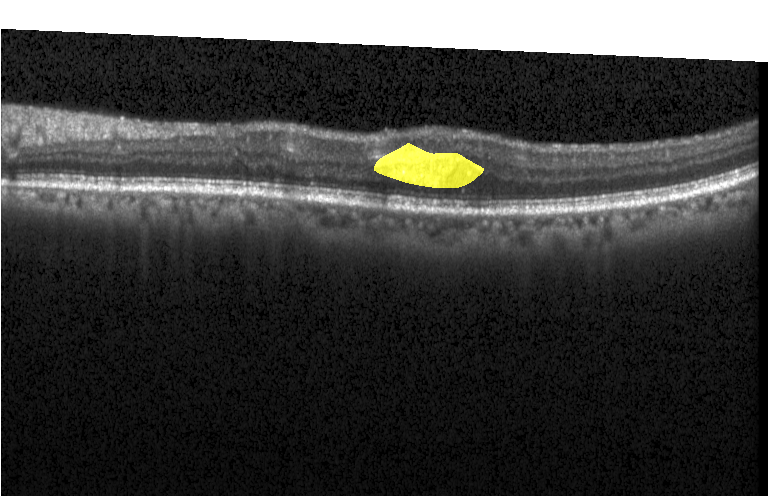}
\includegraphics[trim={0 0 0 3cm},clip,width=1\textwidth]{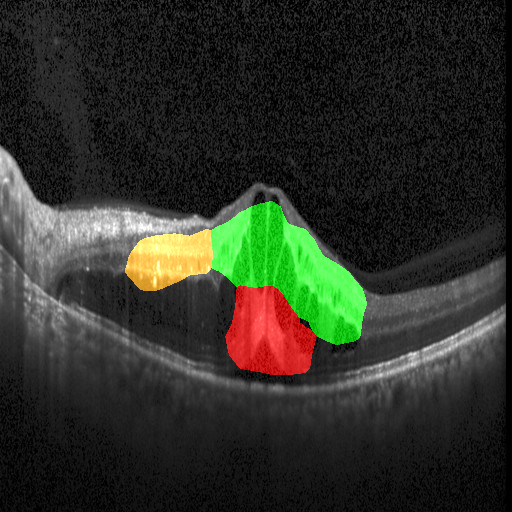}
\caption{TPRO}
\end{subfigure}
\begin{subfigure}{0.13\linewidth}
\includegraphics[trim={0 13cm 0 6cm},clip,width=1\textwidth]{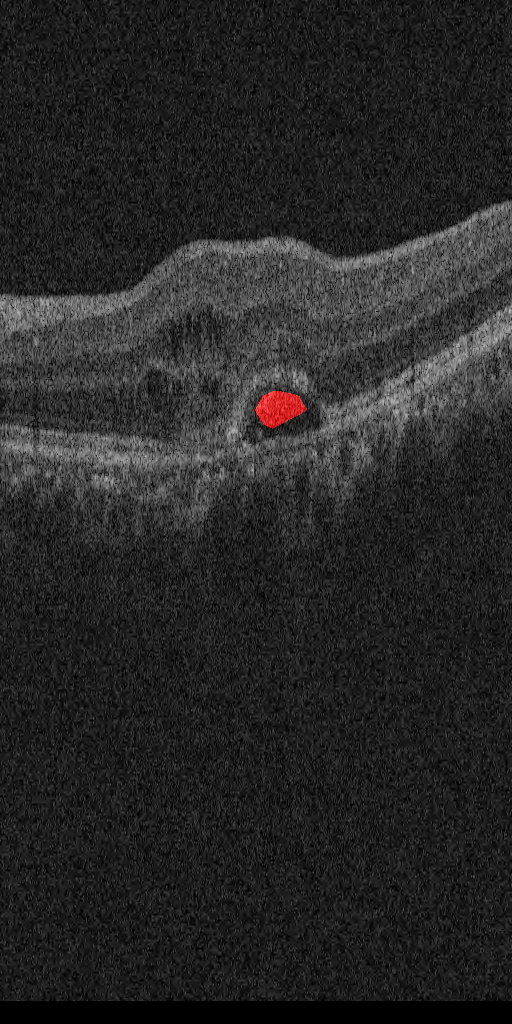}
\includegraphics[trim={0 3cm 0 14cm},clip,width=1\textwidth]{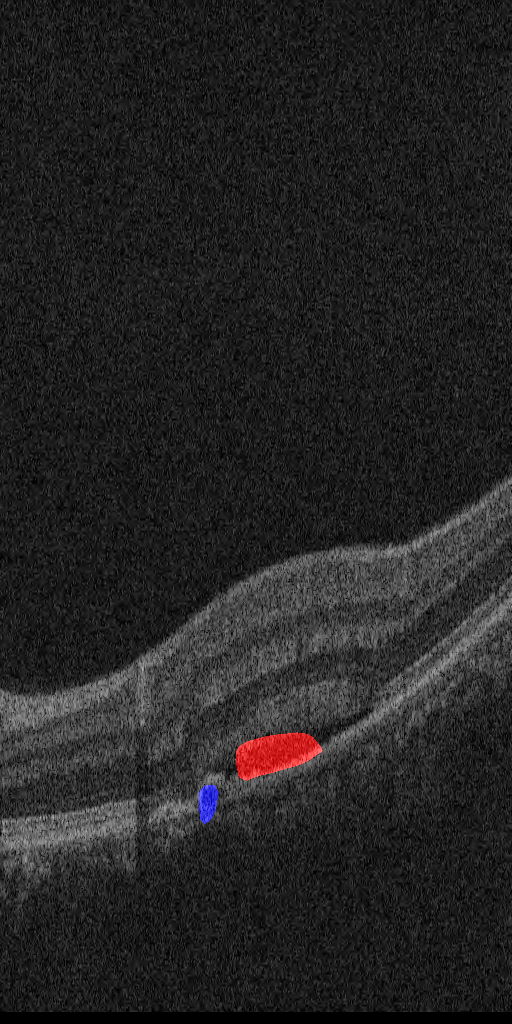}
\includegraphics[trim={0 9cm 0 12cm},clip,width=1\textwidth]{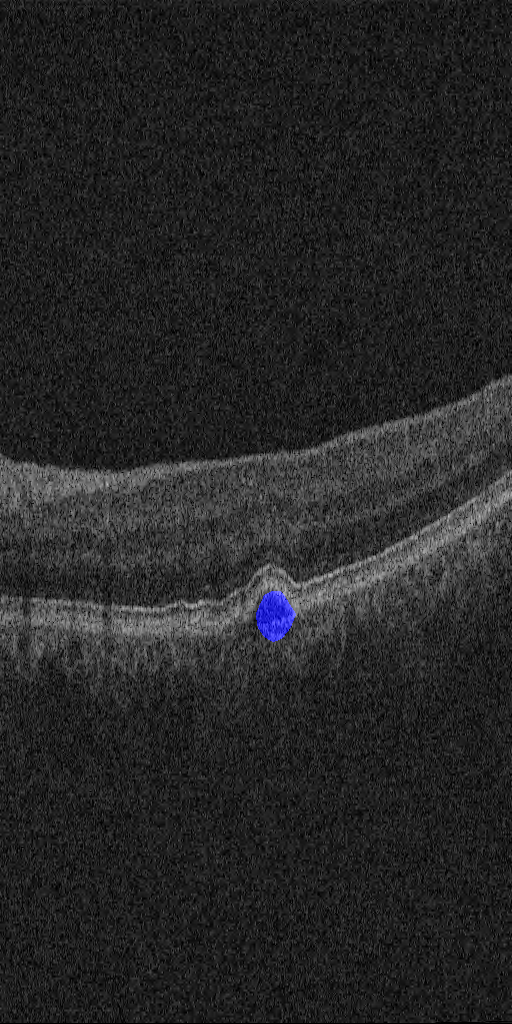}
\includegraphics[width=1\textwidth]{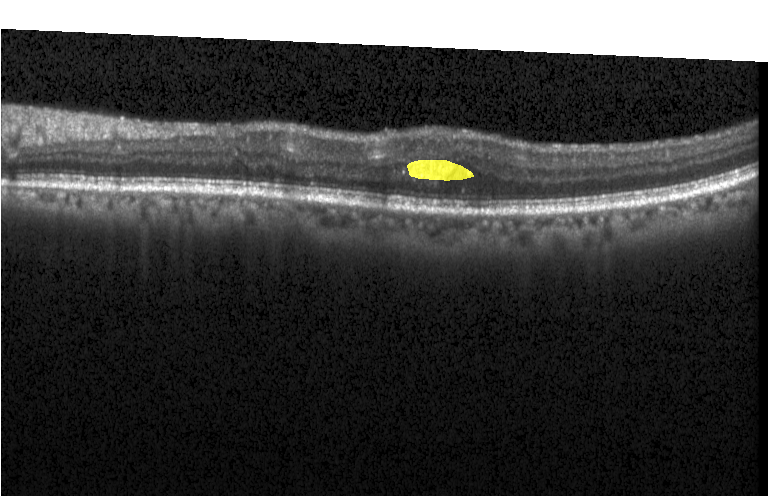}
\includegraphics[trim={0 0 0 3cm},clip,width=1\textwidth]{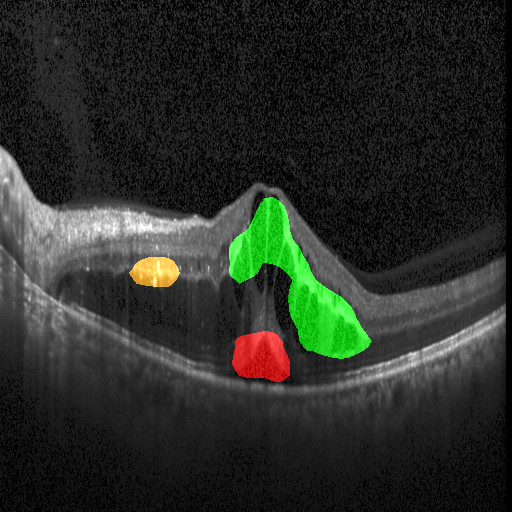}
\caption{\textbf{Ours}}
\end{subfigure}

\caption{Qualitative visualization of pseudo labels generated by baselines and our method on three validation sets. Each row presents a sample with ground truth shown in the second column. Pseudo labels are overlaid on the original images for clarity, with lesion colors as follows: red (SRF), blue (PED), yellow (Fluid), orange (HRD), and green (IRF)}
\label{fig:visualization}
\end{figure*}

In \cref{fig:visualization}, we present several examples comparing the pseudo labels generated by our proposed method to those from other baseline techniques. Our method achieves more accurate localization while minimizing incorrect predictions in irrelevant areas. In contrast, the baseline methods exhibit less stability across images. For example, ReCAM often produces overly sensitive predictions, such as focusing on edge regions for SRF (row 1) or generating misaligned predictions (rows 2 and 5). Similarly, TPRO often highlights irrelevant regions, as evident in its broader false positive coverage compared to our method. The closer alignment of our pseudo labels with the ground truth masks highlights the superior effectiveness of our approach in generating accurate and reliable predictions.

\begin{figure}[ht!]
\centering

\begin{subfigure}{0.12\linewidth}
\includegraphics[trim={0 15cm 0 6cm},clip,width=1\textwidth]{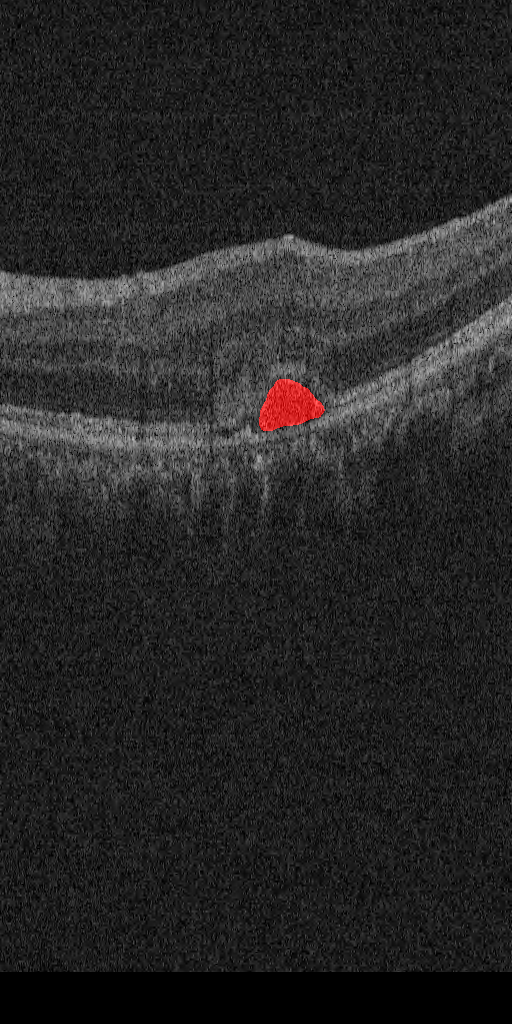}
\includegraphics[trim={0 18cm 0 3cm},clip,width=1\textwidth]{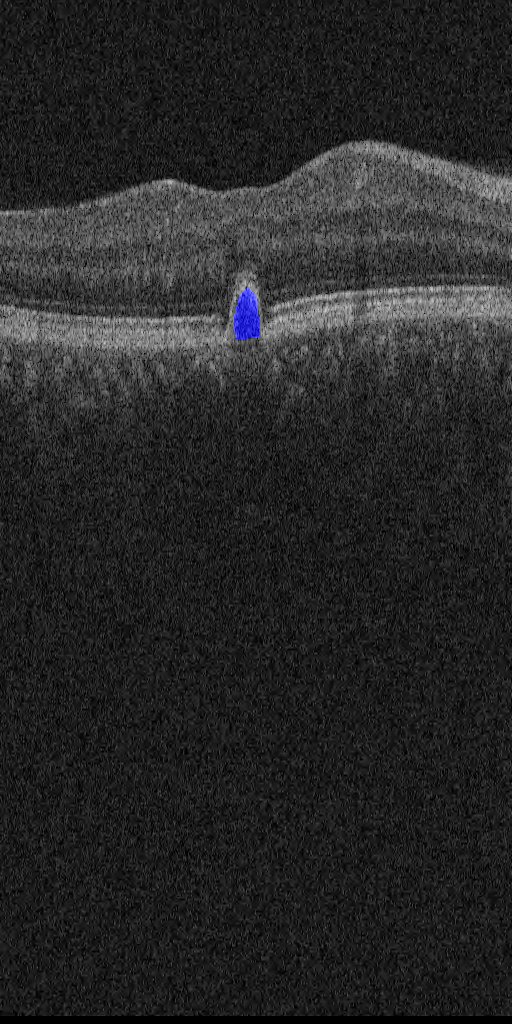}
\includegraphics[trim={0 3cm 0 15cm},clip,width=1\textwidth]{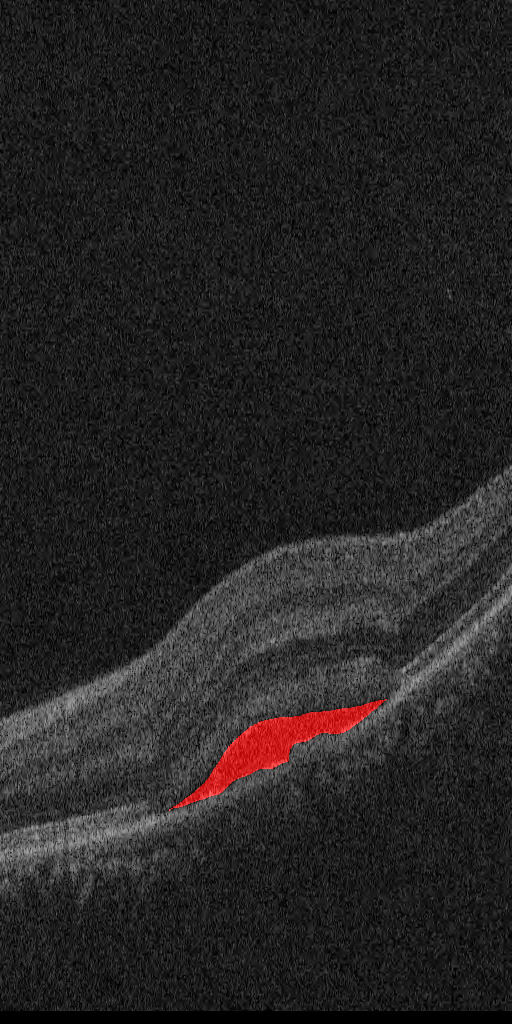}
\caption{GT}
\end{subfigure}
\begin{subfigure}{0.12\linewidth}
\includegraphics[trim={0 15cm 0 6cm},clip,width=1\textwidth]{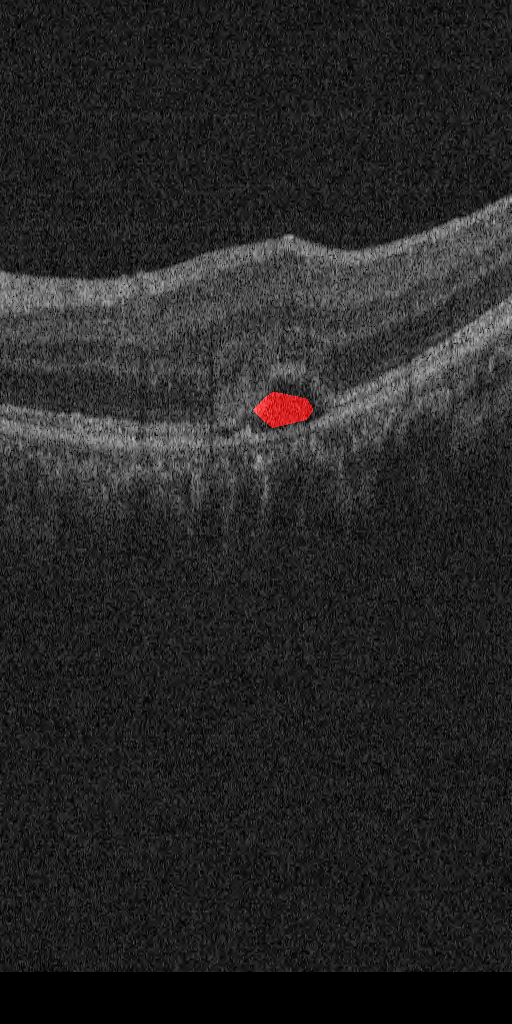}
\includegraphics[trim={0 18cm 0 3cm},clip,width=1\textwidth]{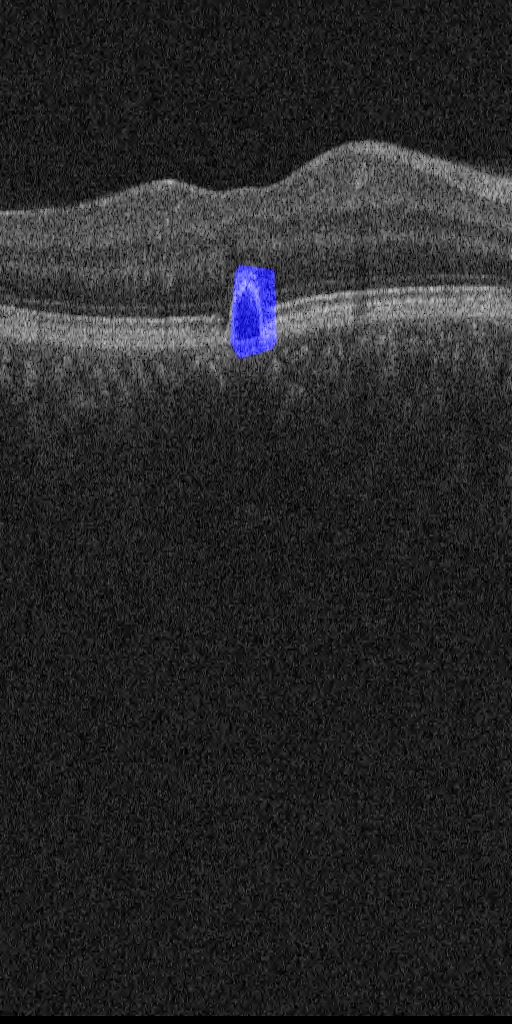}
\includegraphics[trim={0 3cm 0 15cm},clip,width=1\textwidth]{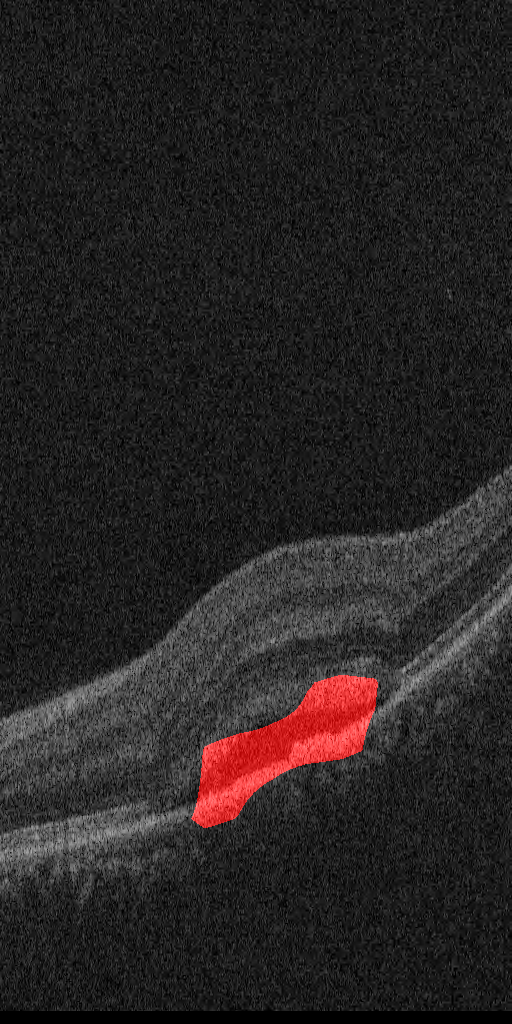}
\caption{$\mathbf{M}$}
\end{subfigure}
\begin{subfigure}{0.12\linewidth}
\includegraphics[trim={0 15cm 0 6cm},clip,width=1\textwidth]{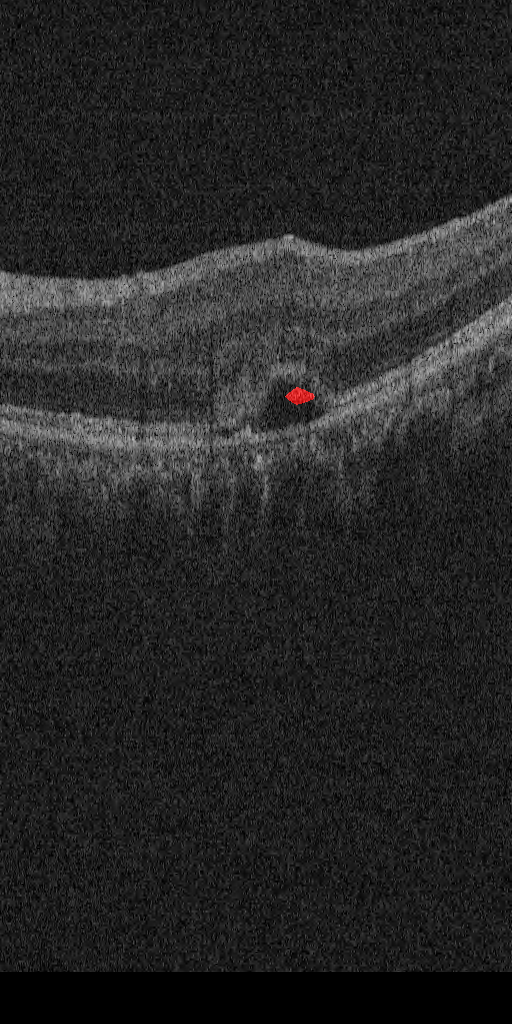}
\includegraphics[trim={0 18cm 0 3cm},clip,width=1\textwidth]{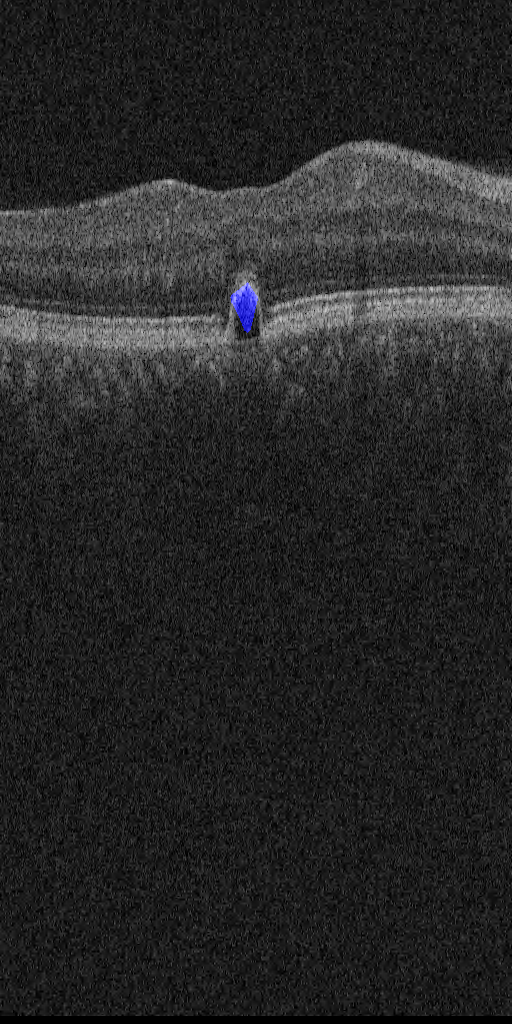}
\includegraphics[trim={0 3cm 0 15cm},clip,width=1\textwidth]{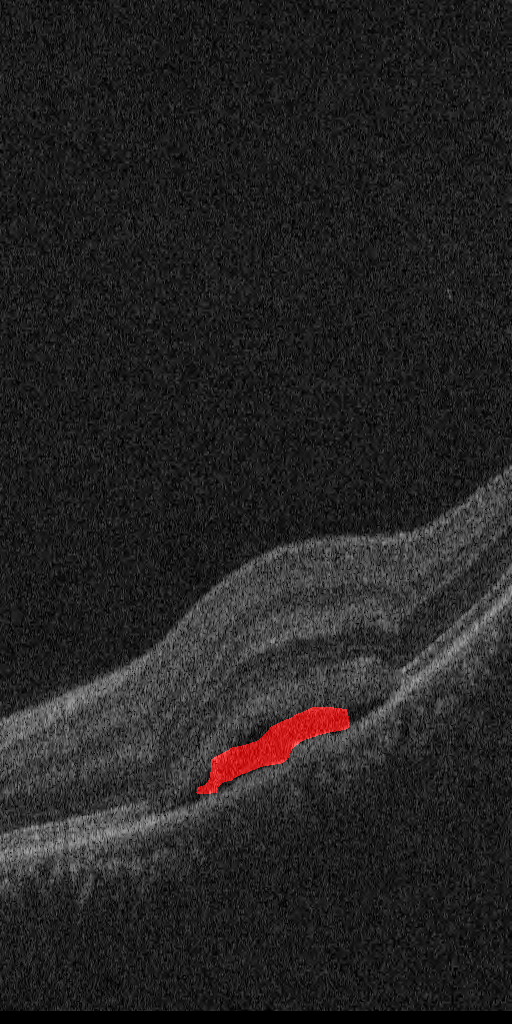}
\caption{$\text{SIM}_3$}
\end{subfigure}
\begin{subfigure}{0.12\linewidth}
\includegraphics[trim={0 15cm 0 6cm},clip,width=1\textwidth]{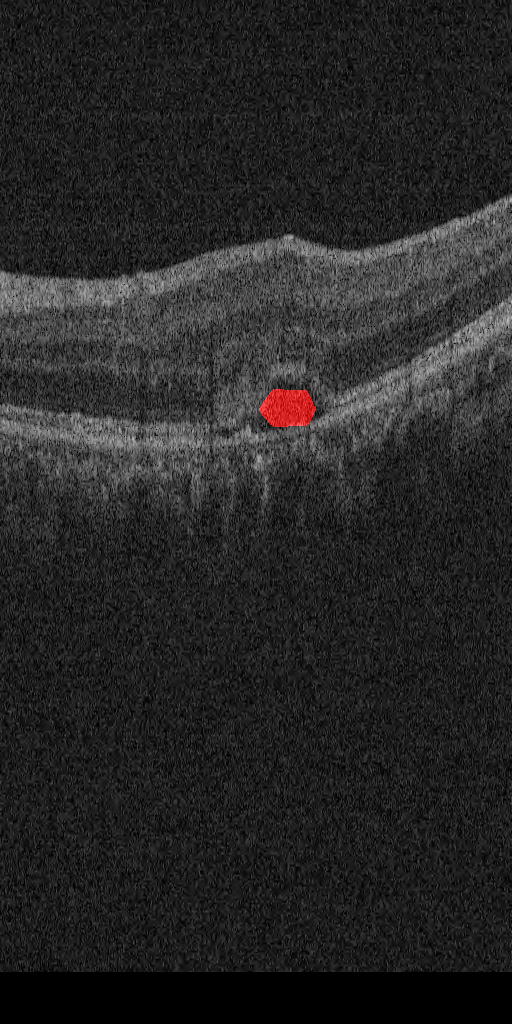}
\includegraphics[trim={0 18cm 0 3cm},clip,width=1\textwidth]{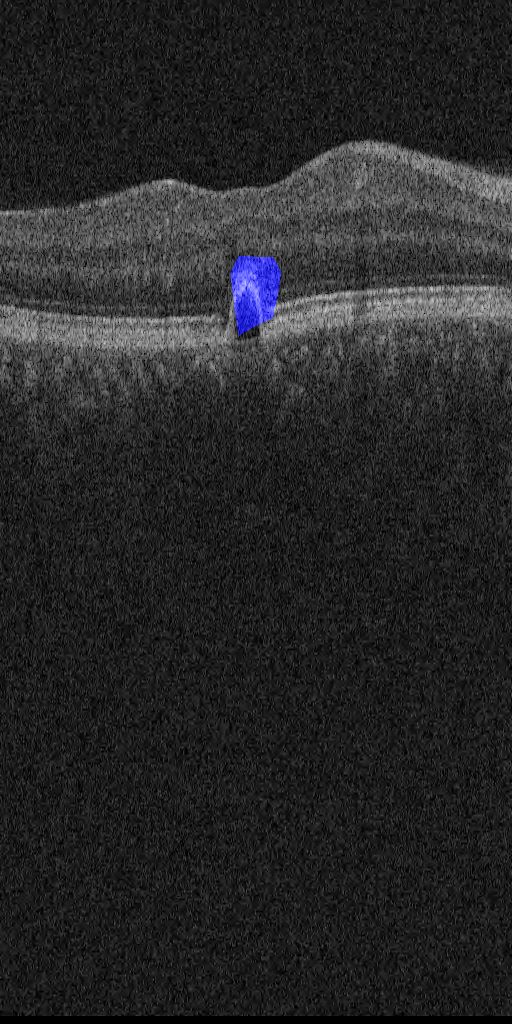}
\includegraphics[trim={0 3cm 0 15cm},clip,width=1\textwidth]{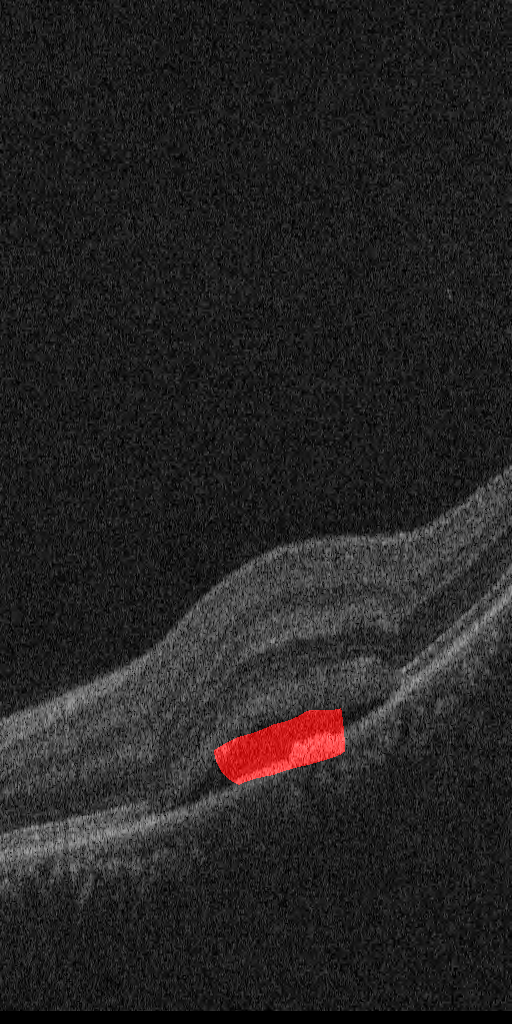}
\caption{$\text{SIM}_4$}
\end{subfigure}
\begin{subfigure}{0.12\linewidth}
\includegraphics[trim={0 15cm 0 6cm},clip,width=1\textwidth]{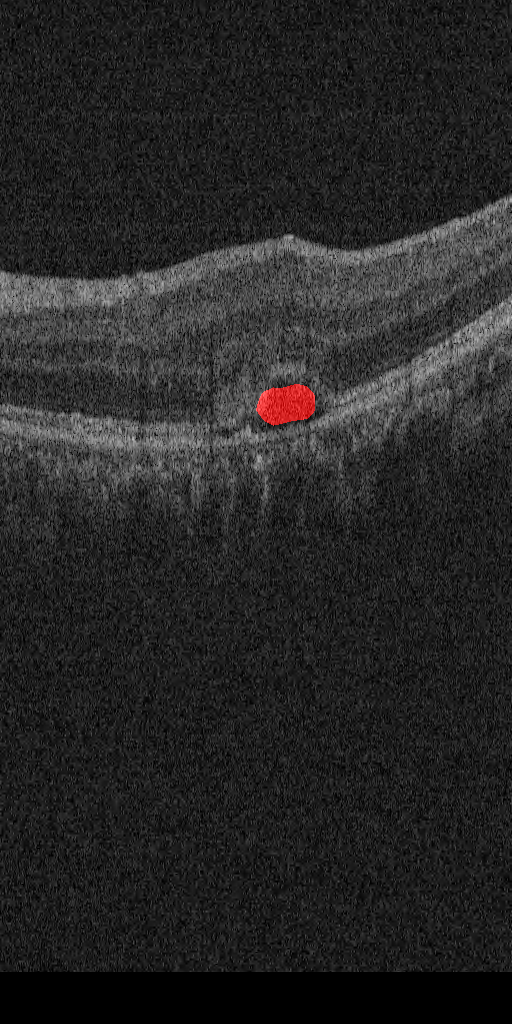}
\includegraphics[trim={0 18cm 0 3cm},clip,width=1\textwidth]{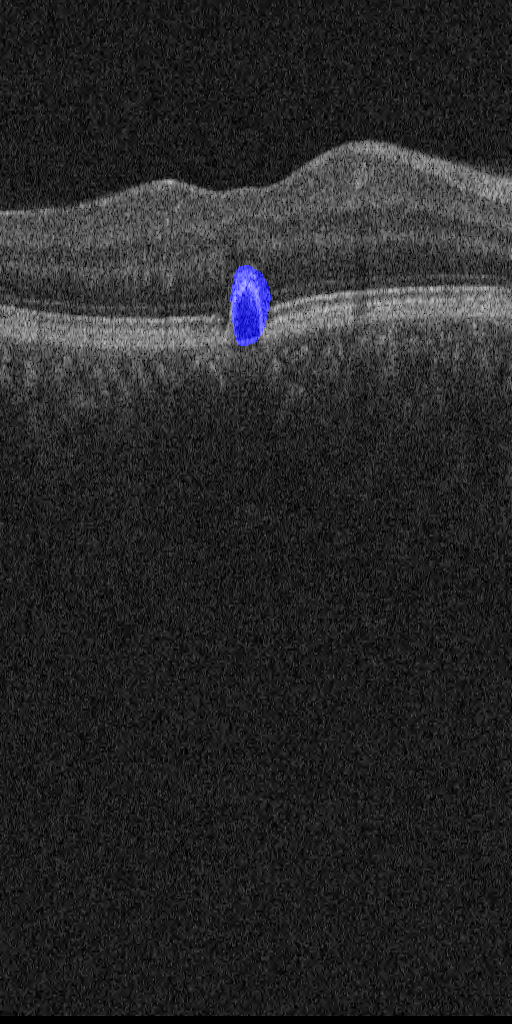}
\includegraphics[trim={0 3cm 0 15cm},clip,width=1\textwidth]{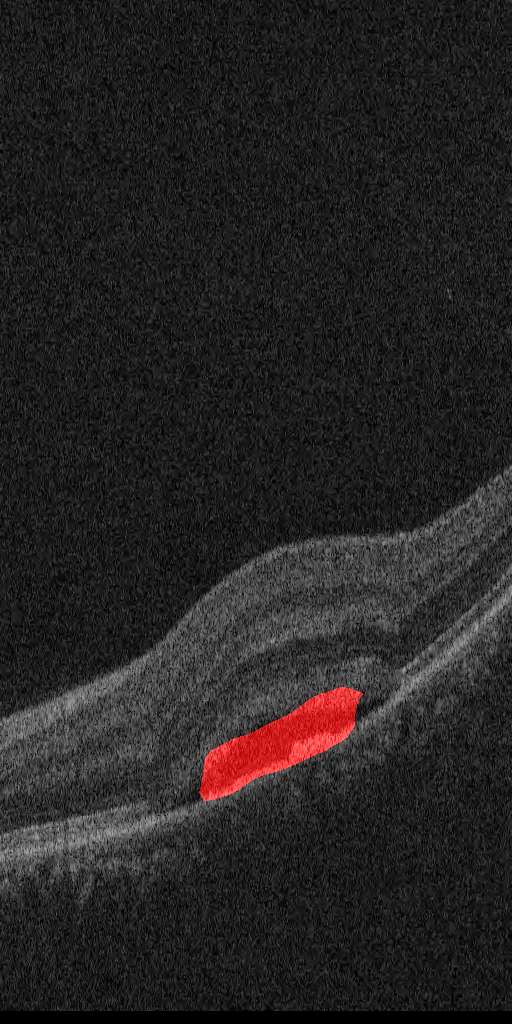}
\caption{\textbf{$\mathbf{M}_{final}$}}
\end{subfigure}

\caption{Visualization of pseudo labels generated from different sources of CAMs. (a) GT represents the ground truth, (b) $\mathbf{M}$ corresponds to the primary branch output, (c) $\text{SIM}_3$ is the similarity map between visual and textual features from stage 3, (d) $\text{SIM}_4$ is the similarity map from stage 4, and (e) $\mathbf{M}_{final}$ represents the proposed combined result integrating the three sources.}
\label{fig:label_informed}
\end{figure}

We present the mIoU performance for different sources of CAMs in \Cref{tab:ablation_clip_loss}. To further illustrate the impact of our proposed three sources of CAMs, we provide examples of the generated pseudo labels overlaid on the original OCT images in \cref{fig:label_informed}. The pseudo labels generated by $\mathbf{M}$, $\text{SIM}_3$, $\text{SIM}_4$, and $\mathbf{M}_{final}$ are described in Eq.~\eqref{heatmaps}. We can observe that $\text{SIM}_3$ is relatively more conservative to smaller regions compared to $\text{SIM}_4$, which makes sense as it is derived from the early stage of the network, where the features primarily capture low-level details and localized patterns. The results generated by the primary branch, $\mathbf{M}$, demonstrate its stable capability to capture lesions. When combined with $\text{SIM}_3$ and $\text{SIM}_4$, our final proposed $\mathbf{M}_{final}$ benefits from the complementary focus of all three sources, producing more stable and accurate results compared to the ground truth.

\begin{figure}[ht]
\centering
    \includegraphics[width=0.6\linewidth]{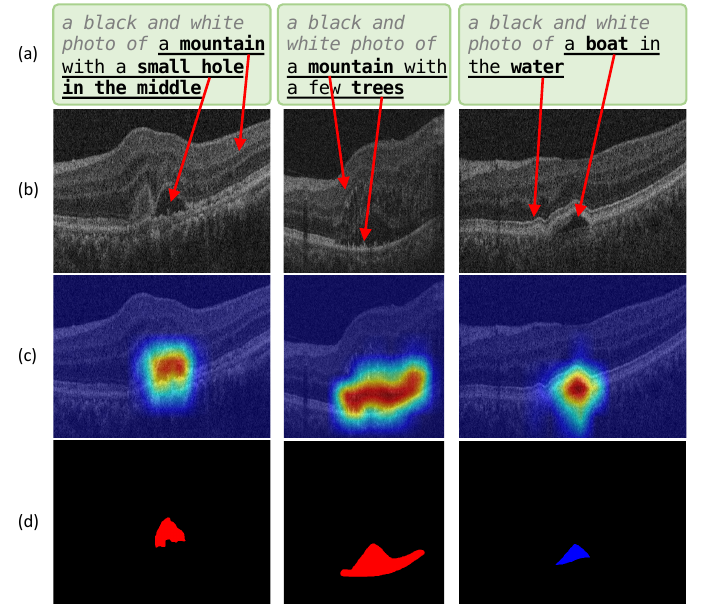}
\caption{Visualization of three examples of CAMs generated by our model with synthetic descriptions on the RESC dataset. Each example consists of: (a) synthetic descriptive text; (b) the original OCT image; (c) CAMs generated by our model; and (d) the ground truth lesion mask for comparison. Red arrows indicate areas in the OCT image that potentially correspond to highlighted keywords in the text.}
\label{fig:quant_caption}
\end{figure}

In \cref{fig:quant_caption}, we present three examples from the RESC dataset, each with synthetic text generated by BLIP and corresponding CAMs produced by our proposed method. Each description begins with the prefix \textit{``a black and white photo of,"} followed by the generated description, which provides a global perspective that leverages non-medical language to convey meaningful information about object relationships and relative positions. By comparing the text and visuals, we can observe that these non-medical descriptions offer effective cues for lesion localization. Red arrows indicate the regions in the OCT image that potentially correspond to key phrases in the synthetic text. For instance, in the first column of (a), the phrase \textit{``a small hole in the middle"} suggests the relative location of the lesion, which aligns well with the position of the SRF lesion. As shown in (c), our model accurately localizes the lesions with high confidence, demonstrating its effectiveness in integrating these textual features.

\section{Discussion and Conclusion}
In this paper, we propose a novel method for OCT lesion segmentation under a weakly supervised setting, using only image-level labels as supervision. Our approach leverages imperfect yet effective information to improve pseudo-label quality and enhance segmentation results. We incorporate structural information to capture potential relationships between certain lesions and their typical layer patterns. Meanwhile, synthetic textual data generated by non-medical pretrained models provides consistent global context, strengthening model robustness. Furthermore, label-derived textual features contribute a lesion-specific local perspective, further refining segmentation accuracy. We validated the effectiveness of our method across three OCT datasets, supported by comprehensive ablation studies examining the contribution of each module. Our approach establishes a new state-of-the-art, outperforming other baselines in weakly supervised lesion segmentation on retinal OCT images.

By incorporating synthetic descriptions generated from non-medical pretrained models, our method effectively leverages existing large-scale vision-language models to enhance segmentation performance. Although these models may introduce some noise due to the lack of alignment with medical-specific features, we have maximized their benefits by successfully integrating textual information into medical imaging analysis. This success presents an opportunity for further improvement: utilizing domain-specific models or incorporating medical reports could provide more specialized guidance and potentially enhance performance even further.

In terms of generalization to other non-OCT modalities, the structural branch relying on features unique to OCT images, such as layer segmentation, can limit the applicability to other types of medical imaging. However, the other components of our model, such as the use of cross-domain foundation models for supplemental textual information, can be readily adopted for various medical imaging modalities to exploit the information embedded in vision-language data.

In conclusion, we have demonstrated that leveraging structural, synthetic, and label-derived features significantly improves weakly supervised OCT lesion segmentation. Our findings validate the feasibility of incorporating textual information in OCT analysis and suggest the potential for integrating domain-specific descriptions in future work. We believe this approach presents a promising opportunity to reduce the need for extensive manual annotation through weakly supervised methods, while effectively leveraging multi-source information to strengthen model performance and support clinical decision-making.

\section*{Acknowledgments}
This work was partially supported by grants PSC-CUNY Research Award 65406-00 53, NSF CCF-2144901, and NIH R21CA258493.


\bibliographystyle{unsrtnat}

\end{document}